\documentclass[3p,authoryear,a4paper]{elsarticle}
\usepackage{graphicx}
\usepackage{slashbox}
\usepackage{color}
\usepackage{dsfont}
\usepackage{amsmath}
\usepackage{amssymb}
\usepackage{amsfonts}
\usepackage{amsbsy}
\usepackage{subcaption}
\usepackage{amsthm}
\usepackage{array}
\usepackage{booktabs} 
\usepackage{verbatim}
\usepackage[english]{babel}
\usepackage{natbib}
\usepackage{ulem}
\usepackage{url}
\usepackage{multirow}
\usepackage{subcaption}
\usepackage{float}
\usepackage{numprint}
\usepackage{enumitem}
\usepackage[title]{appendix}
\usepackage[font=normalsize, labelfont=bf, justification=centering]{caption}
\usepackage{tabularx}
\usepackage{ltablex}
\usepackage{makecell}
\usepackage[hidelinks]{hyperref}
\usepackage[table,xcdraw]{xcolor}
\usepackage[ruled,linesnumbered]{algorithm2e}
\makeatletter
\def\ps@pprintTitle{
  \let\@oddhead\@empty
  \let\@evenhead\@empty
  \def\@oddfoot{\centerline{\thepage}}
  \def\@evenfoot{\thepage\hfill}}
\makeatother

\numberwithin{table}{section}
\numberwithin{figure}{section}

\newcolumntype{R}{>{\raggedleft\arraybackslash}X} 
\newcolumntype{P}[1]{>{\centering\arraybackslash}p{#1}} 

\renewcommand\appendix{\par
\setcounter{section}{0}
\setcounter{subsection}{0}
\setcounter{table}{0}
\setcounter{figure}{0}
\gdef\thetable{\Alph{table}}
\gdef\thefigure{\Alph{figure}}
\gdef\thesection{\Alph{section}}
\setcounter{section}{0}}

\numberwithin{equation}{section}

\newenvironment{arcitem}{
\begin{list}{---}{
\topsep=1pt
\itemsep=1pt 
\parsep=0pt 
\leftmargin=19pt 
}}
{\end{list}}

\newcounter{arclist}

\newcounter{arcenum}
\newenvironment{arcenum}{
\begin{list}{\arabic{arcenum}.}{
\usecounter{arcenum}
\topsep=1pt
\itemsep=1pt 
\parsep=0pt 
\leftmargin=19pt 
}}
{\end{list}}

\begin{document}

\normalem

\begin{frontmatter}

\title{Machine Learning with High-Cardinality Categorical Features \\ in Actuarial Applications}

\cortext[cor]{Corresponding author. }
\address[UMelb]{Centre for Actuarial Studies, Department of Economics, University of Melbourne VIC 3010, Australia}
\address[UNSW]{School of Risk and Actuarial Studies, UNSW Australia Business School, UNSW Sydney NSW 2052, Australia}

\author[UMelb]{Benjamin Avanzi}
\ead{b.avanzi@unimelb.edu.au}

\author[UNSW]{Greg Taylor}
\ead{gregory.taylor@unsw.edu.au}

\author[UNSW]{Melantha Wang\corref{cor}}
\ead{wang.melantha@gmail.com}

\author[UNSW]{Bernard Wong}
\ead{bernard.wong@unsw.edu.au}

\begin{abstract}
High-cardinality categorical features are pervasive in actuarial data (e.g.~occupation in commercial property insurance). Standard categorical encoding methods like one-hot encoding are inadequate in these settings.

In this work, we present a novel \emph{Generalised Linear Mixed Model Neural Network} (``GLMMNet'') approach to the modelling of high-cardinality categorical features. The GLMMNet integrates a generalised linear mixed model in a deep learning framework, offering the predictive power of neural networks and the transparency of random effects estimates, the latter of which cannot be obtained from the entity embedding models. Further, its flexibility to deal with any distribution in the exponential dispersion (ED) family makes it widely applicable to many actuarial contexts and beyond.

We illustrate and compare the GLMMNet against existing approaches in a range of simulation experiments as well as in a real-life insurance case study. Notably, we find that the GLMMNet often outperforms or at least performs comparably with an entity embedded neural network, while providing the additional benefit of transparency, which is particularly valuable in practical applications.

Importantly, while our model was motivated by actuarial applications, it can have wider applicability. The GLMMNet would suit any applications that involve high-cardinality categorical variables and where the response cannot be sufficiently modelled by a Gaussian distribution.
\end{abstract}

\end{frontmatter}

\section{Introduction}\label{sec:intro}
\subsection{Background} \label{ssec:background}
The advances in machine learning (ML) over the past two decades have transformed many disciplines. Within the actuarial literature, we see an explosion of works that apply and develop ML methods for various actuarial tasks \citep[see][]{richman2021b, richman2021a}. Meanwhile, there are many distinctive challenges with insurance data that are not addressed by general-purpose ML algorithms.

One prominent example is the usual presence of high-cardinality categorical features (i.e.~categorical features with many levels or \textit{categories}). These features often represent important risk factors in actuarial data. Examples include the occupation in commercial property insurance, or the cause of injury in workers' compensation insurance. Unfortunately, ML algorithms cannot ``understand'' (process) categorical features on their own.

The classical textbook approach to this problem, and also the most popular one in practice, is \textit{one-hot encoding}. It turns a categorical feature of $q$ unique categories into numeric representations by constructing $q$ binary attributes, one for each category; for example, a categorical feature with three unique categories will be represented as $[1, 0, 0]$, $[0, 1, 0]$ and $[0, 0, 1]$.

One-hot encoding works well with a small number of independent categories, which is the case with most applications in the ML literature \citep[e.g.][]{hastie2009}. However, issues arise as cardinality expands: (i) The orthogonality (i.e.~independence) assumption of one-hot encoding no longer seems appropriate, since the growing number of categories will inevitably start to interact as the feature space gets more and more crowded, (ii) The resultant high-dimensional feature matrix entails computational challenges, especially when used with already computationally expensive models such as neural networks; (iii) The often uneven distribution of data across categories makes it difficult to learn the behaviour of the rare categories.

For a motivating example, consider the workers' compensation scheme of the \citet{stateofnewyork2022}. In the four million claims observed from 2000 to 2022, the cause of injury variable has 78 unique categories, with more than 200,000 observations (about 5\%) for the most common cause (lifting) and less than 1000 observations (0.025\%) for the 10 least common causes (e.g.~crash of a rail vehicle, or gunshot). The uneven coverage of claims presents a modelling challenge and has been documented by actuarial practitioners \citep[e.g.][]{pettifer2012}. Furthermore, the cause of injury variable alone may not serve well to differentiate claim severities. Injuries from the same cause can result in vastly different claims experiences. It seems natural to consider its interaction with the nature of injury and part of body variables, each with 57 reported categories. Exploring $4446~(78 \times 57)$ interaction categories is clearly infeasible with one-hot encoding.

The example above highlights that one-hot encoding is an inadequate tool for handling high-cardinality categorical features. A few alternatives exist but also have their own drawbacks, as listed below. It is worth pointing out that while the following approaches appear very distinct, they share one common feature. They all encourage the categories to ``communicate'' with each other, in the sense that the information learned from one category should also help with learning of the others.

\begin{enumerate}
\def\labelenumi{(\roman{enumi})}
\item
  \emph{Manual (or data-guided) regrouping of categories of similar risk behaviours}. The goal here is to reduce the number of categories so that the refined categories can be tackled with the standard one-hot encoding scheme. This is a working approach, but manual regrouping requires significant domain inputs, which are expensive. Furthermore, data-driven methods such as clustering \citep[see e.g.][]{guiahi2017} require that data be first aggregated by categories, and this aggregation gives away potentially useful information available at more granular levels.
\item
  \emph{Entity embeddings from neural networks}. Proposed by \citet{guo2016}, it seems to be now the most popular approach in the ML-driven stream of actuarial literature \citep{delong2021, shi2021, kuo2021a}. Entity embeddings work to extract a low-dimensional numeric representation of each category, so that categories closer in distance would observe similar response values. However, as the entity embeddings are trained as an early component of a black-box neural network, they offer little transparency towards their effects on the response.
\item \emph{Generalised linear mixed models (GLMM) with the high-cardinality categorical features modelled as random effects}. This is another popular approach among actuaries, partly due to its high interpretability \citep[][]{antonio2007, verbelen2019}. However, GLMMs as an extension to GLMs also inherit their limitations---the same ones that have motivated actuaries to start turning away from GLMs and exploring ML solutions \citep[see, e.g.][]{al-mudafer2022,henckaerts2021}.
\end{enumerate}

Some recent work has appeared in the ML literature aiming to extend GLMMs to take advantage of the more capable ML models, by embedding a ML model into the linear predictor of the GLMMs. The most notable examples include GPBoost \citep{sigrist2022} that combines GLMMs with a gradient boosting algorithm, and LMMNN \citep{simchoni2022} that combines linear mixed models (LMM) with neural networks. Unfortunately, these models present their own limitations in the face of insurance data: GPBoost assumes overly optimistic random effects variance, and LMMNN is limited to a Gaussian-distributed response. While the Gaussian assumption offers computational advantages (in terms of analytical tractability), it is often ill suited to the more general distributions that are required for the modelling of financial and insurance data.

None of the existing techniques appears satisfactory for the modelling of high-cardinality categorical features. This paper seeks an alternative approach that more effectively leverages information in the categories, offers greater transparency than entity embeddings, and demonstrates improved predictive performance. In the next section, we introduce our approach.

\subsection{Contributions}
Our work presents two main contributions.

First, we take inspiration from the latest ML developments and propose a generalised mixed effects neural network called \textbf{GLMMNet}. The GLMMNet is an extension to the LMMNN proposed by \citet{simchoni2022}. In a similar vein, the GLMMNet fuses a deep neural network to the GLMM structure: The \textbf{Net}work component of the model provides it with the flexibility required to capture complex non-linear relationships in the data, and the \textbf{GLMM}-like component of the model allows a probabilistic interpretation and offers full transparency on the categorical variables (through the random effects estimates). In contrast with the Gaussian-based LMMNN, the key enhancement of the GLMMNet is its ability to model the entire class of exponential dispersion (ED) family distributions. The proposed extension requires a significant effort as it involves moving away from analytical results. The flexibility it brings, however, makes the GLMMNet approach widely applicable to many insurance contexts, such as in the prediction of claim frequency or the estimation of pure risk premiums.

Secondly, we provide a systematic empirical comparison of some of the most popular approaches for modelling high-cardinality categorical features. Although the difficulty with such variables has been a long-standing issue in actuarial modelling, to the best of our knowledge, this paper is the first piece of work to extensively compare a range of the existing approaches (including our own GLMMNet). We believe that such a benchmarking study will be valuable, especially to practitioners facing many options available to them.

Importantly, the methodological extensions proposed in this paper, while motivated by actuarial applications, are not limited to this context and can have much wider applicability. The extension of the LMMNN to the GLMMNet would suit any applications where the response cannot be sufficiently modelled by a Gaussian distribution. This unparalleled flexibility with the form of the response distribution, in tandem with the computational efficiency of the algorithm (due to the fast and highly scalable variational inference used to implement GLMMNet), make the GLMMNet a promising tool for many practical ML applications.

\subsection{Outline of Paper}
In Section \ref{sec:glmmnet}, we introduce the GLMMNet, which fuses neural networks with GLMMs to enjoy both the predictive power of deep learning models and the statistical strength---specifically, the interpretability and likelihood-based estimation---of GLMMs. In Section \ref{sec:simulation}, we illustrate and compare the proposed GLMMNet against a range of alternative approaches across a spectrum of simulation environments, each to mimic some specific elements of insurance data commonly observed in practice. Section \ref{sec:case-study} presents a real-life insurance case study and Section \ref{sec:conclusion} concludes.

The code used in the numerical experiments is available on \url{https://github.com/agi-lab/glmmnet}.

\section{GLMMNet: A Generalised Mixed Effects Neural Network} \label{sec:glmmnet}
This section describes our GLMMNet in detail. We start by defining some notation in Section~\ref{ssec:problem}. We then provide an overview of GLMMs in Section~\ref{ssec:glmm} to set the stage for the introduction of the GLMMNet. Sections~\ref{ssec:glmmnet}--\ref{ssec:glmmnet-predict} showcase the architectural design of the GLMMNet and provide the implementation details.

\subsection{Setting and Notation} \label{ssec:problem}
We first introduce some notation to formalise the problem discussed above.

We write \(Y\) to denote the \emph{response variable}, which is typically one of claim frequency, claim severity or risk premium for most insurance problems. We assume that the distribution of \(Y \in \mathcal{Y}\) depends on some \emph{covariates} or \emph{features}. In this work, we draw a distinction between standard features \(\mathbf{x} \in \mathcal{X}\) (i.e.~numeric features and low-dimensional categorical features) and the high-cardinality categorical features \(\mathbf{z} \in \mathcal{Z}\) consisting of \(K\) features each with \(q_i~(i = 1, \cdots, K)\) categories. The latter is the subject of our interest. All categorical features have been one-hot encoded in the representations $\mathbf{x}$ and $\mathbf{z}$.

We assume \(\mathbf{x} \in \mathbb{R}^p\) and \(\mathbf{z} \in \mathbb{R}^q\), where \(p \in \mathbb{Z}\) and \(q = \sum_{i=1}^K q_i \in \mathbb{Z}\) represent the respective dimensions of the vectors. The empirical observations (i.e.~data) are denoted \(\mathcal{D}=(y_i, \mathbf{x}_i, \mathbf{z}_i)_{i=1}^n\), where \(n\) is the number of observations in the dataset. For convenience, let us also write \(\mathbf{y}=[y_1, y_2, \cdots, y_n]^\top \in \mathbb{R}^n, \mathbf{X} = [\mathbf{x}_1, \mathbf{x}_2, \cdots, \mathbf{x}_n]^\top \in \mathbb{R}^{n \times p}\) and \(\mathbf{Z} = [\mathbf{z}_1, \mathbf{z}_2, \cdots, \mathbf{z}_n]^\top \in \mathbb{R}^{n \times q}\). In general, we use bold font to denote vectors and matrices.

The purpose of any predictive model is to learn the conditional distribution \(p(y|\mathbf{x}, \mathbf{z})\). It is generally assumed that the dependence of the response \(Y\) on the covariates is through a true but unknown function \(\mu:(\mathcal{X}, \mathcal{Z}) \to \mathcal{Y}\), such that
\begin{align}
p(y|\mathbf{x}, \mathbf{z}) = p(y|\mu(\mathbf{x}, \mathbf{z}), \boldsymbol{\phi}), \label{eq:predictive}
\end{align}
where \(\boldsymbol{\phi}\) represents a vector of any additional or auxiliary parameters of the likelihood. Most commonly $\mu(\mathbf{x}, \mathbf{z})$ will be the conditional mean \(\mathbb{E}(Y|\mathbf{x}, \mathbf{z})\), although it can be interpreted more generally. The predictive model then approximates \(\mu(\cdot)\) by some parametric function \(\hat{\mu}(\cdot)\). Specifically, a linear regression model will assume a linear structure for \(\mu(\cdot)\) with \(\boldsymbol{\phi} = \sigma^2_\epsilon\) to account for the error variance. Indeed, for most ML models, the function $\mu(\cdot)$ is the sole subject of interest (as opposed to the entire distribution $p(y|\mathbf{x}, \mathbf{z})$). There are many different options for choosing the function $\mu(\cdot)$. We discuss and compare the main ones in this paper; see also Section \ref{ssec:candidates}.


In what follows, for notational simplicity we will now assume that \(\mathbf{z}\) consists of only one (high-cardinality) categorical feature, i.e.~\(K = 1\) and \(q = q_1\). The extension to multiple categorical features, however, is very straightforward; see also Section~\ref{sssec:re}.

\subsection{Generalised Linear Mixed Models} \label{ssec:glmm}
Mixed models were originally introduced to model multiple correlated measurements on the same subject (e.g. longitudinal data), but they also offer an elegant solution to the modelling of high-cardinality categorical features \citep{frees2014}. We now give a brief review of mixed models in the latter context. For further details, the books by \citet{mcculloch2001, gelman2007, denuit2019glm} are all excellent references for this topic.

Let us start with the simplest linear model case. A one-hot encoded linear model assumes
\begin{align}
y | \mathbf{x}, \mathbf{z} \sim \mathcal{N}(\mu(\mathbf{x}, \mathbf{z}), \sigma_\epsilon^2), \text{ with } \mu(\mathbf{x}, \mathbf{z}) = \beta_0 + \mathbf{x}^\top \boldsymbol{\beta} + \mathbf{z}^\top \boldsymbol{\alpha}, \label{eq:nopool}
\end{align}
where \(\mathbf{x}\) are standard features (e.g.~sum insured, age), \(\mathbf{z}\) is a \(q\)-dimensional binary vector representation of a high-cardinality categorical variable (e.g.~occupation), $\beta_0 \in \mathbb{R}$, $\boldsymbol{\beta} = [\beta_1, \beta_2, \cdots, \beta_p]^\top \in \mathbb{R}^p$, and \(\boldsymbol{\alpha} = [\alpha_1, \alpha_2, \cdots, \alpha_q]^\top \in \mathbb{R}^q\) with \(\sum_{i=1}^q \alpha_i = 0\) are the regression coefficients to be estimated. Note that the additional constraint \(\sum_{i=1}^q \alpha_i = 0\) is required to restore identifiability of parameters.

In high-cardinality settings, the vector \(\mathbf{z}\) becomes high-dimensional, and hence so must be \(\boldsymbol{\alpha}\). High cardinality therefore introduces a large number of parameters to the estimation problem. Moreover, in most insurance problems, the distribution of categories will be highly imbalanced, meaning that some categories will have much more exposure (i.e.~data) than others. The rarely observed categories are likely to produce highly variable parameter estimates. As established in earlier sections, all this makes it extremely difficult to estimate the \(\boldsymbol{\alpha}\) accurately \citep{gelman2007}.

Equation \eqref{eq:nopool} is also referred to as the \textit{no pooling} model by \citet{antonio2014}, as each $\alpha_i$, $i = 1, 2, \cdots, q$ has to be learned independently. At the other end of the spectrum, there is a \textit{complete pooling} model, which simply ignores the \(\mathbf{z}\) and assumes
\begin{align}
\mu(\mathbf{x}, \mathbf{z}) = \beta_0 + \mathbf{x}^\top \boldsymbol{\beta}. \label{eq:pool}
\end{align}
In the middle ground between the overly noisy estimates produced by \eqref{eq:nopool} and the over-simplistic estimates from \eqref{eq:pool}, we can find the (linear) mixed models, which assume
\begin{align}
\mu(\mathbf{x}, \mathbf{z}) = \beta_0 + \mathbf{x}^\top \boldsymbol{\beta} + \mathbf{z}^\top \mathbf{u}, \label{eq:LMM}
\end{align}
where \(\mathbf{u} = [u_1, \cdots, u_q]^\top\) with \(u_i \stackrel{iid}{\sim} \mathcal{N}(0, \sigma_u^2)\) are known as the \emph{random effects} characterising the deviation of individual categories from the population mean, in contrast with the \emph{fixed effects} \(\beta_0\) and \(\boldsymbol{\beta}\). Model \eqref{eq:LMM} is also known as a \emph{random intercept} model.

The central idea in \eqref{eq:LMM} is that instead of assuming some fixed coefficient for each category, we assume that the effects of individual categories come from a distribution, so we only need to estimate (far fewer) parameters that govern the distribution of random effects rather than learning an independent parameter per category. This produces the equivalent effect of partial pooling \citep{gelman2007}: categories with a smaller number of observations will have weaker influences on parameter estimates, and extreme values get regularised towards the collective mean (modelled by the fixed effects, i.e.~\(\beta_0 + \mathbf{x}^\top \boldsymbol{\beta}\)).

In the same way that linear mixed models extend linear models, GLMMs extend GLMs by adding random effects capturing between-category variation to complement the fixed effects in the linear predictor.

Suppose that we have \(q\) categories in the data, each with \(n_i~(i = 1,2, \cdots, q)\) observations (so the total number of observations is \(n = \sum_{i=1}^q n_i\)). A GLMM is then defined by four components:

\begin{enumerate}
\def\labelenumi{(\arabic{enumi})}
\item
  \emph{Unconditional distribution of random effects}. We assume that the random effects \(u_j \stackrel{iid}{\sim} f(u_j|\boldsymbol{\gamma}), ~j= 1, 2, \cdots, q\) depend on a small set of parameters $\boldsymbol{\gamma}$. It is typically assumed that $u_j$ follows a Gaussian distribution with zero mean and variance $\sigma_u^2$, i.e.~\(u_j \stackrel{iid}{\sim} \mathcal{N}(0, \sigma_u^2)\).
\item
  \emph{Response distribution}. GLMMs assume that the responses \(Y_i, i = 1, 2, \cdots, n\), given the random effect \(u_{j[i]}\) for the category it belongs to (where $j[i]$ means that the $i$-th observation belongs to the $j$-th category, $j = 1, 2, \cdots, q$), are conditionally independent with an ED distribution, i.e.
  \begin{align}
  p\left(y_{i} | u_{j[i]}, \theta_{i}, \phi \right)= \exp \left[\frac{y_{i} \theta_{i}- b\left(\theta_{i}\right)}{\phi}+c\left(y_{i}, \phi\right)\right], \label{eq:EDF}
  \end{align}
  where \(\theta_{i}\) denotes the location parameter and \(\phi\) the dispersion parameter for the ED density, \(b(\cdot)\) and \(c(\cdot)\) are known functions. It follows from the properties of ED family distributions that
  \begin{align}
  \mu_{i} & = \mathbb{E}(Y_{i} | u_{j[i]}) = b'(\theta_{i}), \label{eq:mean} \\
  \operatorname{Var}(Y_{i} | u_{j[i]}) &= \phi b''(\theta_{i}) = \phi V(\mu_{i}), \label{eq:var}
  \end{align}
  where \(b'(\cdot)\) and \(b''(\cdot)\) denote the first and second derivatives of \(b(\cdot)\) and \(V(\cdot)\) is commonly referred to as the variance function. Equation \eqref{eq:mean} implies that the conditional distribution of \(Y_{i} | u_{j[i]}\) in \eqref{eq:EDF} is completely specified by the conditional mean \(\mu_{i} = \mathbb{E}(Y_{i} | u_{j[i]})\) and the dispersion parameter \(\phi\).
\item
  \emph{Linear predictor}. The linear predictor includes both fixed and random effects:
  \begin{align}
  \eta_{i} = \beta_0 + \mathbf{x}_{i}^\top \boldsymbol{\beta} + \mathbf{z}_{i}^\top \mathbf{u}, ~~ i = 1, \cdots, n, \label{eq:eta}
  \end{align}
  where \(\mathbf{x}_{i}\) denotes a vector of fixed effects variables (e.g.~sum insured, age) and \(\mathbf{z}_{i}\) a vector of random effects variables (e.g.~a binary representation of a high-cardinality feature such as injury code).
\item
  \emph{Link function}. Finally, a monotone differentiable function \(g(\cdot)\) links the conditional mean \(\mu_{ij}\) with the linear predictor \(\eta_{ij}\):
  \begin{align}
  g(\mu_{i}) =g(\mathbb{E}[Y_{i}|u_{j[i]}]) = \eta_{i}, ~~ i = 1, \cdots, n, ~ j = 1, \cdots, q, \label{eq:link}
  \end{align}
  completing the model.
\end{enumerate}
As explained earlier, the addition of random effects allows GLMMs to account for correlation within the same category, without overfitting to an individual category as is the case of a one-hot encoded GLM. The shared parameters that govern the distribution of random effects essentially enable information to transfer between categories, which is helpful for learning.


\subsection{Model Architecture} \label{ssec:glmmnet}
In Section \ref{ssec:background}, we briefly reviewed previous attempts at embedding ML into mixed models. In particular, the LMMNN structure proposed by \citet{simchoni2022} is the closest to what we envision for a mixed effects neural network. However, it suffers from two major shortcomings that restrict its applicability to insurance contexts: First, the LMMNN assumes a Gaussian response with an identity link function, for which analytical results can be obtained, but which is ill suited to modelling skewed, heavier-tailed distributions that dominate insurance and financial data. Second, LMMNNs model the random effects in a non-linear way (through a sub-network in the structure). This complicates the interpretation of random effects, which carries practical significance in many applications.

\begin{figure}[H]
\begin{center}
\includegraphics[width=0.65\linewidth]{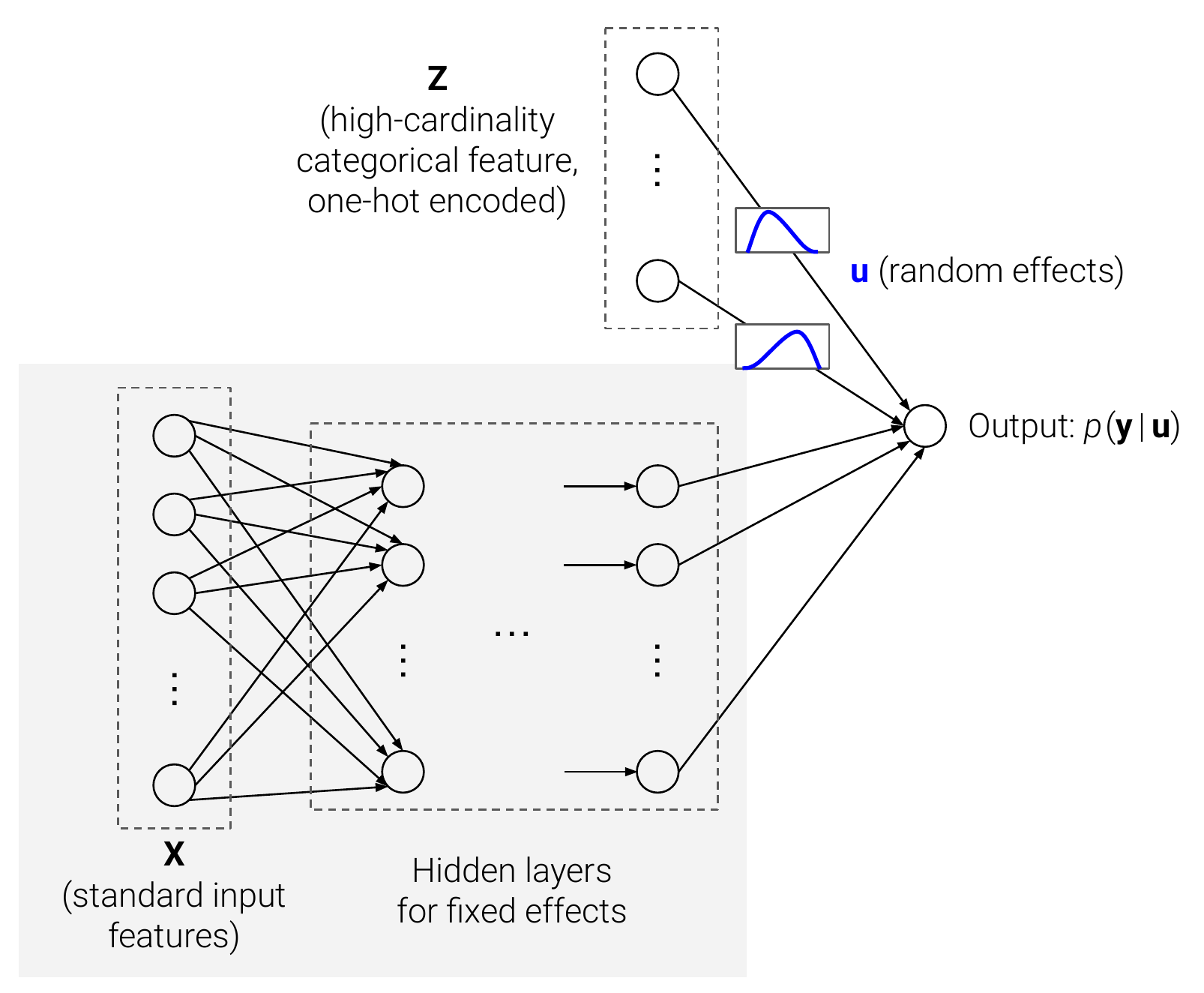}
\end{center}
\caption{Architecture of GLMMNet}\label{fig:glmmnet}
\end{figure}

The GLMMNet aims to address these concerns. The architectural skeleton of the model is depicted in Figure~\ref{fig:glmmnet}. We adopt a very similar structure to that of \citet{simchoni2022}, except that we remove the sub-network that they used to learn a non-linear function of \(\mathbf{Z}\). As noted earlier, the main purpose of our modification is to retain the interpretability of random effect predictions from the GLMM's linear predictor. In addition, we also want to avoid an over-complicated structure for the random effects, whose role is to act as a regulariser \citep{gelman2007}.

In mathematical terms, we assume that the conditional \(\mathbf{y}|\mathbf{u}\) follows an ED family distribution, with
\begin{align*}
g(\boldsymbol{\mu})=g(\mathbb{E}[\mathbf{y}| \mathbf{u}])=f(\mathbf{X})+\mathbf{Z}\mathbf{u}, \quad \mathbf{u} \sim \mathcal{N}(\mathbf{0}, \boldsymbol{\Sigma}),
\end{align*}
where \(g(\cdot)\) is the link function and \(f(\cdot)\) is learned through a neural network.

In the following, we give a detailed description of each component in the architecture plotted above, followed by a discussion on how the collective network can be trained in Section \ref{ssec:glmmnet-training}. We remark that, in this work we have allowed \(f(\cdot)\) to be fully flexible to take full advantage of the predictive power of neural networks. We recognise that in some applications, interpretability of the fixed effects may also be desirable. In such cases, it is possible to replace the fixed effects module (described in Section~\ref{sssec:fe}) by a Combined Actuarial Neural Network \citep[CANN;][]{wuethrich2019} structure.

\subsubsection{Fixed Effects} \label{sssec:fe}
The biggest limitation of GLMMs lies in their linear structure in \eqref{eq:eta}: similar to GLMs, features need to be hand-crafted to allow for non-linearities and interactions \citep{richman2021b, richman2021a}. The proposed GLMMNet addresses this issue by utilising a multi-layer network component for the fixed effects, which is represented as the shaded structure in Figure \ref{fig:glmmnet}. For simplicity here we consider a fully connected feedforward neural network (FFNN), although many other network structures, e.g. convolutional neural networks (CNN) can also be easily accommodated.

A FFNN consists of multiple layers of neurons (represented by circles in the diagram) with non-linear activation functions to capture potentially complex relationships between input and output vectors; we refer to \citet{goodfellow2016} for more details. Formally, for a network with \(L\) hidden layers and \(q_l\) neurons in the \(l\)-th layer for \(l=1,\cdots,L\), the \(l\)-th layer is defined in Equation \eqref{eq:hidden-layer} below.
\begin{align}
\mathbf a^{(l)}: \mathbb{R}^{q_{l-1}} & \to \mathbb{R}^{q_l}, \notag \\
\mathbf v & \mapsto \mathbf a^{(l)}(\mathbf v) = \left[a_1^{(l)}(\mathbf v), \cdots, a_{q_l}^{(l)}(\mathbf v)\right]^\top, \label{eq:hidden-layer}
\end{align}
with
\begin{align}
a_j^{(l)}(\mathbf v) = \varphi^{(l)} \left(b_j^{(l)} + \mathbf w_j^{(l)\top} \mathbf v\right), \quad j = 1, \cdots, q_l, \label{eq:hidden-activ}
\end{align}
where \(\varphi^{(l)}: \mathbb{R} \to \mathbb{R}\) is called the activation function for the \(l\)-th layer, \(b_j^{(l)} \in \mathbb{R}\) and \(\mathbf w_j^{(l)} \in \mathbb{R}^{q_{l-1}}\) respectively represent the bias term (or intercept in statistical terminology) and the network weights (or regression coefficients) for the \(j\)-th neuron in the \(l\)-th layer, and \(q_0\) is defined as the number of (preprocessed) covariates entering the network.

The network therefore provides a mapping from the (preprocessed) covariates \(\mathbf x\) to the desired output layer through a composition of hidden layers:
\begin{align}
f: \mathbb{R}^{q_0} &\to \mathbb{R}^{q_{L+1}} \notag \\
\mathbf x &\mapsto f(\mathbf x) = \left(\mathbf a^{(L+1)} \circ \cdots \circ \mathbf a^{(1)} \right) (\mathbf x). \label{eq:NN-activ}
\end{align}

Training a network involves making many specific choices; Appendix~\ref{app:nn} gives more details.

\subsubsection{Random Effects} \label{sssec:re}
Below we give a description of the random effects component of GLMMNet, which is the top (unshaded) structure in Figure \ref{fig:glmmnet}. For illustration, we stay with the case of a single high-cardinality categorical feature with \(q\) unique levels. The column of the categorical feature is first one-hot transformed into a binary matrix \(\mathbf{Z}\) of size \(n \times q\), which forms the input to the random effects component of the GLMMNet.

Here comes the key difference from the fixed effects network: The weights in the random effects layer, denoted by \(\mathbf{u}=[u_1, u_2, \cdots, u_q]^\top\), are not fixed values but are assumed to come from a distribution. This way, instead of having to estimate the effects of every individual category---many of which may lack sufficient data to support a reliable estimate, we only need to estimate the far fewer parameters that govern the distribution of $\mathbf{u}$ (in our case, just a single parameter \(\sigma_u^2\) in \eqref{eq:prior}).

A less visible yet important distinction of our GLMMNet from the LMMNN \citep{simchoni2022} is that \emph{our model takes a Bayesian approach to the estimation of random effects (as opposed to an exact likelihood approach in the LMMNN)}. The Bayesian approach helps sidestep some difficult numerical approximations of the (marginal) likelihood function (as detailed in Section~\ref{ssec:glmmnet-training}). We should point out that, although the Bayesian approach is popular for estimating GLMMs \citep[e.g.][]{brms2017}, there are further computational challenges that prevent them from being applied to ML mixed models; we elaborate on this in Section \ref{ssec:glmmnet-training}.

Following the literature \citep{mcculloch2001, antonio2007}, we assume that the random effects follow a normal distribution with zero mean, and that the random effects are independent across categories:
\begin{align}
\mathbf{u} \sim \mathcal{N}(\mathbf{0}, \sigma_u^2 \mathbf{I}), \label{eq:prior}
\end{align}
or equivalently, \(u_i \stackrel{iid}{\sim} \mathcal{N}(0, \sigma_u^2)\) for \(i = 1, 2, \cdots, q\). This is taken as the prior distribution of the random effects \(\mathbf{u}\).

In the context of our research problem, we are interested in the predictions for \(\mathbf{u}\), which indicate the deviation of individual categories from the population mean, or in other words, the excess risk they carry. Under the Bayesian framework, \emph{given the posterior distribution \(p(\mathbf{u} | \mathcal{D})\)}, we can simply take the posterior mode \(\hat{\mathbf{u}} = \mathop{\mathrm{argmax}}_\mathbf{u} p(\mathbf{u} | \mathcal{D})\) as a point estimate, which is also referred to as the maximum a posteriori (MAP) estimate. Alternatively, we can also derive interval predictions from the posterior distribution to determine whether the deviations as captured by \(\mathbf{u}\) carry statistical significance.

We should note that the estimation of the posterior distribution, which we treated as given in the above, represents a major challenge in Bayesian inference. We come back to this topic and discuss how we tackle it in Section~\ref{ssec:glmmnet-training}.

It should also be pointed out that the extension to multiple categorical features is very straightforward: all we need is to add another random effects layer (and one more variance parameter to estimate for each additional feature).

\subsubsection{Response Distribution} \label{sssec:rd}
As explained earlier, one novel contribution of GLMMNet is its ability to accommodate many non-Gaussian distributions, notably gamma (commonly used for claim severity), (over-dispersed) Poisson (for claim counts), and Bernoulli (for binary classification tasks). Our extension to non-Gaussian distributions is an important pre-requisite for insurance applications.

Precisely, we assume that the responses \(Y_i, i = 1, 2, \cdots, n\), given the random effect \(u_{j[i]}\) for the category it belongs to (where $j[i]$ means that the $i$-th observation belongs to the $j$-th category), are conditionally independent with a distribution in the ED family. Equation \eqref{eq:mean} implies that the conditional distribution of \(Y_{i} | u_{j[i]}\) is completely specified by the conditional mean \(\mu_{i}\) and the dispersion parameter \(\phi\), so we can write\footnote{The assumption of constant dispersion parameter $\phi$ is made to comply with the standard assumptions of GLM/GLMMs, but can be relaxed if desired by a slight modification. For the purpose of illustration in this work, we do not pursue this path down further but note that it can be an interesting direction for future research.} \[y_i | u_{j[i]} \sim \mathrm{ED}({\mu}_i, \phi).\]

In the GLMMNet, we assume that a link function \(g(\cdot)\) connects the conditional mean \(\mu_{i}\) with the non-linear predictor \(\eta_{i}\) formed by adding together the output from the fixed effects module and the random effects layer:
\begin{align}
g(\mu_{i}) = \eta_{i} = f(\mathbf{x}_i) + \mathbf{z}_i^\top \mathbf{u} = f(\mathbf{x}_i) + u_{j[i]}, ~~ i = 1, \cdots, n, ~ j=1, \cdots, q. \label{eq:glmmnet-predictor}
\end{align}
In implementation, we use the inverse link \(g^{-1}(\cdot)\) as the activation function for the final layer, to map the output \([f(\mathbf{x}_i), u_{j[i]}]\) from the fixed effects module and the random effects layer to a prediction for the conditional mean response \(\mu_i = g^{-1}(f(\mathbf{x}_i) + u_{j[i]}).\)

Finally, as GLMMNet operates under probabilistic assumptions, the dispersion parameter \(\phi\) (independent of covariates in our setting) can be estimated under maximum likelihood as part of the network. This is implemented by adding an input-independent but trainable weight to the final layer of GLMMNet \citep{keras}, whose value will be updated by (stochastic) gradient descent on the loss function.

\subsection{Training of GLMMNet: Variational Inference} \label{ssec:glmmnet-training}
In a LMMNN \citep{simchoni2022}, the network is trained by minimising the negative log-likelihood of \(\mathbf{y}\). As the LMMNN assumes a Gaussian density for \(\mathbf{y} | \mathbf{u}\) and \(\mathbf{u} \sim \mathcal{N}(\mathbf{0}, \boldsymbol{\Sigma})\), the marginal likelihood of \(\mathbf{y}\) can be derived analytically, i.e.
\(\mathbf{y} \sim \mathcal{N}(f(\mathbf X), \mathbf Z \boldsymbol\Sigma \mathbf Z^\top + \sigma_\epsilon^2 \mathbf I)\).
The network weights and biases, as well as the variance and covariance parameters \(\sigma_\epsilon^2\) and \(\sigma_u^2\) are learned by minimising the negative log-likelihood
\begin{align*}
-\log \mathcal{L}= \frac{1}{2}(\mathbf{y}-f(\mathbf{X}))^\top \mathbf{V}^{-1}(\mathbf{y}-f(\mathbf{X})) + \frac{1}{2} \log \operatorname{det}(\mathbf{V}) + \frac{n}{2} \log (2 \pi),
\end{align*}
where \(\mathbf{V} = \mathbf Z \boldsymbol\Sigma \mathbf Z^\top + \sigma_\epsilon^2 \mathbf I\) and \(\boldsymbol \Sigma := \operatorname{Cov}(\mathbf u) = \sigma_u^2 \mathbf{I}\).

In making the extension to GLMMNet, however, we no longer obtain a closed-form expression for the marginal likelihood:
\begin{align}
\mathcal{L}(\mathbf{y}; \boldsymbol{\beta}, \sigma_u, \phi) = \prod_{i=1}^n p(y_i | \boldsymbol{\beta}, \sigma_u, \phi) = \prod_{i=1}^n \int p(y_i | \boldsymbol{\beta}, u_{j[i]}, \phi) f(u_{j[i]} | \sigma_u) \mathrm{d} u_{j[i]} \label{eq:glmm-likelihood}
\end{align}
where \(\boldsymbol{\beta}\) denote the weights and biases in the fixed effects component of the GLMMNet (Section \ref{sssec:fe}), \(j[i]\) indicates the category to which the \(i\)-th observation belongs, \(\sigma_u^2\) is the variance parameter of the random effects, and \(\phi\) is the usual dispersion parameter for the ED density for the response.

We take a Bayesian approach to circumvent the difficult numerical approximations required to compute the integral in Equation \eqref{eq:glmm-likelihood}. Under the Bayesian framework, \(\pi(u_j) = f(u_j | \sigma_u) = \mathcal{N}(0, \sigma_u^2), ~j = 1, \cdots, q\), is taken as the prior for the random effects. Our goal is to make inference on the posterior of the random effects, given by
\begin{align}
p(\mathbf{u} | \mathcal{D})=\frac{\pi(\mathbf{u})p(\mathcal{D}|\mathbf{u})}{p(\mathcal{D})}=\frac{\pi(\mathbf{u})p(\mathcal{D}|\mathbf{u})}{\int \pi(\mathbf{u} )p(\mathcal{D}|\mathbf{u} ) \mathrm{d} \mathbf{u}}, \label{eq:posterior}
\end{align}
where \(\pi(\mathbf{u}) = \prod_{j=1}^q \pi(u_j)\) and $\mathcal{D}$ represents the data. Note that we again encounter an intractable integral in \eqref{eq:posterior}. The traditional solution is to use Markov chain Monte Carlo (MCMC) to sample from the posterior without computing its exact form; however, the computational burden of MCMC techniques restricts its applicability to large and complex models like a deep neural network. This partly explains why Bayesian methods, although frequently used for estimating GLMMs \citep[e.g.][]{brms2017}, tend not to be as widely adopted among ML mixed models \citep[e.g.][]{sigrist2022,hajjem2014}.

With our GLMMNet, we apply \emph{variational inference} to solve this problem. Variational inference is a popular approach among ML researchers working with Bayesian neural networks, due to its computational efficiency \citep{blei2017,zhang2019}. Variational inference does not try to directly estimate the posterior, but proposes a surrogate parametric distribution \(q \in \mathcal{Q}\) to approximate it, therefore turning the difficult estimation into a highly efficient optimisation problem. In practice, the choice of the surrogate distribution is often made based on simplicity or computational feasibility. One particularly popular option is to use a mean-field distribution family \citep{blei2017}, which assumes independence among all latent variables. In this work, we consider a diagonal Gaussian distribution for the surrogate posterior, \[q_{\boldsymbol{\vartheta}}(\mathbf{u})=\mathcal{N}({\boldsymbol{\mu}}_u, {\boldsymbol{\Sigma}}_u),\] where \(\boldsymbol{\mu}_u\) is a \(q\)-dimensional vector of the surrogate posterior mean of the \(q\) random effects, and \(\boldsymbol{\Sigma}_u = \mathrm{diag}(\sigma_1^2, \sigma_2^2, \cdots, \sigma_q^2)\) is a diagonal matrix of the posterior variance. More complex distributions, such as those that incorporate dependencies among latent variables, may provide a more accurate approximation. In this work, however, as we are mainly concerned with estimating the mean and dispersion of the posterior random effects, the chosen diagonal Gaussian surrogate distribution should suffice for this purpose.

The vector \(\boldsymbol{\mu}_u\) and the diagonal elements of \(\boldsymbol{\Sigma}_u\) together form the \emph{variational parameters} of the diagonal Gaussian, denoted by \(\boldsymbol{\vartheta}\). The task here is to find the variational parameters \(\boldsymbol{\vartheta}^*\) that make the approximating density \(q_{\boldsymbol{\vartheta}} \in \mathcal{Q}\) closest to the true posterior, where closeness is defined in the sense of minimising the Kullback-Leibler (KL) divergence \citep{kullback1951}. In mathematical terms, this means
\begin{align}
q_{\boldsymbol{\vartheta}^{*}}(\mathbf{u}) &= \mathop{\mathrm{argmin}}_{q_{\boldsymbol{\vartheta}}(\mathbf{u}) \in \mathcal{Q}} \mathrm{KL}[q_{\boldsymbol{\vartheta}}(\mathbf{u}) \| p(\mathbf{u} | \mathcal{D})] \label{eq:glmm-vi} \\
&= \mathop{\mathrm{argmin}}_{q_{\boldsymbol{\vartheta}}(\mathbf{u}) \in \mathcal{Q}} \mathbb{E}_{q_{\boldsymbol{\vartheta}}(\mathbf{u})} [\log q_{\boldsymbol{\vartheta}}(\mathbf{u}) - \log p(\mathbf{u} | \mathcal{D})] \\
&= \mathop{\mathrm{argmin}}_{q_{\boldsymbol{\vartheta}}(\mathbf{u}) \in \mathcal{Q}} \mathbb{E}_{q_{\boldsymbol{\vartheta}}(\mathbf{u})} [\log q_{\boldsymbol{\vartheta}}(\mathbf{u}) - \log \pi(\mathbf{u}) - \log p(\mathcal{D} | \mathbf{u})] \\
&= \mathop{\mathrm{argmin}}_{q_{\boldsymbol{\vartheta}}(\mathbf{u}) \in \mathcal{Q}} \left(\operatorname{KL}[q_{\boldsymbol{\vartheta}}(\mathbf{u}) \| \pi(\mathbf{u})]-\mathbb{E}_{q_{\boldsymbol{\vartheta}}(\mathbf{u})}[\log p(\mathcal{D} | \mathbf{u})]\right).
\end{align}

We take the term inside the outermost bracket, which is called the \emph{evidence lower bound} (ELBO) loss or the negative of \emph{variational free energy} in \citet{neal1998}, to be the loss function for the GLMMNet:
\begin{align}
\mathcal{L}_\mathrm{GLMMNet} &= \operatorname{KL}[q_{\boldsymbol{\vartheta}}(\mathbf{u}) \| \pi(\mathbf{u})]-\mathbb{E}_{q_{\boldsymbol{\vartheta}}(\mathbf{u})}[\log p(\mathcal{D} | \mathbf{u})] \label{eq:glmm-ELBO} \\
&= \mathbb{E}_{q_{\boldsymbol{\vartheta}}(\mathbf{u})}\left[\log \frac{q_{\boldsymbol{\vartheta}}(\mathbf{u})}{\pi(\mathbf{u})} \right]- \mathbb{E}_{q_{\boldsymbol{\vartheta}}(\mathbf{u})}[\log p(\mathcal{D} | \mathbf{u})]. \label{eq:glmm-ELBO-E}
\end{align}
We see that the loss function used to optimise the GLMMNet is a sum of two parts: a first part that captures the divergence between the (surrogate) posterior and the prior, and a second part that captures the likelihood of observing the given data under the posterior. This decomposition echoes the trade-off between the prior beliefs and the data in any Bayesian analysis \citep{blei2017}.

The loss function in \eqref{eq:glmm-ELBO-E} can be calculated via Monte Carlo, that is, by generating posterior samples under the current estimates of the variational parameters $\boldsymbol{\vartheta}$, evaluating the respective expressions at each of the sampled values, and taking the average to approximate the expectation. The derivatives of the loss function can be computed via automatic differentiation \citep{keras} and used by standard gradient descent algorithms for optimisation \citep[e.g.][]{kingma2014}.

\subsection{Prediction from GLMMNet} \label{ssec:glmmnet-predict}
Recall that \(\mathbf{x}\) denotes the standard input features and \(\mathbf{z} \in \mathbb{R}^q\) is a binary representation of the high-cardinality feature. Let \((\mathbf{x}^*, \mathbf{z}^*)\) represent a new data observation for which a prediction should be made. Note that when making predictions, it is possible to encounter a category that was never seen during training, in which case \(\mathbf{z}^*\) will be a zero vector (as it does not belong to any of the already-seen categories) and the predictor in Equation \eqref{eq:glmmnet-predictor} reduces to \(\eta^* = f(\mathbf{x}^*)\).

To make predictions from GLMMNet, we take expectations under the posterior distribution on the random effects \citep{blundell2015,jospin2022}:
\begin{align}
p(y^* | \mathbf{x}^*, \mathbf{z}^*) &= \mathbb{E}_{p(\mathbf{u} | \mathcal{D})} \left[p(y^* | \mathbf{x}^*, \mathbf{z}^*, \mathbf{u}) \right] \notag \\
& \approx \mathbb{E}_{q_{\boldsymbol{\vartheta}^{*}}(\mathbf{u})} \left[p(y^* | \mathbf{x}^*, \mathbf{z}^*, \mathbf{u}) \right] \label{eq:glmmnet-pred}
\end{align}
where \(q_{\boldsymbol{\vartheta}^{*}}(\cdot)\) denotes the optimised surrogate posterior. As a result of Equation \eqref{eq:glmmnet-pred}, any functional of \(y^* | \mathbf{x}^*, \mathbf{z}^*\), e.g.~the expectation, can be computed in an analogous way.

The expectation in \eqref{eq:glmmnet-pred} can be approximated by drawing Monte Carlo samples from the (surrogate) posterior, as outlined in Algorithm \ref{alg:GLMMNet-predict}. When there are multiple high-cardinality features, each will have its own binary vector representation and optimised surrogate posterior, and the expectation in Equation~\eqref{eq:glmmnet-pred} will be taken with respect to the joint posterior (which is simply a product of the surrogate posteriors when we assume independence between these features).

It is also worth pointing out that the model averaging in line \ref{line:ensemble} has the equivalent effect of training an ensemble of networks, which is usually otherwise computationally impractical \citep{blundell2015}.
\begin{algorithm}[htbp!]
\SetKwInOut{Input}{Input}
\SetKwInOut{Output}{Output}
\caption{Prediction from GLMMNet}\label{alg:GLMMNet-predict}
\Input{A new data observation $(\mathbf{x}^*, \mathbf{z}^*)$; number of Monte Carlo iterations $N$}
\Output{Predictive distribution: $p(y^* | \mathbf{x}^*, \mathbf{z}^*)$}
\For{$i = 1$ to $N$}{
  Simulate $\mathbf{u}_{(i)}$ from the surrogate posterior $q_{\boldsymbol{\vartheta}^{*}}(\mathbf{u})$\;
  Set $p_{(i)} = p(y^* | \mathbf{x}^*, \mathbf{z}^*, \mathbf{u}_{(i)}) = \mathrm{ED}(\mu_{(i)}, \phi)$ where $\mu_{(i)} = g^{-1}\left(f(\mathbf{x}^*) + \mathbf{z}^{*\top}\mathbf{u}_{(i)}\right)$ and $\phi$ is the dispersion parameter for the ED distribution (learned in the GLMMNet)\;
}
\Return $p(y^* | \mathbf{x}^*, \mathbf{z}^*) = \frac{1}{N} \sum_{i=1}^N p_{(i)}$\; \label{line:ensemble}
\end{algorithm}

\section{Comparison of GLMMNet with Other Leading Models} \label{sec:simulation}

In this section, we compare the performance of the GLMMNet against the most popular existing alternatives for handling high-cardinality categorical features. We list the candidate models in Section~\ref{ssec:candidates} and the performance metrics we use to evaluate the models in Section~\ref{ssec:metrics}. Control over the true data generating process via simulation allows us to highlight the respective strengths of the models. We challenge the models under environments of varying levels of complexity, e.g.~low signal-to-noise ratio, highly imbalanced distribution of categories, and/or skewed response distributions. The simulation environments detailed in Section~\ref{ssec:sim-env} are designed to mimic these typical characteristics of insurance data. Section~\ref{ssec:sim-results} summarises the results. Ensuing insights are used to inform the real data case study in Section~\ref{sec:case-study}.

\subsection{Models for Comparison} \label{ssec:candidates}
Listed below are the leading models widely used in practice that allow for high-cardinality categorical features. They will be implemented and used for comparison with the GLMMNet in this experiment. 
\begin{arcenum}
\item
  GLM with 
  \begin{arcitem}
  \item \texttt{GLM\_ignore\_cat}: complete pooling, i.e.~ignoring the categorical feature altogether. In the notation of Section~\ref{ssec:problem}, it can be represented as $\mu(\mathbf{x}, \mathbf{z}) = g^{-1}(\beta_0 + \mathbf{x}^\top \boldsymbol{\beta})$, where $g(\cdot)$ is the link function, $\beta_0$ and $\boldsymbol{\beta}$ are the regression parameters to be estimated;
  \item \texttt{GLM\_one\_hot}: one-hot encoding, which can be represented as $\mu(\mathbf{x}, \mathbf{z}) = g^{-1}(\beta_0 + \mathbf{x}^\top \boldsymbol{\beta} + \mathbf{z}^\top \boldsymbol{\alpha})$ where $\boldsymbol{\alpha}$ are the additional parameters for each category in $\mathbf{z}$;
  \item \texttt{GLM\_GLMM\_enc}: GLMM encoding, which can be represented as $\mu(\mathbf{x}, \mathbf{z}) = g^{-1}(\beta_0 + \mathbf{x}^\top \boldsymbol{\beta} + z'\alpha)$ where $z'$ represents the encoded value (a scalar) of $\mathbf{z}$ and $\alpha$ the additional parameter associated with it (more details in Appendix~\ref{app:glmm-enc});
\end{arcitem}
\item
  Gradient boosting machine (GBM, with trees as base learners) under the same categorical encoding setup as above:
  \begin{arcitem}
      \item \texttt{GBM\_ignore\_cat}: $\mu(\mathbf{x}, \mathbf{z}) = \mu(\mathbf{x})$ where $\mu(\cdot)$ is a weighted sum of tree learners;
      \item \texttt{GBM\_one\_hot}: $\mu(\mathbf{x}, \mathbf{z})$ where $\mu(\cdot, \cdot)$ is a weighted sum of tree learners;
      \item \texttt{GBM\_GLMM\_enc}: $\mu(\mathbf{x}, \mathbf{z}) = \mu(\mathbf{x}, z')$ where $z'$ represents the encoded value (a scalar) of $\mathbf{z}$;
  \end{arcitem}
\item
  Neural network with entity embeddings (\texttt{NN\_ee}). Here, $\mu(\mathbf{x}, \mathbf{z})$ is composed of multiple layers of interconnected artificial neurons, where each neuron receives inputs, performs a weighted sum of these inputs, and applies an activation function to produce its output (see also Section~\ref{sssec:fe}). Entity embeddings add a linear layer between each one-hot encoded input and the first hidden layer and are learned as part of the training process. This is currently the most popular approach for handling categorical inputs in deep learning models and serves as our \textbf{target benchmark};
\item
  GBM with pre-learned entity embeddings (\texttt{GBM\_ee}), which is similar to \texttt{GBM\_GLMM\_enc} except with the $z'$ replaced by a vector of entity embeddings learned from \texttt{NN\_ee};
\item
  Mixed models:
  \begin{arcitem}
  \item
    \texttt{GLMM} with $\mu(\mathbf{x}, \mathbf{z}) = g^{-1}(\beta_0 + \mathbf{x}^\top \boldsymbol{\beta} + \mathbf{z}^\top \mathbf{u})$ where $\mathbf{u}$ are the random effects (Section~\ref{ssec:glmm});
  \item
    \texttt{GPBoost} \citep{sigrist2021,sigrist2022} with $\mu(\mathbf{x}, \mathbf{z}) = g^{-1}(f(\mathbf{x}) + \mathbf{z}^\top \mathbf{u})$ where $f(\cdot)$ is an ensemble of tree learners;
  \item
    The proposed \texttt{GLMMNet}, with $\mu(\mathbf{x}, \mathbf{z}) = g^{-1}(f(\mathbf{x}) + \mathbf{z}^\top \mathbf{u})$ where $f(\cdot)$ is a FFNN (Section~\ref{sssec:fe}).
  \end{arcitem}
\end{arcenum}

Each model has its own features and properties, which renders the collection of models suitable for different contexts; Table~\ref{tab:candidates} gives a high-level summary of their different characteristics. In the table, ``interpretability" refers to the ease with which a human can understand a model, as well as the ability to derive insights from the model. For instance, GLMs have high interpretability, because GLMs have a small number of parameters and each parameter carries a physical meaning, which contrasts with the case of a standard deep neural network characterised by a large number of parameters that are meaningless on their own. ``Operational complexity" refers to the resources and time required to implement, train and fine-tune the model.

\begin{table}[ht]
\centering
\begin{footnotesize}
\begin{tabular}{p{1.4cm}p{2cm}p{3cm}p{3cm}p{2.3cm}p{1.5cm}}
\toprule
 & \multicolumn{3}{c}{\textbf{Predictive performance}} & \multicolumn{2}{c}{\textbf{Other considerations}} \\ \cmidrule(lr){2-4} \cmidrule(lr){5-6}
 & \textit{Handling non-linearities} & \textit{Categorical treatment} & \textit{Response distribution} & \textit{Interpretability} & \textit{Operational complexity} \\ \midrule
GLM & None (unless performed manually) & Optional categorical encoding methods can be used to preprocess features, e.g.~one-hot encoding & ED family & High (fully transparent model) & Low \\ \midrule
GBM & Good & Optional categorical encoding methods can be used to preprocess features, e.g.~one-hot encoding & Gaussian (alternatives can be accommodated by specifying a different loss function) & Medium (through partial dependence plots and feature importance analysis) & Moderate \\ \midrule
GLMM & None (unless performed manually) & Through random effects, which are an addition to GLMs (see Section~\ref{ssec:glmm}) & ED family & High (fully transparent model) & Low \\ \midrule
Entity embeddings (neural network) & Excellent & Through entity embeddings & Gaussian (alternatives can be accommodated by specifying a different loss function) & Low & High \\ \midrule
GPBoost & Good & Through random effects (see Section~\ref{ssec:candidates}) & Selected ED family distributions (Bernoulli, Poisson, gamma, Gaussian) & Medium (transparent random effects) & Moderate \\ \midrule
GLMMNet & Excellent & Through random effects (see Section~\ref{ssec:glmmnet}) & ED family & Medium (transparent random effects) & High \\ \bottomrule
\end{tabular}
\caption{Overview of different characteristics of candidate models} \label{tab:candidates}
\end{footnotesize}
\end{table}

Based on this qualitative comparison only, the GLMMNet has good potential to improve on neural networks with entity embeddings, because it can similarly capture complex relationships while also offering a wider range of response distributions and better interpretability. However, further investigations are needed to understand and illustrate in which circumstances the GLMMNet approach can outperform other approaches, which is the focus of the rest of this section.

\subsection{Model Evaluation Criteria} \label{ssec:metrics}
For all experiments presented in this paper (including the real data case study in Section~\ref{sec:case-study}), we randomly split the data into a training and a test set. The training set is used to select hyperparameters (by further splitting into an inner-training and a validation set) and fit the pool of candidate models. The different models are then evaluated and compared based on their performance on the test dataset. Specifically, to quantify the predictive performance, we consider the metrics listed below:
\begin{arcitem}
\item
  \emph{Accuracy of point predictions}. We report the root mean squared error (RMSE) and the mean absolute error (MAE), which are respectively given by
  \begin{align*}
  \mathrm{RMSE}=\sqrt{\frac{1}{n^*} \sum_{i=1}^{n^*}\left(y_i^*-\hat{y}_i^*\right)^2}, ~~ \mathrm{MAE}=\frac{1}{n^*} \sum_{i=1}^{n^*}\left|y_i^*-\hat{y}_i^*\right|,
  \end{align*} where \(n^*\) is the number of test observations, \(\mathbf{y}^*\) is the target output and \(\hat{\mathbf{y}}^*\) is the (point) prediction of the mean response.
\item
  \emph{Average accuracy of point predictions per category}. In order to gain insights into the category-specific accuracy of point predictions, we consider the RMSE of average prediction for each category: \[\mathrm{RMSE\_avg}=\sqrt{\frac{1}{q}\sum_{j=1}^q (\overline{y}_j^*-\overline{\widehat{y}}^*_j)^2}, \quad \overline{y}_j^*=\frac{1}{n_j^*}\sum_{i:j[i]=j}^{n^*}y_i^*, ~~ \overline{\widehat{y}}_j^*=\frac{1}{n_j^*}\sum_{i:j[i]=j}^{n^*}\widehat{y}_i^*,\] where \(q\) is the number of categories, \(n_j^*\) is the number of test observations in the \(j\)-th category, \(\overline{y}_j^*\) and \(\overline{\widehat{y}}_j^*\) respectively denote the average of the response variable \(y\) and its prediction \(\hat{y}\) for all observations in the \(j\)-th category. This metric serves to measure and compare how well a model is able to capture the between-category differences. This can be interpreted as, for example, the average accuracy of loss predictions on each sub-portfolio, which has practical significance.
\item
  \emph{Accuracy of probabilistic predictions}. The quantification of probabilistic accuracy is especially relevant to actuarial applications \citep{embrechts2022}. In this work, we consider two \emph{proper scoring rules}, the continuous ranked probability score (``CRPS'') and the negative log-likelihood (``NLL''), both of which are commonly used for assessing probabilistic forecasts \citep{gneiting2007,al-mudafer2022,delong2021a}. The CRPS is defined as \begin{equation}
  \mathrm{CRPS} = \frac{1}{n^*}\sum_{i=1}^{n^*} \mathrm{CRPS}(\hat{F}_i, y_i), \text{ where }
  \mathrm{CRPS}(\hat{F}_i, y_i) = \int_{-\infty}^{\infty}(\hat{F}_i(z)-\mathds{1}_{\{z \geq y_i\}})^{2} \mathrm{d} z, \label{eq:crps}
  \end{equation}
  and where \(\hat F_i(\cdot)\) represents the distribution function (df) of a probabilistic forecaster for the \(i\)-th response value and \(y_i\) represents its observed value. The CRPS is a quadratic measure of the difference between the forecast predictive df and the empirical df of the observation. Hence, the smaller it is the better. The integral in \eqref{eq:crps} has been evaluated analytically and implemented in R for many distributions; we refer to \citet{jordan2019} for details.

  Computation of the NLL is more straightforward: \begin{equation}
  \mathrm{NLL} = \frac{1}{n^*}\sum_{i=1}^{n^*} \mathrm{NLL}(\hat{F}_i, y_i), \text{ where }
  \operatorname{NLL}(\hat{F}_i, y_i) = -\log \hat{f}_i(y_i). \label{eq:nll}
  \end{equation} and where \(\hat{f}_i(\cdot)\) is the density of the probabilistic forecaster.

  For models that do not produce a distributional forecast, we construct an artificial predictive distribution by using the point forecast as the mean and estimating a dispersion parameter for the assumed distribution \citep[see Section 4.6 of][]{denuit2019glm}.
\end{arcitem}

\subsection{Simulation Datasets} \label{ssec:sim-env}
We generate \(n\) observations from a (generalised) non-linear mixed model with \(q\) categories (\(q \le n\)). We assume that the responses \(y_i, i = 1, \cdots, n\), each belonging to a category \(j[i] \in \{1, \cdots, q\}\), given the random effects \(\mathbf{u}=(u_j)_{j=1}^q\), are conditionally independent with a distribution in the ED family (e.g.~gamma); denoted as \(y_i | u_{j[i]} \sim \mathrm{ED}({\mu}_i, \phi),\) where \({\mu}_i=\mathbb{E}(y_i | u_{j[i]})\) represents the mean and \(\phi\) represents a dispersion parameter shared across all observations. A link function \(g(\cdot)\) connects the conditional mean \({\mu}_i\) with a non-linear predictor in the form of
\begin{align}
g(\mu_i)=g(\mathbb{E}[{y}_i| {u}_{j[i]}])=f(\mathbf{x}_i)+u_{j[i]}, \quad i = 1, \cdots, n, ~j = 1, \cdots, q, \label{eq:sim-formula}
\end{align}
where \(\mathbf{x}_i\) is a vector of fixed effects covariates for the \(i\)-th observation, \(\mathbf{z}_i\) is a vector of random effects variables (in our case, a binary vector representation of a high-cardinality categorical feature), the random effects \({u}_{j} \stackrel{iid}{\sim} \mathcal{N}(0, \sigma_u^2)\) measure the deviation of individual categories from the population mean, and \(f(\cdot)\) is some complex non-linear function of the fixed effects.

Here, we consider
\begin{align}
f(\mathbf{x}) &= 10 \sin \left(\pi x_{1} x_{2}\right)+20\left(x_{3}-0.5\right)^{2}+10 x_{4}+5 x_{5}, \label{eq:friedman-1}\\
x_i & \stackrel{iid}{\sim} \mathcal{U}(0,1), ~ i = 1, \cdots, 10, \notag
\end{align}
which was studied in \citet{friedman1991} and has become a standard simulation function for regression models (used in e.g. \citealt{kuss2005}; also available in the scikit-learn Python library by \citealt{sklearn}). In our application, we use the Friedman function to test the model's ability to filter out irrelevant predictors (note that \(x_6, \cdots, x_{10}\) is not used in \(f\)) as well as detect interactions and non-linearities of input features, all in the presence of a high-cardinality grouping variable that is subject to random effects.

We set up the desired simulation environments by adjusting the following parameters; refer to Appendix \ref{app:data-gen} for details.

\begin{arcitem}
\item
  \textit{Signal-to-noise ratio}: a three-dimensional vector that captures the relative ratio of signal strength (as measured by \(\mu_f\), the mean of \(f(\mathbf{x})\)), random effects variance (\(\sigma_u^2\)), and variability of the response (\(\sigma_\epsilon^2\); noise, or equivalently the irreducible error, as this component captures the unexplained inherent randomness in the response).
\item
  \textit{Response distribution}: distributional assumption for the response, e.g.~Gaussian, gamma, or any other member of the ED family.
\item
  \textit{Inverse link}: inverse of the link function, i.e.~the inverse function of \(g(\cdot)\) in Equation~\eqref{eq:sim-formula}.
\item
  \textit{Distribution of categories}: whether to use balanced or skewed distribution for the allocation of categories. A ``balanced'' distribution allocates approximately equal number of observations to each category; a ``skewed'' distribution generates categories from a (scaled) beta distribution; see Appendix~\ref{app:data-gen}.
\end{arcitem}

Table \ref{tab:scenarios} lists the simulation environments considered in our experiment. Each of the environments in experiments 2--5 has been configured to mimic one specific characteristic of practical insurance data, such as a low signal-to-noise ratio (the specification of $[8,1,4]$ was selected based on estimated parameters from the real insurance case study in Section \ref{sec:case-study}), an imbalanced distribution across categories, or a highly skewed response distribution for claim severity. We give a brief description of each environment below.

\begin{arcitem}
    \item \textit{Experiment 1} simulates the base scenario that adheres to the assumptions of a Gaussian GLMMNet.
    \item \textit{Experiment 2} simulates a gamma-distributed response, which is often used to model claim severity in lines of business (LoB) such as auto insurance or general liability.
    \item \textit{Experiment 3} simulates a skewed distribution of categories, which is a common characteristic of high-cardinality categorical features (e.g. car make, injury code).
    \item \textit{Experiments 4--5} incrementally increase the level of noise in the data and simulate LoBs that are difficult to model by covariates, such as commercial building insurance, catastrophic events and cyber risks.
    \item \textit{Experiment 6} represents the most challenging scenario, incorporating the complexities of all the other experiments.
\end{arcitem}

\begin{table}[ht]
{\centering
\begin{small}
\begin{tabular}{@{}cccccc@{}}
\toprule
Exp ID & \multicolumn{2}{c}{Signal-to-noise} & Response distribution & Inverse link & Distribution of categories \\ \midrule
1 (base) & $[4, 1, 1]$ & (high) & Gaussian & Identity & Balanced \\
2 & $[4, 1, 1]$ & (high) & \textbf{Gamma} & \textbf{Exponential} & Balanced \\
3 & $[4, 1, 1]$ & (high) & Gaussian & Identity & \textbf{Skewed} \\
4 & $\mathbf{[4, 1, 2]}$ & (medium) & Gaussian & Identity & Balanced \\
5 & $\mathbf{[8, 1, 4]}$ & (low) & Gaussian & Identity & Balanced \\
6 & $\mathbf{[8, 1, 4]}$ & (low) & \textbf{Gamma} & \textbf{Exponential} & \textbf{Skewed} \\ \bottomrule
\end{tabular}
\end{small}
\caption{Parameters used for five different simulation environments.\\Bold face indicates changes from the base scenario (i.e. experiment 1).} \label{tab:scenarios}
}
\end{table}

For each scenario, we generate 5,000 training observations and 2,500 testing observations.

\subsection{Results} \label{ssec:sim-results}
In all scenarios studied, the GLMMNet outperforms or at least performs on par with the target benchmark model of an entity embedded neural network. In low to medium noise level environments, the GLMMNet usually outperforms the second best model (i.e.~the entity embedded neural network) by a significant margin. The two outperform all the other mixed models, including GLMM and GPBoost.

\begin{figure}[ht]
{\centering \subfloat[Mean absolute error\label{fig:boxplots-1-1}]{\includegraphics[width=0.49\linewidth]{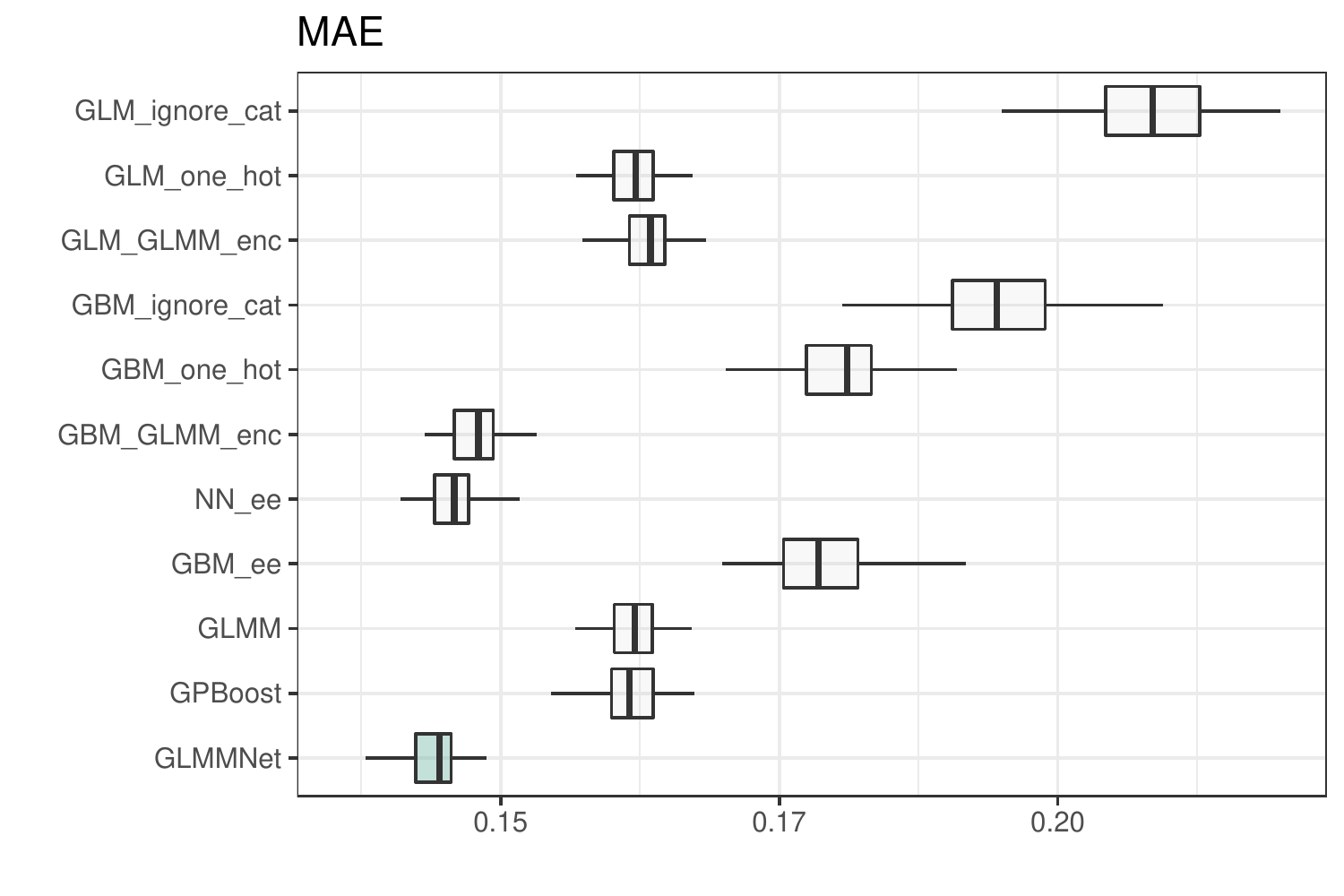} }\subfloat[Root mean square error\label{fig:boxplots-1-2}]{\includegraphics[width=0.49\linewidth]{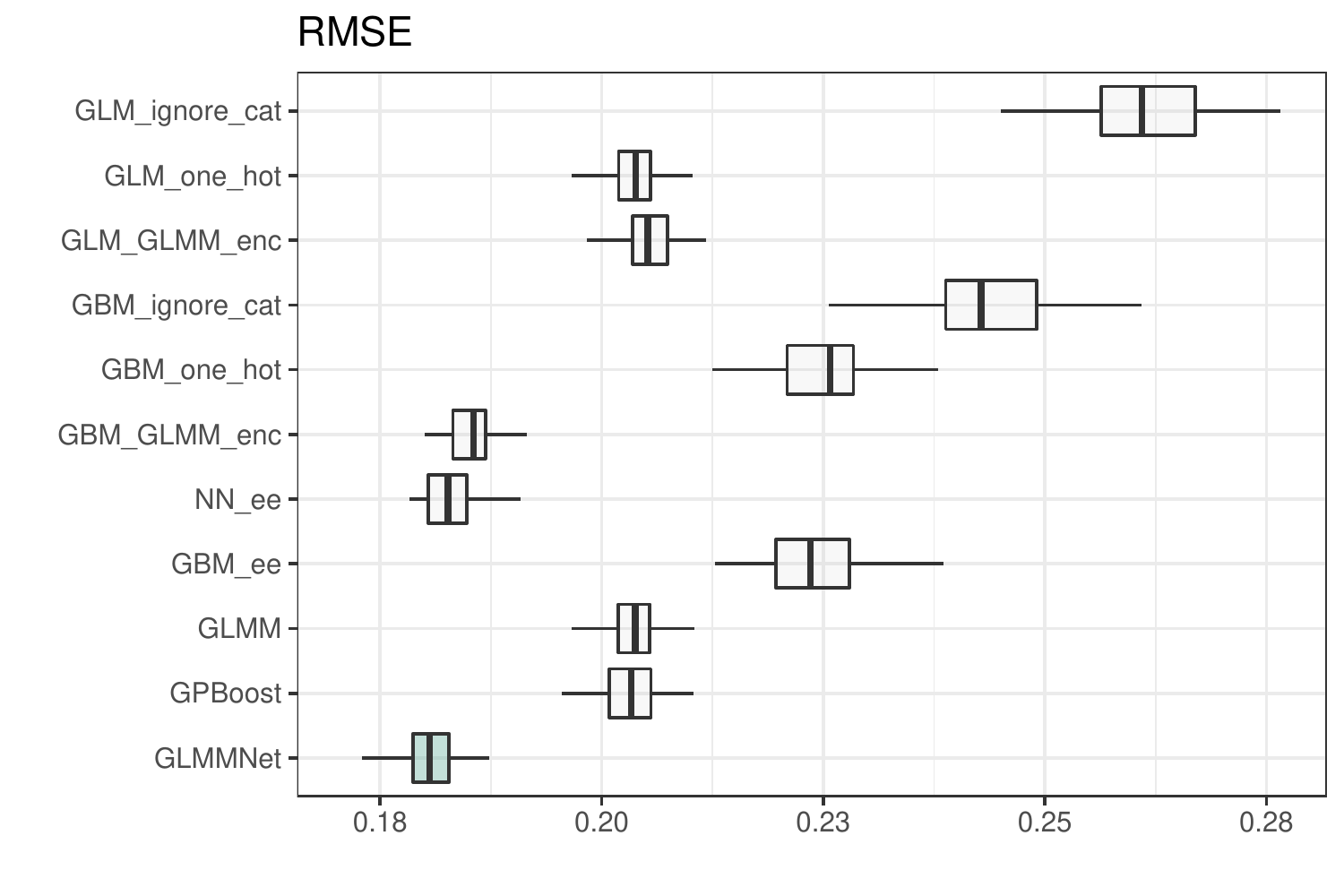} }\newline\subfloat[Continuously ranked probability score (CRPS)\label{fig:boxplots-1-3}]{\includegraphics[width=0.49\linewidth]{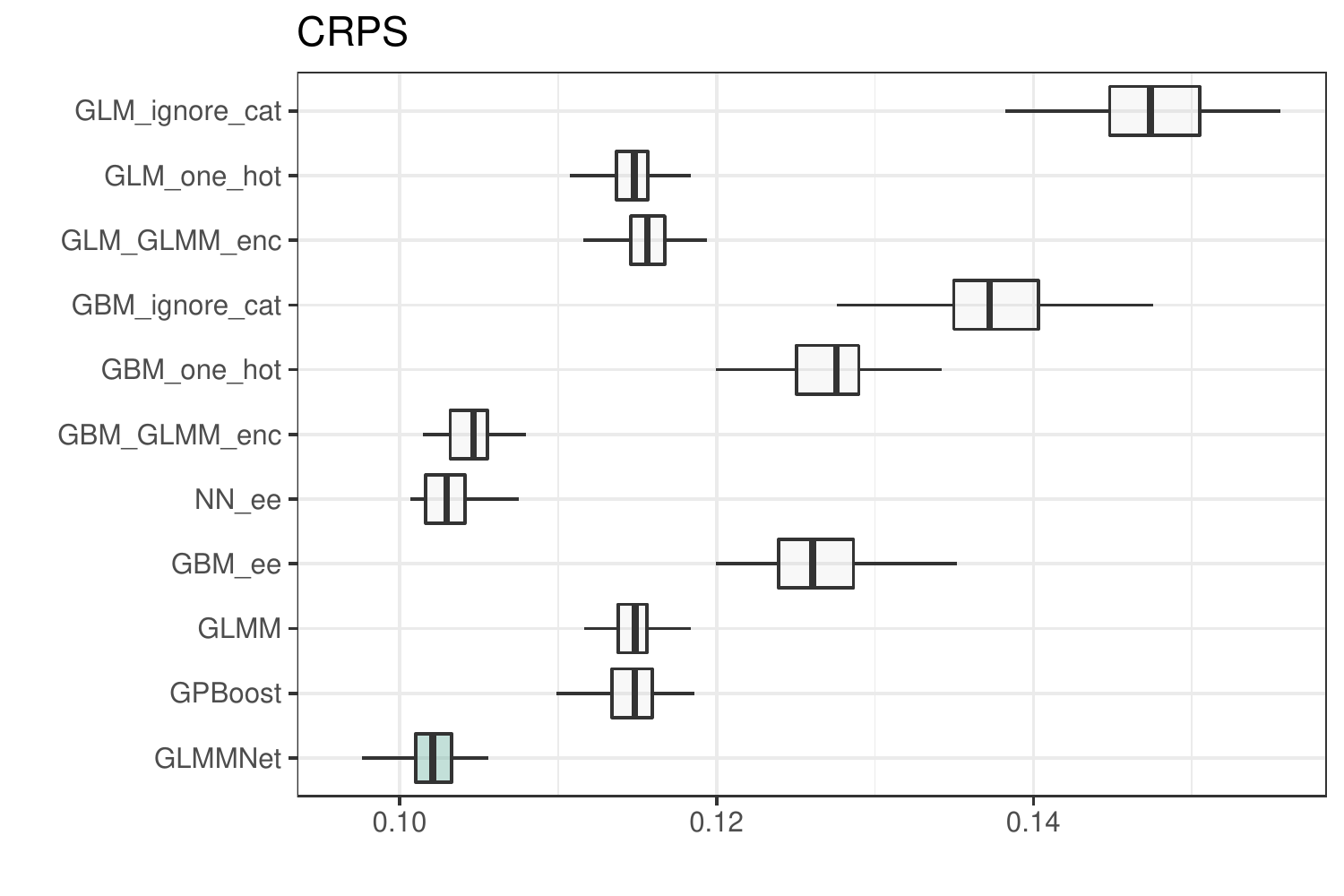} }\subfloat[RMSE of average prediction per category\label{fig:boxplots-1-4}]{\includegraphics[width=0.49\linewidth]{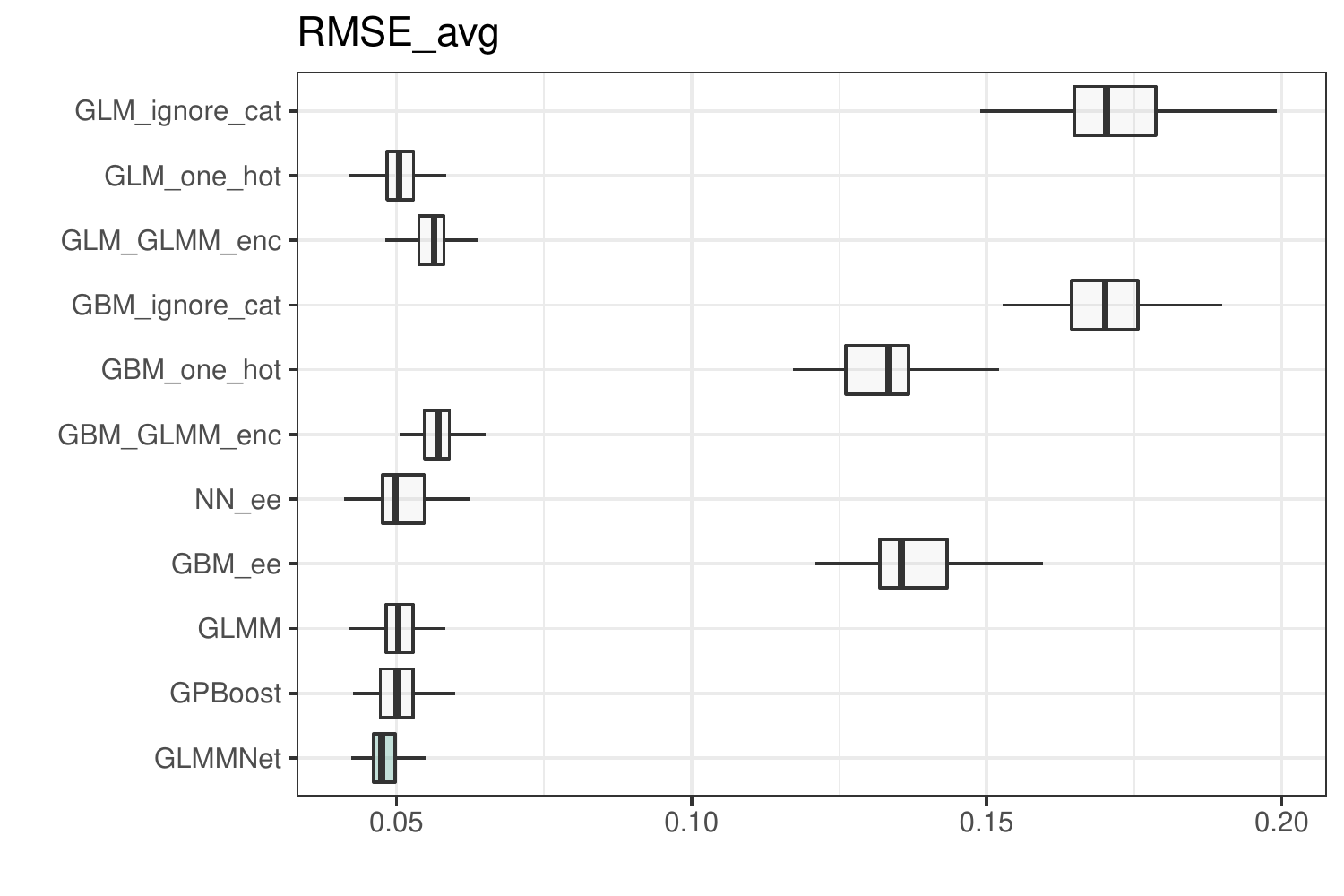} }
}

\caption{Boxplots of out-of-sample performance metrics of the different models; GLMMNet highlighted in green. Each experiment is repeated 50 times, with 5,000 training observations and 2,500 testing observations each.}\label{fig:boxplots-1}
\end{figure}

Figure \ref{fig:boxplots-1} compares the out-of-sample performance of the candidate models (listed in Section~\ref{tab:candidates}) over 50 simulation runs (of 5,000 training and 2,500 testing observations each) under a base scenario (experiment 1 in Table \ref{tab:scenarios}); the proposed GLMMNet is highlighted in green. For all metrics considered, smaller values mean a better fit.

From Figure \ref{fig:boxplots-1}, we see that \texttt{GLMMNet} is leading in all metrics. In particular, on average it performs better than the current state-of-the-art neural network with entity embeddings (\texttt{NN\_ee}), which is also the next best model. A paired sample Wilcoxon signed rank test on the out-of-sample MAE rejects the null hypothesis---that \texttt{GLMMNet} and \texttt{NN\_ee} perform the same---with strong evidence (\(p < 0.001\)), in favour of the alternative that \texttt{GLMMNet} produces a smaller error. The same conclusion holds for all metrics tested. Indeed, upon closer examination of the paired sample data, we find that for every single simulation run, the performance of \texttt{GLMMNet} surpasses that of \texttt{NN\_ee}. This suggests that optimising the networks via the (more suitable) mixed model likelihood provides a significant benefit to their predictive performance, in terms of both point predictions and (more notably) probabilistic predictions.

Ignoring \texttt{GLMMNet}, the best performing models in panels (a)--(c) are \texttt{NN\_ee} and \texttt{GBM\_GLMM\_enc} (a boosting model with GLMM encoding). Both models significantly outperform the linear family of models (GLM or GLMM), showing that a flexible model structure is required to capture the non-linearities in the data.

More importantly, without a suitable way of accounting for the high-cardinality categorical feature, a flexible model structure alone is not enough to achieve good performance. For example, the struggle of tree-based models in dealing with categorical features has been well documented \citep{prokhorenkova2018}. In this experiment, with the usual one-hot encoding, the GBM model (\texttt{GBM\_one\_hot}) performs even worse than its GLM counterpart (\texttt{GLM\_one\_hot}); although the reverse is true when both are fitted without the categorical variable altogether. This confirms the motivation for this research: more flexible ML models, when dealing with financial or insurance data, can often be constrained by their capability to model categorical variables with many levels.

The ranking of models in Figure~\ref{fig:boxplots-1}(d)---which is designed to compare the models' ability to explain between-category variations---is slightly different. \texttt{GLM\_one\_hot} and \texttt{GLMM} perform better, moving up in the ranking to join the (consistently) top-performing \texttt{GLMMNet} and \texttt{NN\_ee}. This is a result of the balance property (for GLMs with canonical links) which ensures that the sum of fitted values must equal exactly the sum of observed values for any level of the categorical features in the training set \citep{wuethrich2021}. In this case, the training RMSE\_avg will be zero by definition. In practice, it is remarkable that this quality often carries over to an out-of-sample set as well, as demonstrated by the performance of \texttt{GLM\_one\_hot} and \texttt{GLMM} here.

\begin{figure}[!htbp]
\begin{center}
\includegraphics[width=\linewidth]{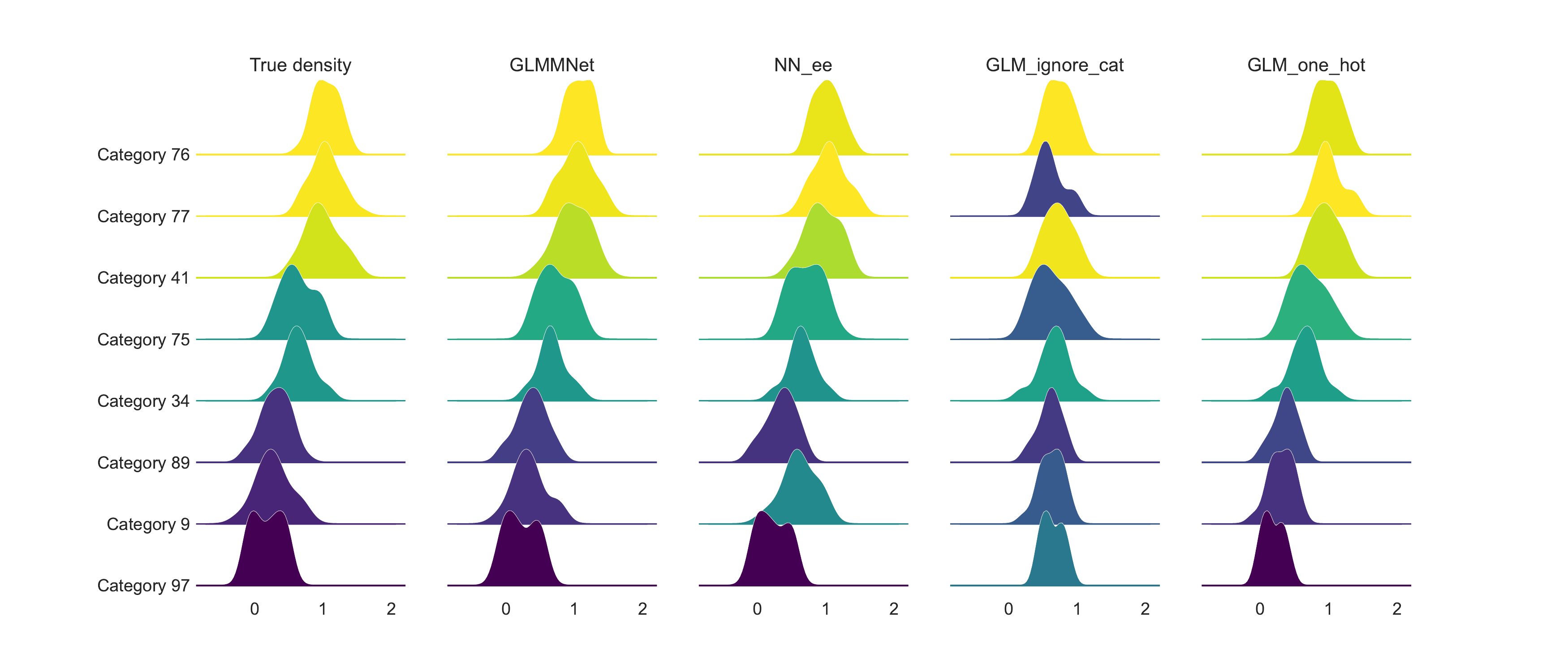} 
\caption{True \textit{versus} predicted densities for (selected) categories on the test set in experiment 1}\label{fig:densities-1}
\end{center}
\end{figure}

Figure~\ref{fig:densities-1} gives a more insightful picture of per-category predictive accuracy by plotting the densities of predictions from selected models against the true underlying density that generated the observations (leftmost). The plot shows a selection of representative categories (top three with the highest response values, two in the middle, and bottom three with the lowest), but the conclusions drawn here hold generally for other categories too. The colours roughly correspond to the range of values under which the predictions fall; darker colours represent smaller values (relative to an average observation), and lighter represent larger values. Ideally, the predicted density shapes should match as closely as possible to the true densities, and the colours should also agree as much as possible.

In line with the previous observations, \texttt{GLMMNet} produces very accurate density estimates and models the variation in the data extremely well. \texttt{NN\_ee}, on the other hand, clearly overestimates the response for Category 9, likely due to an overfit to the training data points. Unsurprisingly, ignoring the categorical variable (\texttt{GLM\_ignore\_cat}) almost eliminates the between-category differences and leads to very poor estimates. When equipped with one-hot encoding, the GLM (\texttt{GLM\_one\_hot}) is able to sense the differences in the mean (as indicated by the colours), as the balanced distribution of categories provides sufficient data points for learning the between-category differences. However, \texttt{GLM\_one\_hot} tends to produce narrower densities than the truth. This indicates a failure to capture the remaining, or within-category, variation in the data, confirming that its linear structure is too restrictive for this dataset.

The results from experiments 2--3 are mainly consistent with what we have seen in experiment 1 (see Appendix~\ref{app:sim23}). \texttt{GLMMNet} takes a strong lead in both cases and showcases a statistically significant advantage over its closest competitor \texttt{NN\_ee} in all cases (\(p < 0.01\) from the corresponding Wilcoxon signed rank tests).

The result in experiment 2 (gamma-distributed response) is much within expectation; the GLMMNet by design takes good care of the response distribution via its loss function. The result in experiment 3 (skewed distribution of categories) further suggests that the predictive strength of the GLMMNet is, at least to some degree, immune to the shape of the categorical distribution.

As the noise level increases (experiments 4--6), the differences in model performance become smaller, and the predictive advantage of the GLMMNet starts to fade away (see Appendix~\ref{app:sim46}). In experiments 5--6, the GBM family of models surpasses both \texttt{GLMMNet} and \texttt{NN\_ee} in CRPS measures. One possible explanation is that the variance of the random effects here is too small with respect to the error variance to warrant the need for accurately modelling the categorical variable. Although \texttt{GLMMNet} may not always provide a boost to the overall predictive performance, it remains one of the top performers in terms of its ability to capture the individual category means (as measured by RMSE\_avg). The latter may be of practical significance in many applications.

\begin{figure}[htbp!]
{\centering \subfloat[Exp 5, CRPS\label{fig:boxplots-add-1}]{\includegraphics[width=0.49\linewidth]{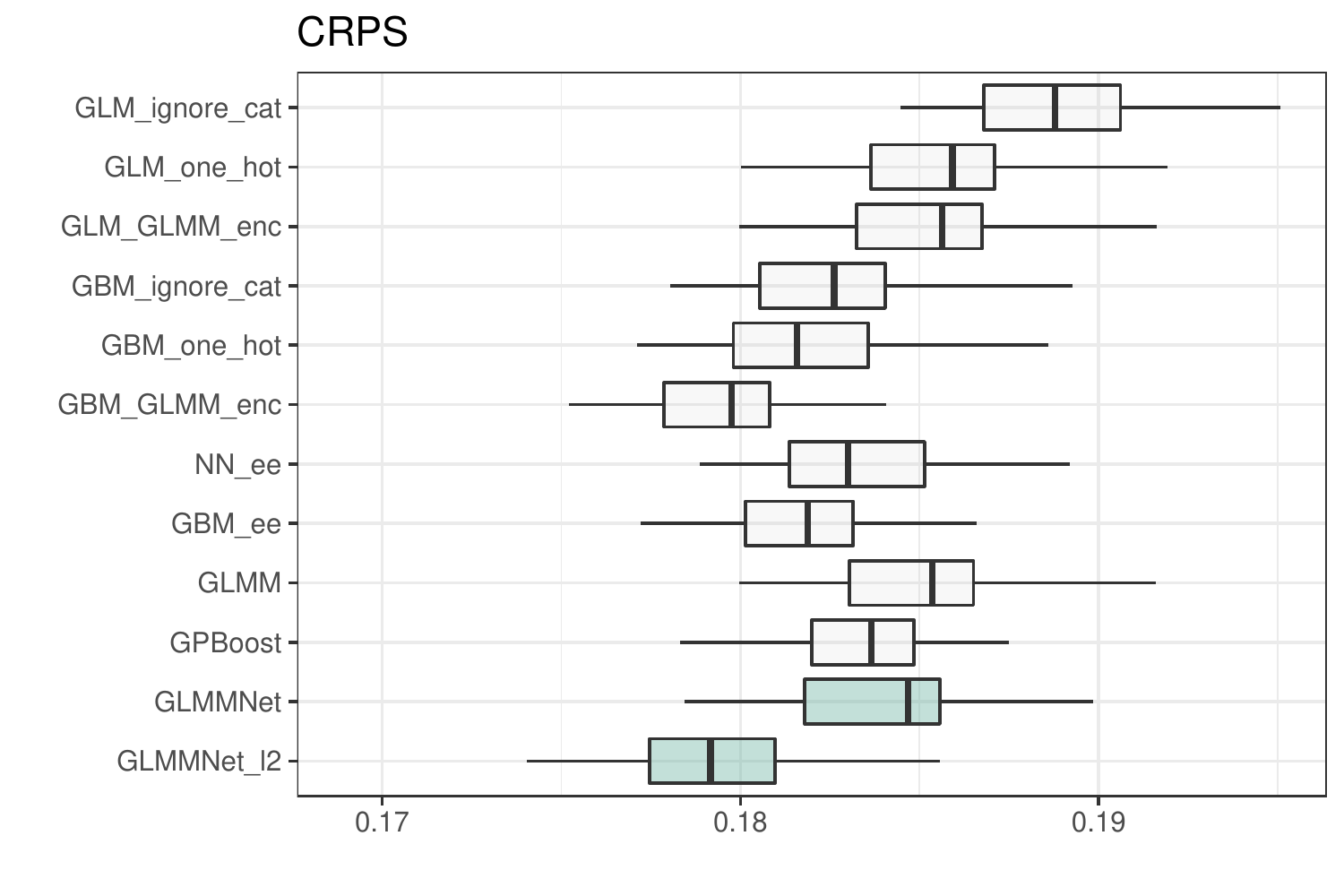} }\subfloat[Exp 5, RMSE of average prediction per category\label{fig:boxplots-add-2}]{\includegraphics[width=0.49\linewidth]{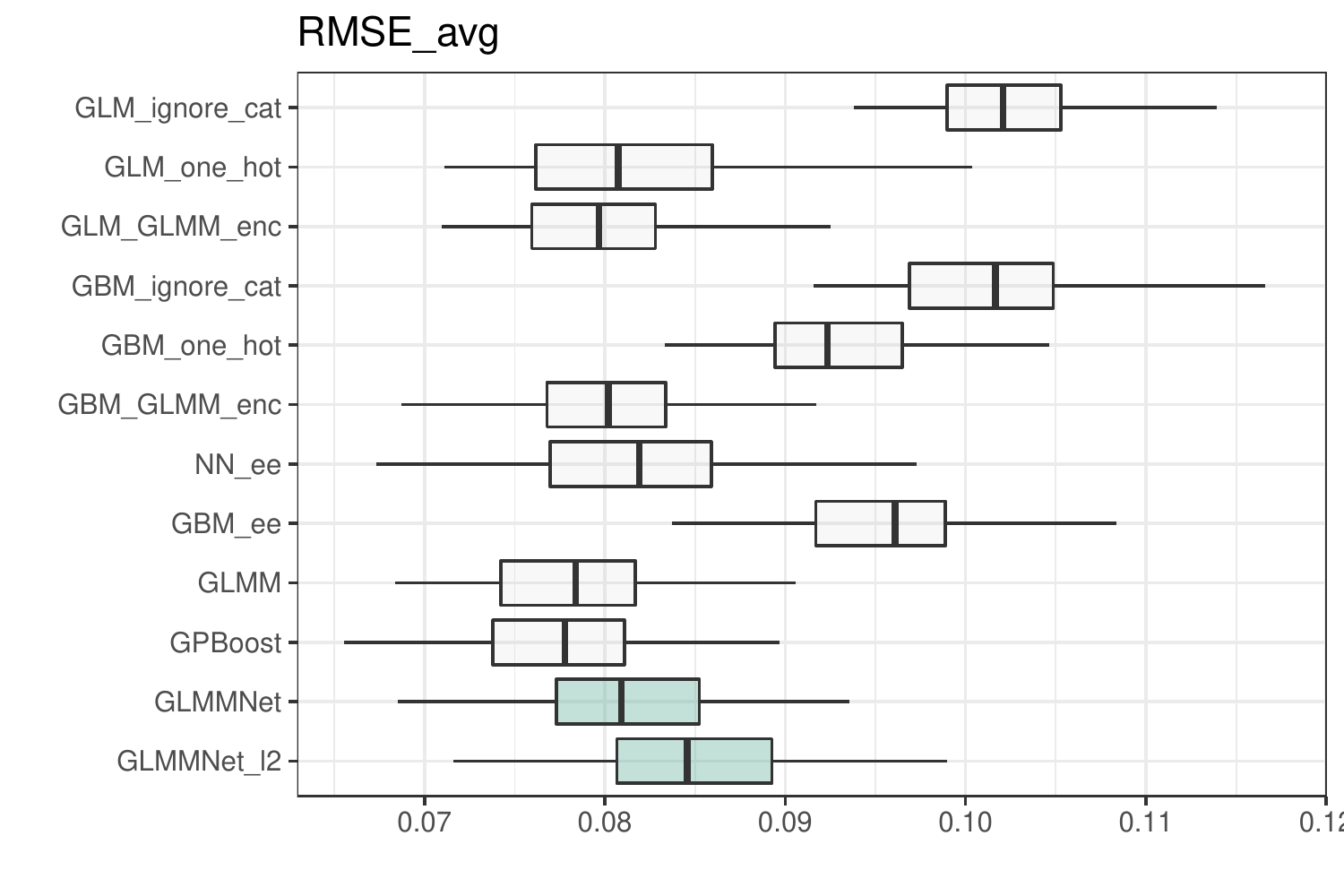} }
}
\caption{Boxplots of out-of-sample performance metrics of the different models in experiment 5 (middle; high noise Gaussian), \textbf{shown with an $\ell_2$-regularised GLMMNet}. Each experiment is repeated 50 times, with 5,000 training observations and 2,500 testing observations each.\vspace{2mm}}\label{fig:boxplots-add}
\end{figure}

The outperformance of GBM over deep learning models as observed in experiments 5 and 6 was already discussed in the prior literature, see, e.g. \citet{borisov2021}, though it remains an open question what causes this gap in performance. It is possible that neural networks are more likely to overfit to the high noise in the background, or that they may get stuck in a suboptimal local minimum and cannot gain enough momentum from the data to escape. In general, the high level of noise in the environment prevents the network-based models (including the proposed GLMMNet) from fully realising their predictive power. Further investigations suggest that adding regularisation helps improve the overall performance of the GLMMNet to a great extent, as shown in panel (a) of Figure \ref{fig:boxplots-add}, though it seems to bias the point predictions (panel b).

Contrasting experiments 5 and 6 shows that the \texttt{GLMMNet} ranks higher when the response moves away from Gaussian (see Figure~\ref{fig:boxplots-4} in Appendix~\ref{app:sim46}). The predictive power that the \texttt{GLMMNet} regains in experiment 6 is likely a result of the change in the true distribution of the response. When the shape of the response looks less normal, the outperformance of GBM models over the GLMMNet becomes less distinctive. This comparison highlights the advantage of the GLMMNet in using a loss function that is better aligned with the distribution of the response. We expect this effect to be more pronounced when we work with heavier-tailed distributions that deviate even more from the normal.

Across all experiments, we find that the \texttt{GLMMNet}, \texttt{NN\_ee}, and \texttt{GBM\_GLMM\_enc} consistently outperform the other models in terms of predictive performance. In particular, we find that our GLMMNet outperforms or at least performs on par with the target benchmark model of an entity embedded neural network in all scenarios studied. That said, each model has its own strengths and limitations, and the choice of which model to use will depend on the specific needs of the modelling problem. Table~\ref{tab:strengths} lists some of these considerations. We should also note that \texttt{GBM\_ee} has similar properties to \texttt{GBM\_GLMM\_enc} and is therefore not included in the table for brevity of presentation (except for the fact that \texttt{GBM\_ee} is more complicated to implement as it requires training an entity-embedded neural network beforehand).

\newcommand{\tabitem}{~~\llap{\textbullet}~~}
\begin{table}[ht]
\centering
\begin{footnotesize}
\begin{tabular}{p{1.5cm}p{6.5cm}p{6.5cm}}
\toprule
 & \multicolumn{1}{c}{Strengths} & \multicolumn{1}{c}{Limitations} \\ \midrule
\texttt{NN\_ee} & \raggedright \tabitem Strong predictive performance, particularly in low-medium noise environments & {\raggedright \tabitem Compromised performance in high noise environments} \newline {\raggedright \indent \tabitem Limited interpretability} \\ \midrule
\texttt{GBM\_GLMM\_enc} & \raggedright \tabitem Good predictive performance, particularly in high noise environments \newline \raggedright \indent \tabitem Simpler structure and faster to train  & {\raggedright \tabitem Compromised performance in low-medium noise environments} \newline {\raggedright \indent \tabitem Limited interpretability of the effects of the categories} \\ \midrule
\texttt{GLMMNet} & \raggedright  \tabitem Consistently outstanding predictive performance, particularly in low-medium noise environments and/or when the response deviates from the Gaussian distribution \newline \raggedright \indent \tabitem Transparency on the category effects (through random effects predictions) \newline \raggedright \indent \tabitem Offers naturally probabilistic estimates & {\raggedright \tabitem Compromised performance in high noise environments when used without additional regularisation} \newline {\raggedright \indent \tabitem Limited interpretability of the fixed effects component} \\ \bottomrule
\end{tabular}
\caption{Model comparison: strengths and limitations of the top performing models} \label{tab:strengths}
\end{footnotesize}
\end{table}

\section{Application to Real Insurance Data} \label{sec:case-study}
In this section, we apply GLMMNet to a real (proprietary) insurance dataset (described in Section~\ref{ssec:data}), and discuss its performance relative to the other models we have considered (Section~\ref{ssec:real-results}). We also demonstrate, in Section~\ref{ssec:practical-insights}, how such an analysis can potentially inform practitioners in the context of technical pricing.

\subsection{Description of Data} \label{ssec:data}
The data for this analysis were provided by a major Australian insurer. The original data cover 27,351 commercial building and contents reported claims by small and medium-sized enterprises (SME) over the period between 2010 and 2015. The analysis seeks to construct a regression model to predict the ultimate costs of the reported claims based on other individual claim characteristics that can be observed early in the claims process (e.g.~policy-level details and claim causes). A description of the (post-engineered) input features is given in Appendix~\ref{app:predictors}.

\begin{figure}[htbp!]
\begin{center}
\includegraphics[width=0.5\linewidth]{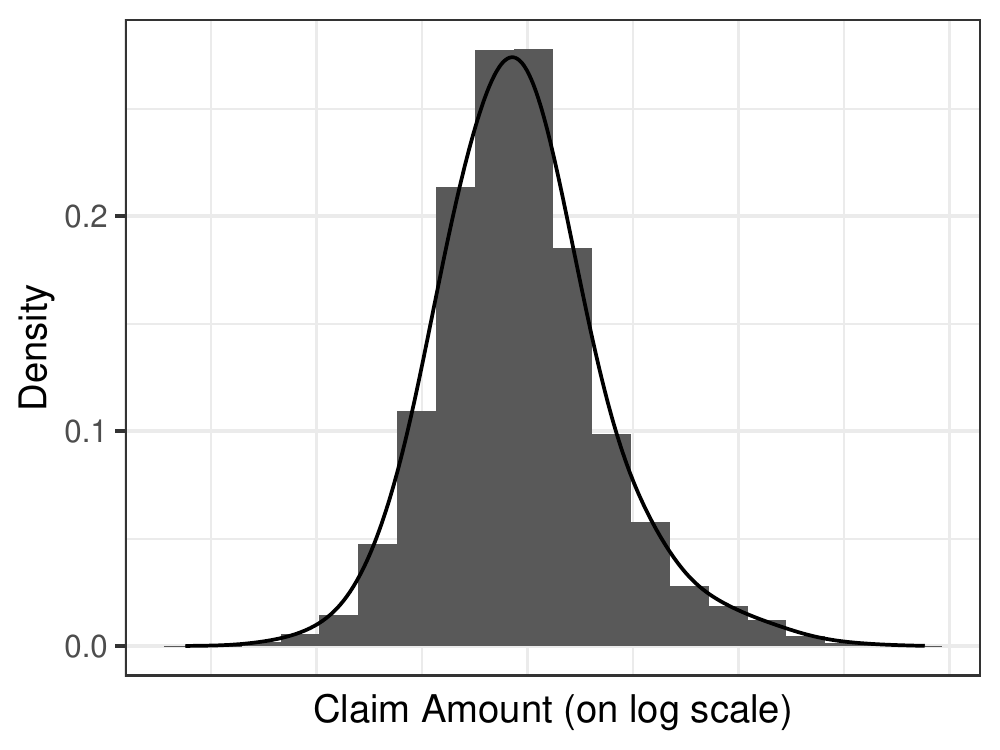}
\caption{Histogram of claim amounts (on log scale). The $x$-axis numbers have been deliberately removed for confidentiality reasons.}\label{fig:hist}
\end{center}
\end{figure}

\begin{figure}[htbp!]
\subfloat[P-P plot (lognormal)\label{fig:incurred-3}] {\includegraphics[width=0.5\linewidth]{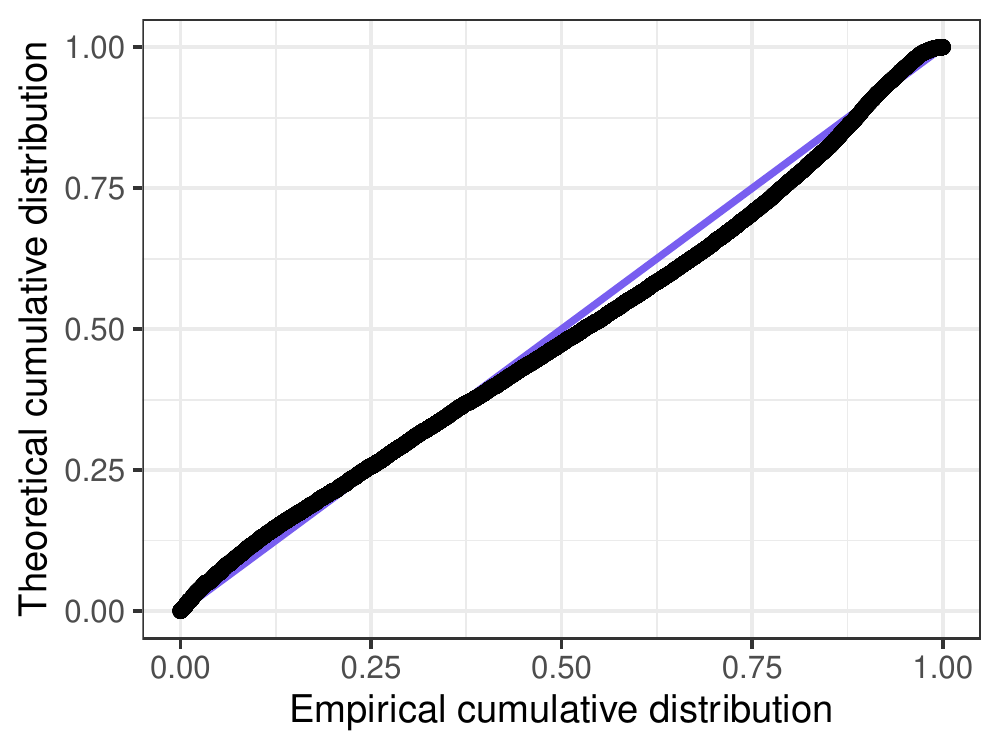} }\subfloat[P-P plot (loggamma)\label{fig:incurred-4}]{\includegraphics[width=0.5\linewidth]{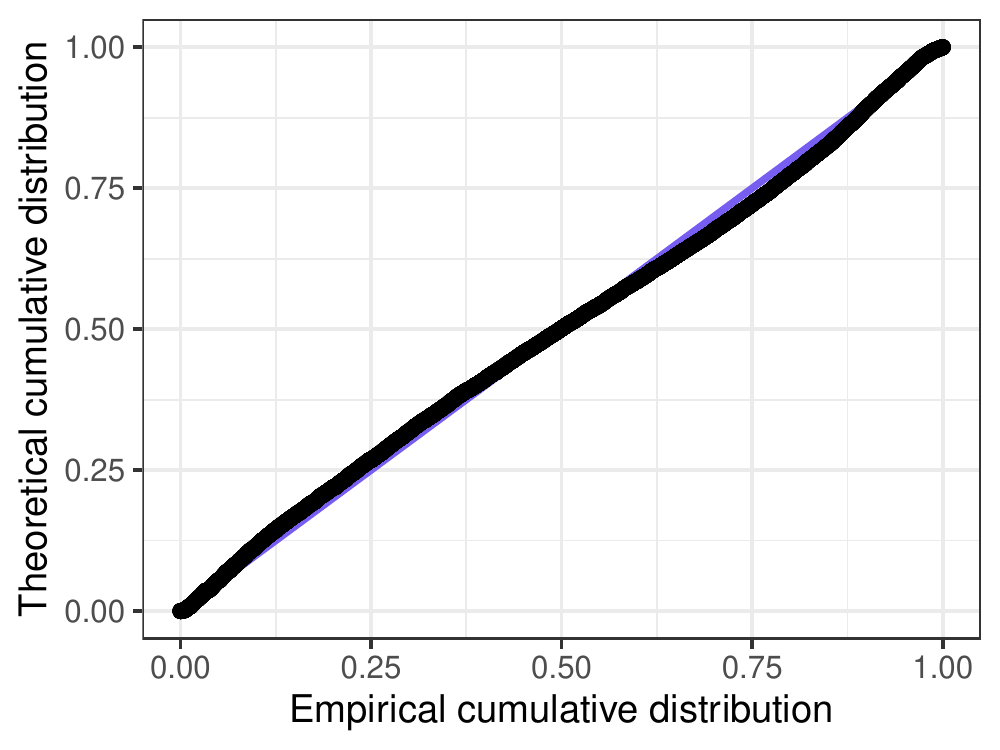} }
\caption{Probability-probability (P-P) plot of empirical \textit{versus} fitted lognormal and loggamma (theoretical) distributions. The parametric distributions are fitted to the (unlogged) marginal, i.e. without any covariates.}\label{fig:incurred}
\end{figure}

The response variable is the claim amount (claim severity), which has a very skewed distribution, as shown in Figure~\ref{fig:hist}: even on a log scale, we can still observe some degree of positive skewness. In search for a suitable distribution to model the severity, we fit both lognormal and loggamma distributions to the (unlogged) marginal. Figure~\ref{fig:incurred} indicates that both models may be suitable, with loggamma being slightly more advantageous. We will test both lognormal and loggamma distributions in our experiments below (Section~\ref{ssec:real-results}); specifically, we fit Gaussian and gamma models to the log-transformed response.

The high-cardinality categorical variable of interest is the occupation of the business, which is coded through the Australian and New Zealand Standard Industrial Classification (ANZSIC) system. The ANZSIC system hierarchically classifies occupations into Divisions, Subdivisions, Groups and Classes (from broadest to finest). See Appendix~\ref{app:anzsic} for an example.

\begin{figure}[ht!]
\begin{center}
\subfloat[Number of claims from the top 20 common occupation classes\label{fig:cat-1}]
{\includegraphics[width=0.49\linewidth]{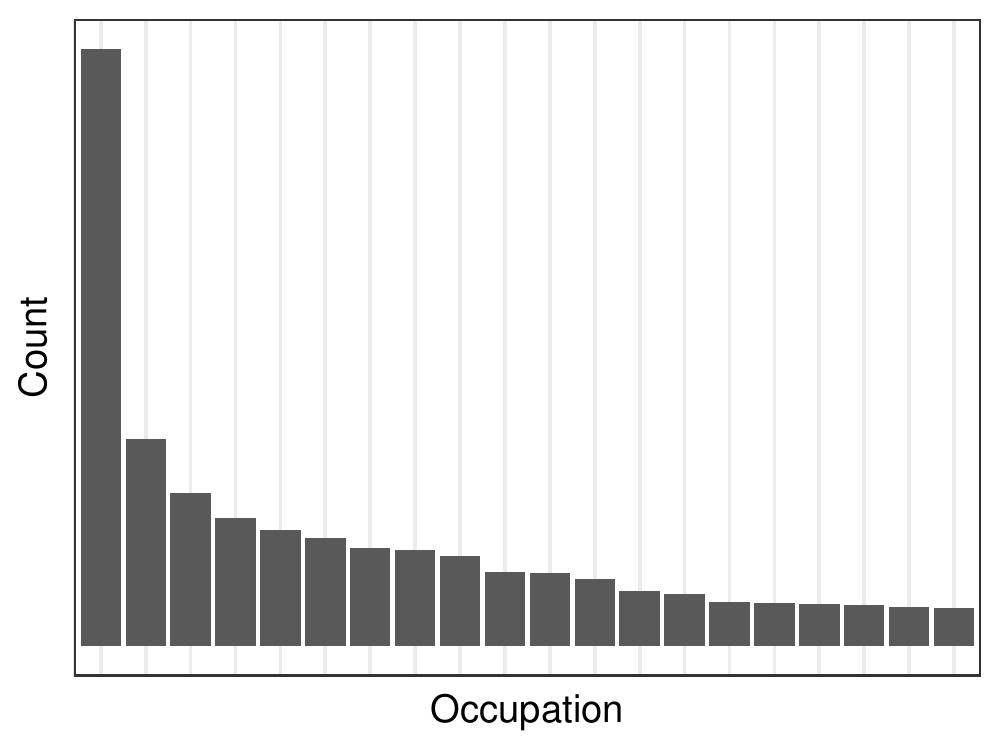} }
\subfloat[Average claim size for the (same) top 20 common occupation classes\label{fig:cat-2}]{\includegraphics[width=0.49\linewidth]{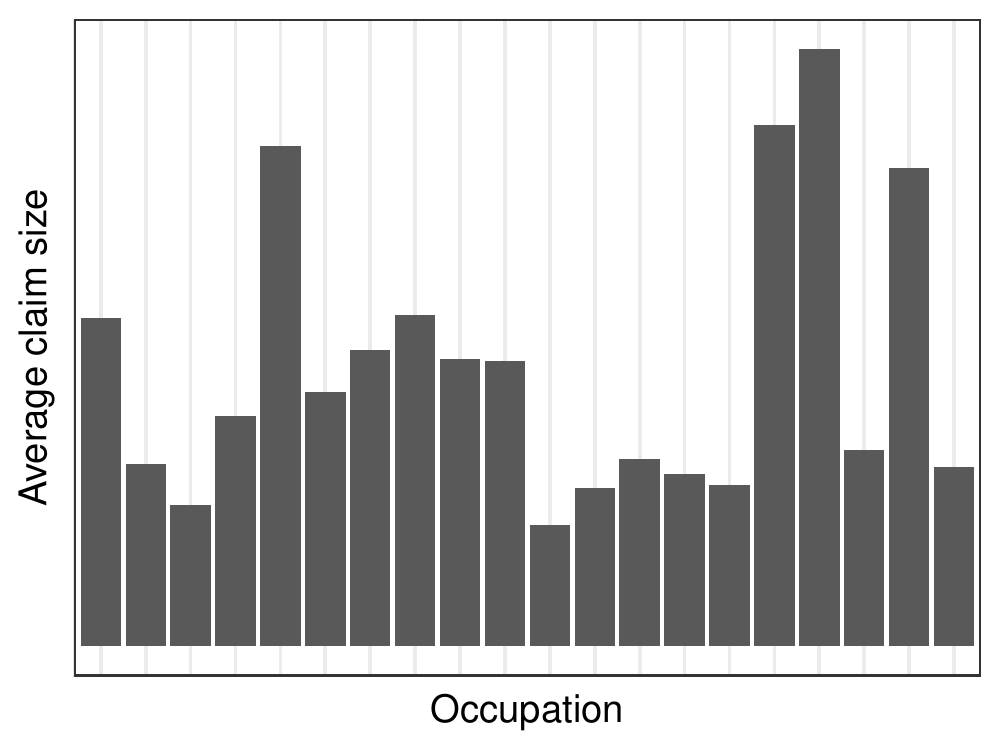}}
\caption{Skewed distribution of occupation class and heterogeneity in experience. The axes labels have been removed to preserve confidentiality.}\label{fig:cat}
\end{center}
\end{figure}

We will look at occupations at the Class level.\footnote{We chose not to incorporate the hierarchical information. Instead, we directly modelled the lowest level (i.e. Class) as a flat categorical feature. The decision was made in light of (1) the increased complexity of estimation in hierarchical models, and (2) the lack of alignment between the hierarchical system built for general-purpose occupation classification and the factors that differentiate claim severities.} There are over 300 unique levels of such occupation classes in this data set. Panel (a) of Figure \ref{fig:cat} highlights the skewed distribution of occupations: the most common occupation has more than double the number of observations than the second most common, and the number of observations decays rapidly for the rarer classes. Panel (b) shows the heterogeneity in claims experiences between different occupation classes; calculations reveal a coefficient of variation of 160\% for the occupation means. One challenge in modelling the occupation variable lies in determining how much confidence we can have in the observed claims experiences; intuitively, we should trust the estimates more when there are more data points, and less when there are few data points. As explained in Section~\ref{ssec:glmm}, the mixed effects models are one solution to this problem, which we will explore further in this section.

The dataset under consideration displays many typical characteristics of insurance data that we studied in Section~\ref{sec:simulation} (see Section~\ref{ssec:sim-env} in particular). For example, it is characterised by an extremely skewed response distribution (Figure~\ref{fig:hist}), a low signal-to-noise ratio (there is little information that covariates can add to the marginal fits, as implied by Figure~\ref{fig:incurred}) and a highly disproportionate distribution of categories (Figure~\ref{fig:cat}). These features of the dataset best align with those of Experiment 6 in Section~\ref{sec:simulation}. If the conclusions generalise, we expect that a regularised version of our GLMMNet will show strong performance.

\subsection{Modelling Results} \label{ssec:real-results}
We consider the same candidate models as before; see Section \ref{ssec:candidates}. Having noted the underperformance of GLMMNet in the presence of high noise and the benefit of adding regularisation to it, we also consider an \(\ell_2\)-regularised GLMMNet (\texttt{GLMMNet\_l2}) in the hope that the regularisation would help improve performance.

We follow the learning pipeline described at the start of Section~\ref{ssec:metrics} and split the data into a 90\% training set and a 10\% test set. Table~\ref{tab:cs-results} presents the out-of-sample metrics for the top performing lognormal and loggamma models respectively, compared against a one-hot encoded GLM as a baseline (full results in Appendix~\ref{app:case-study}).\footnote{In Table~\ref{tab:cs-results} and the relevant tables in Appendix~\ref{app:case-study}, the median absolute error (MedAE) and NLL are calculated on the original scale of the data (i.e. after the predictions are back-transformed). The CRPS is calculated on the log-transformed scale, because there is no closed-form formula for the loggamma CRPS. As a result of this, the CRPS as presented here will be more tolerant of wrong predictions for extreme observations than the NLL.} Note that we have selected slightly different metrics from the simulation experiments to better accommodate the heavier-tailed response. The RMSE, for example, is no longer a suitable measure as it penalises very heavily for predictions that are far from the observed values, which can happen very frequently when there are a few extreme observations in the data.

\begin{table}[!htbp]
\begin{center}
\begin{small}
\begin{tabular}{@{}lcccccc@{}}
\toprule
& \multicolumn{3}{c}{\textbf{Lognormal (out-of-sample)}} & \multicolumn{3}{c}{\textbf{Loggamma (out-of-sample)}} \\ 
& MedAE  & CRPS  & NLL & MedAE  & CRPS  & NLL \\ \midrule
GLM\_one\_hot & 4108 & 0.7931 & 9.623 & 1946 & 0.8557 & 9.751 \\
GBM\_GLMM\_enc & 3870 & 0.7666 & 9.584 & \textbf{1536} & 0.7626 & 9.578 \\
NN\_ee & 4086 & 0.7666 & 9.584 & 1606 & \textbf{0.7612} & 9.578 \\
GBM\_ee & 3828 & 0.7665 & 9.584 & 1549 & 0.7629 & 9.579 \\
GLMM & 3864 & 0.7666 & 9.584 & 1570 & 0.7629 & \textbf{9.577} \\
GLMMNet & 3783 & 0.7751 & 9.595 & 1633 & 0.7662 & 9.583 \\
GLMMNet\_l2 & \textbf{3549} & \textbf{0.7634} & \textbf{9.580} & 1618 & 0.7626 & \textbf{9.577} \\ \bottomrule
\end{tabular}
\end{small}
\end{center}
\caption{Comparison of lognormal and loggamma model performance (median absolute error, CRPS, negative log-likelihood) on the test (out-of-sample) set. The best values are bolded.} \label{tab:cs-results}
\end{table}

The three metrics tell consistent stories most of the time. The results for CRPS and NLL are almost perfectly aligned, which is not surprising given that both are measures of how far the observed values depart from the probabilistic predictions. In general, we consider the CRPS and NLL measures to be more reliable estimates of the goodness-of-fit of the models than the MedAE, which only takes into account the point predictions but not the uncertainty associated with them.

The results mostly match our expectation. Among the lognormal models, the regularised \texttt{GLMMNet\_l2} takes the lead, followed by \texttt{GBM\_GLMM\_enc}, \texttt{NN\_ee}, \texttt{GBM\_ee} and \texttt{GLMM}, whose performances are so close together that it is impossible to distinguish between the four. The \texttt{GLM} family of models performs poorly on this data, much worse than their mixed model counterpart, i.e.~\texttt{GLMM}, confirming that the presence of the high-cardinality categorical variable interferes with their modelling capabilities. Among the loggamma models, although \texttt{NN\_ee} outperforms \texttt{GLMMNet\_l2} in CRPS, \texttt{GLMMNet\_l2} remains the most competitive model in terms of NLL (on par with \texttt{GLMM}). One practical consideration to note is that when the environment is highly noisy and the underlying model is uncertain, such as here, the interpretability of the model becomes even more important. In such cases, models that are more interpretable are preferred, which in turn, highlights the advantages of \texttt{GLMMNet} over \texttt{NN\_ee}.

An important observation is that adding the \(\ell_2\) regularisation term clearly helps GLMMNet navigate the data better. In the lognormal models, the regularised network represents a 1.5\% improvement in the CRPS and NLL from the original \texttt{GLMMNet}, which suggests that the original \texttt{GLMMNet} overfits to the training data (as can be confirmed from its deterioration in the test performance). Indeed, the amelioration of the score proves statistically significant ($p < 0.001$) under the Diebold-Mariano test \citep{gneiting2014}. The results for the loggamma models lead to similar conclusions.

Comparing the lognormal and loggamma models, we find that assuming a loggamma distribution for the response improves the model fit in almost all instances (except the GLM models) and across all three performance metrics. While the CRPS and NLL values are reasonably similar between the two tables, the median absolute error (MedAE) halves when moving from lognormal to loggamma. An intuitive explanation is that the lognormal models struggle more with fitting to the extreme claims and thus tend to over-predict the middle-ranged values in an attempt to boost the predictions. The loggamma models, on the other hand, are more comfortable with the extremes, as those can be captured reasonably well by the long tail. These results confirm the need to consider alternative distributions beyond the Gaussian---one major motivation for our proposed extension of the LMMNN to the GLMMNet. Even within the Gaussian framework, as demonstrated by the lognormal models presented here, the GLMMNet can still be useful.

Overall, as we have seen in the simulation experiments, the differences in model performance are relatively small due to the low signal-to-noise ratio. Only a limited number of risk factors were observed, and they are not necessarily the true drivers of the claims cost. In the experiments above, a simpler structure like GLMM sometimes performs comparably with the more complex and theoretically more capable GLMMNet. That said, the results clearly demonstrate that the GLMMNet is a promising model in terms of predictive performance. We expect to see better results with less noisy data or a greater volume of data, e.g.~when more information becomes available as a policy progresses and claims develop, reducing the amount of noise in the system. This has been demonstrated in the simulation experiments (Section \ref{ssec:sim-results}).

\subsection{GLMMNet: Decomposing Random Effects Per Individual Category} \label{ssec:practical-insights}
One main advantage of mixed effects models in dealing with high-cardinality categorical features is the transparency they offer on the effects of individual categories. This is particularly important as it is often the case that the high-cardinality categorical feature is itself a key risk factor. For example, in this data set, there is a lot of heterogeneity in claims experience across different occupations, but the lack of data for some makes it hard to determine how much trust can be given to the experience.

\begin{figure}[!htbp]
\begin{center}
\includegraphics[width=0.75\linewidth]{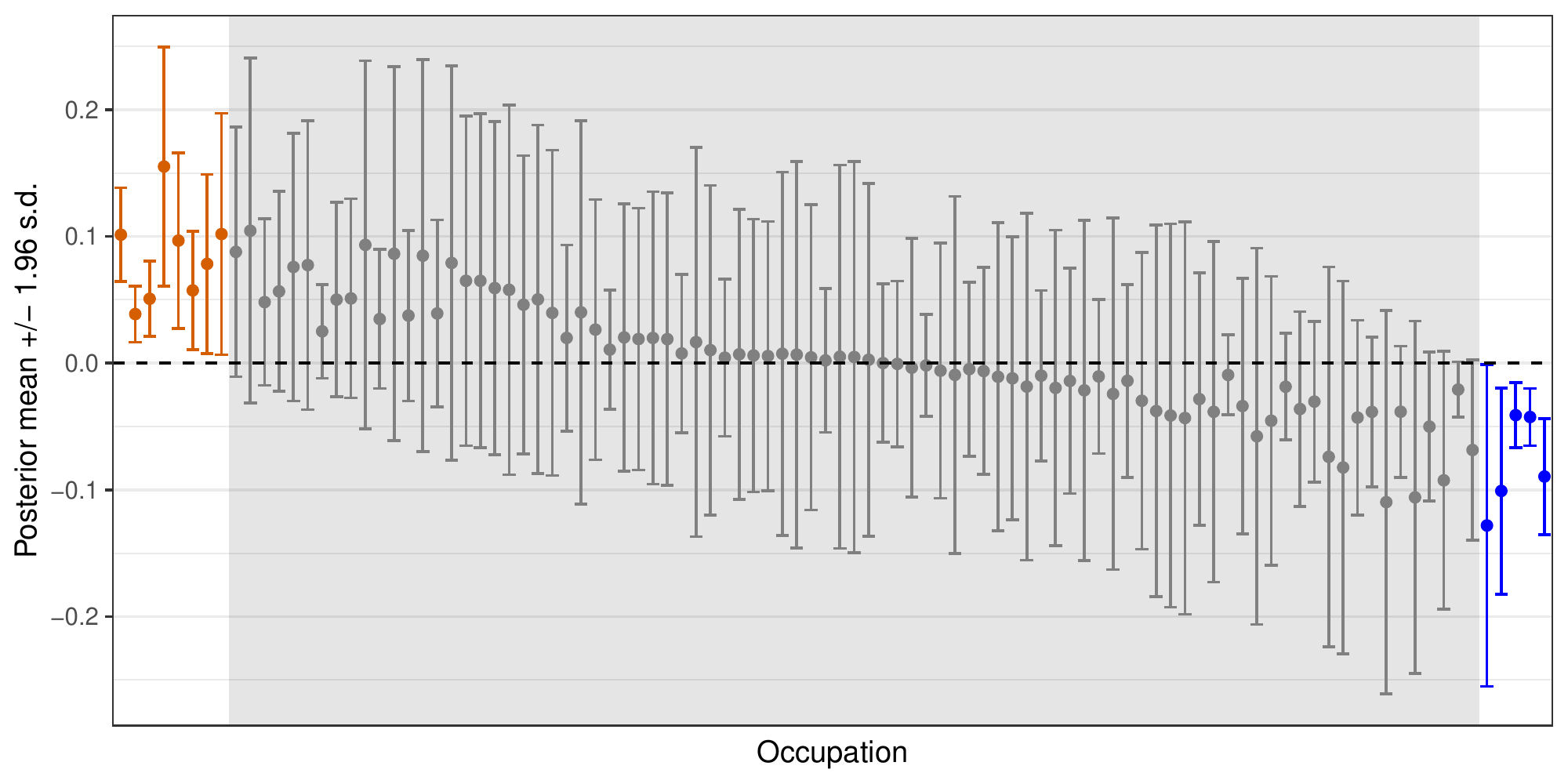} 
\caption{Posterior predictions of (a randomly selected sample of) the random effects in 95\% confidence intervals from the loggamma GLMMNet; ordered by decreasing $z$-scores. Occupations that do not overlap with the zero line are highlighted in vermillion (if above zero) and blue (if below zero) respectively. Occupations that \textit{do} overlap with the zero line are in the shaded region. The \textit{x}-axis labels have been removed to preserve confidentiality.}\label{fig:pr-1}
\end{center}
\end{figure}

The GLMMNet can provide useful insights in this regard. Figure \ref{fig:pr-1} shows the posterior predictions for the effects of individual occupations (ordered by decreasing $z$-scores), plotted with 95\% confidence intervals. Unsurprisingly, occupations with a larger number of observations generally have tighter intervals (i.e.~lower standard deviations); see Figure~\ref{fig:pr-2} of Appendix~\ref{app:case-study}. This prevents us from overtrusting extreme estimates, which may happen simply due to the small sample size. In the context of technical pricing, such an analysis can be helpful for identifying occupation classes that are statistically more or less risky. This is not possible with the often equally or less high-performing neural networks with entity embeddings. With the latter, we can analyse the embedding vectors to identify occupations that are ``similar'' to each other in the sense of producing similar output, but we cannot go any further.

\section{Conclusions} \label{sec:conclusion}
High-cardinality categorical features are often encountered in real-world insurance applications. The high dimensionality creates a significant challenge for modelling. As discussed in Section~\ref{ssec:background}, the existing techniques prove inadequate in addressing this issue. Recent advancements in ML modeling hold promise for the development of more effective approaches.

In this paper, we took inspiration from the latest developments in the ML literature and proposed a novel model architecture called GLMMNet that targets high-cardinality categorical features in insurance data. The proposed GLMMNet fuses neural networks with GLMMs to enjoy both the predictive power of deep learning models and the transparency and statistical strength of GLMMs.

GLMMNet is an extension to the LMMNN proposed by \citet{simchoni2022}. The LMMNN is limited to the case of a Gaussian response with an identity link function. GLMMNet, on the other hand, allows for the entire class of ED family distributions, which are better suited to the modelling of skewed distributions that we see in insurance and financial data.

Importantly, in making this extension, we have made several modifications to the original LMMNN architecture, including:
\begin{arcitem}
\item
  \emph{The removal of a non-linear transformation on the high-cardinality categorical features \(Z\)}. This is to ensure the interpretability of random effect predictions, which carries practical significance (as discussed in Section~\ref{ssec:practical-insights}).
\item
  \emph{The shift from optimising the exact likelihood to a variational inference approach}. As discussed in Section~\ref{ssec:glmmnet-training}, the integral in the likelihood function does not generally have an analytical solution (with only a few exceptions, and the Gaussian example is one of those). The shift to a variational inference approach circumvents the complicated numerical approximations to the integral. This opens the door to a flexible range of alternative distributions for the response, including Bernoulli, Poisson, Gamma---any member of the ED family. This flexibility makes the GLMMNet widely applicable to many insurance contexts (and beyond).
\end{arcitem}

In our experiments, we found that the GLMMNet often performed better than or at least comparably with the benchmark model of an entity embedded neural network, which is the most popular approach among actuarial researchers working with neural networks. Although they can both be outperformed by boosting models in an environment of low signal-to-noise ratio, adding regularisation to the GLMMNet was sufficient to bring the model back to the top. The GLMMNet's outperformance over the entity embedded neural network is an achievement worth celebrating, as it comes with the additional benefits of transparency on the random effects as well as the ability to produce probabilistic predictions (e.g.~uncertainty estimates).

To sum up, the proposed GLMMNet has at least the following advantages over the existing approaches:
\begin{arcitem}
\item
  Improved predictive performance under most scenarios we have tested;
\item
  Interpretability with the random effects;
\item
  Flexibility with the form of the response distribution;
\item
  Ability to handle large datasets (due to the computationally efficient implementation through variational inference).
\end{arcitem}

All these make GLMMNet a worthy addition to the actuaries' modelling toolbox for handling high-cardinality categorical features.

\section*{Acknowledgements}
This work was presented at the 2022 Australasian Actuarial Education and Research Symposium (AAERS) in November 2022 (Canberra, Australia). The authors are grateful for the constructive comments received from colleagues present at the event.

This research was supported under Australian Research Council's Discovery Project DP200101859 funding scheme.  Melantha Wang acknowledges financial support from UNSW Australia Business School. The views expressed herein are those of the authors and are not necessarily those of the supporting organisations. 

The authors declare that they have no conflicts of interest.

\section*{Data and Code}
The code used in the numerical experiments, as well as the simulation datasets, is available on \url{https://github.com/agi-lab/glmmnet}.


\bibliographystyle{elsarticle-harv}
\bibliography{libraries}

\newpage
\appendix
\counterwithin{figure}{section}
\counterwithin{table}{section}
\section{GLMMNet Implementation} \label{app:practical}
Below we outline a few practical considerations that arise from the implementation of GLMMNet.

\subsection{Fixed Prior}
The random effects layer of GLMMNet (Section \ref{sssec:re}) requires the specification of a prior distribution. We mentioned that we would use a Gaussian prior as per common practice with GLMMs, but we did not specify the exact parameters for it. With GLMMNet, it is possible to have a trainable prior whose parameters can be found by gradient descent with respect to the loss function in \eqref{eq:glmm-ELBO}; this approach of estimating the prior from the data is known as \emph{empirical Bayes} \citep{casella1985}. We experimented with both fixed and trainable priors in our implementation. We found that having a trainable prior leads to worse performance in general. \citet{blundell2015} also reached the same conclusion from their experiments with Bayesian neural networks and they speculated that the algorithm would be more tempted to update the prior parameters than the posterior when the prior were trainable. We choose to work with fixed priors.

As to what values to use for the fixed prior, we reference the Stan documentation by \citet{gelman2020}, which recommends the use of weakly informative priors, i.e.~priors that will give way to the likelihood in the presence of sufficient data but will dominate in the absence of data. The exact prior parameters to use will depend on the data and the task at hand.

\subsection{Reparametrisation of variational parameters}
We use the softplus bijector function, defined as \(\sigma(\lambda) = \log(1+\exp(\lambda)), ~\lambda \in \mathbb{R}\), to reparametrise the scale parameters (\(\sigma_j, ~j=1, \cdots, q\)) in the diagonal Gaussian distribution (i.e.~the surrogate posterior). This ensures that the surrogate scale parameters always stay within their support (i.e.~in the positive region).

Furthermore, we found that it is important to shrink the value of the \(\sigma_j\)'s at the start of the learning process to help guide the algorithm in the right direction. In our implementation, this is achieved by adding a constant multiplier (e.g.~0.01) to the parametrisation of \(\sigma_j\)'s. In the absence of such a constraint, the algorithm tends to return unreasonably large \(\sigma_j\) values, which significantly deteriorates the performance of GLMMNet. It appears that the unguided GLMMNet converges to strange local minima where the model attributes all the variation to the noise.

We note that, in theory, adding a constant multiplier as we did does not modify the solution space and thus should return the same results regardless. We were able to empirically verify that if we gave the model enough training time, it was able to figure out the right range of values for the parameters at the end. The tweak we made, however, helps the algorithm immediately converge to a better local minimum. One possible explanation is that by reducing the size of the gradients on the scale parameters, it helps the model focus on learning the more important location (mean parameters) of the posterior random effects.

\subsection{Initialisation of trainable parameters}
The network weights and biases in GLMMNet are initialised by Tensorflow's default Glorot uniform initialiser \citep{glorot2010}. The dispersion parameter for the ED family, which is also learned as part of GLMMNet, is initialised with a fixed estimate roughly calibrated to the data at hand. It does not seem to have a major impact on GLMMNet's performance, but we still consider it good practice to run some small scale experiments with this choice of initial value.

\clearpage
\section{Notes on Numerical Experiments}
\subsection{GLMM Encoding} \label{app:glmm-enc}
GLMM encoding can be regarded as an easier and more flexible alternative to the machine learning mixed models---e.g.~GPBoost \citep{sigrist2021,sigrist2022} or GLMMNet, which often present a convoluted estimation procedure. While simple enough, \citet{pargent2022} found that this encoding scheme performed very well across a range of prediction tasks and a variety of datasets.

In the numerical experiments, we implemented a cross-validated version of GLMM encoding, as presented in Algorithm \ref{alg:GLMM-encoding}. The encoded values \(\mathbf{z}'\) then take the place of the original categorical feature to enter any subsequent ML model, e.g.~a gradient boosting machine.

\begin{algorithm}[htbp!]
\caption{GLMM Encoding, adapted from Pargent et al. (2022)}\label{alg:GLMM-encoding}
\KwData{$\mathcal{D}=(y_i, \mathbf{z}_i)_{i=1}^n$}
\KwResult{$\mathbf{z}'=\psi(\mathbf z) \in \mathbb{R}^n$, a numeric representation of the categorical feature $\mathbf{z}$}
Randomly partition $\mathcal{D}$ into $K$ subsets of equal size $\mathcal{D} = \{\mathcal{D}_1, \mathcal{D}_2, \cdots, \mathcal{D}_K\}$\;
\For{$k = 1$ to $K$}{
  Fit a random intercept model $\boldsymbol \eta=\beta_0 \boldsymbol{1}+\mathbf{Zu}$ with $\mathbf{u}\sim \mathcal{N}(\mathbf{0}, \sigma^2_u \mathbf{I})$ on $\mathcal{D}_k^\text{train} =\mathcal{D} \backslash \mathcal{D}_k$\;
  \For{$\{y_i, \mathbf{z}_i\} \in \mathcal{D}_k$}{
    \eIf{category of the $i$-th observation $j[i]$ is in $\mathcal{D}_k^\text{train}$}{
      Predict $z'_i = \hat{y}_i= g^{-1}(\beta_0+u_{j[i]})$\;
    }{
      Predict $z'_i = \hat{y}_i= g^{-1}(\beta_0)$\;
    }
  }
}
\end{algorithm}

\subsection{Neural Networks} \label{app:nn}
Training a network involves making many specific choices. Below we briefly describe the specifications we use for the two network models, \texttt{NN\_ee} and \texttt{GLMMNet}. To allow a fair comparison, where possible, we keep those choices consistent across both networks.
\begin{arcitem}
\item
  \emph{Architecture for the hidden layers}. We choose to use three hidden layers and \([64, 32, 16]\) hidden units (neurons) in each layer. This choice was made following the recommendation of \citet{ferrario2020}, where the authors suggested that the first hidden layer should be large enough to allow new features to be constructed from the raw input covariates and that successive layers should compress information. Some quick experimentation confirms that this choice of the hidden units works relatively well.
\item
  \emph{Activation functions}. We use the ReLU function defined by \(f(x) = \max(x, 0)\) for hidden layer activations and the inverse of the link function for final layer activation (see Section \ref{sssec:rd}).
\item
  \emph{Optimiser}. For both networks we use Tensorflow's default Adam optimiser with learning rate 0.001 \citep{kingma2014}, which is a state-of-the-art optimisation algorithm with demonstrated superior performance for a range of predictive tasks. This has also been used in many actuarial applications, e.g. \citet{richman2021d}.
\item
  \emph{Early stopping}. In fitting the networks, we further split the training data into an inner-training set and a validation set. We decide the number of epochs to train the networks based on when the validation performance (as captured by the validation loss) stops improving for a fixed number of epochs.
\item
  \emph{Loss function}. Section \ref{ssec:glmmnet-training} discusses in detail the loss function we use to optimise \texttt{GLMMNet}. For \texttt{NN\_ee}, we follow standard practice and use the squared error loss function to optimise the network. While it is possible to design a likelihood-based loss function for \texttt{NN\_ee}, this involves further complications (such as reparametrisation of the dispersion parameters) above the scope of this project.
\item
  \emph{Embedding dimension} (for \texttt{NN\_ee} only). The dimension \(d\) of the embedding space is a network hyperparameter that should be chosen through experimentation. \citet{lakshmanan2020} suggest to use ``the fourth root of the total number of unique categorical elements'' (p.48), which we used as a reference.
\end{arcitem}

\subsection{Learning Pipeline}
As per common practice, the data are split into training and testing sets. The training set is used for learning (i.e. fitting of the models) and the testing set is used for assessing and comparing the performance of the optimised candidate models.

In this work, we only perform a minimal search for hyperparameters, as hyperparameter tuning is not the main purpose of this study. We use the validation set approach. For models that involve hyperparameters, we split the training set into an inner-training set and a validation set. Models are fitted on the inner-training set, and hyperparameters are selected based on their performance on the validation set. This approach is less systematic than cross validation, but from experience, it usually yields reasonably similar results at a much more sustainable computational cost.

\subsection{Simulation Environments} \label{app:data-gen}
\begin{figure}[htbp]
\begin{center}
\subfloat[Balanced\label{fig:cat-dist-1}]{\includegraphics[width=0.46\linewidth]{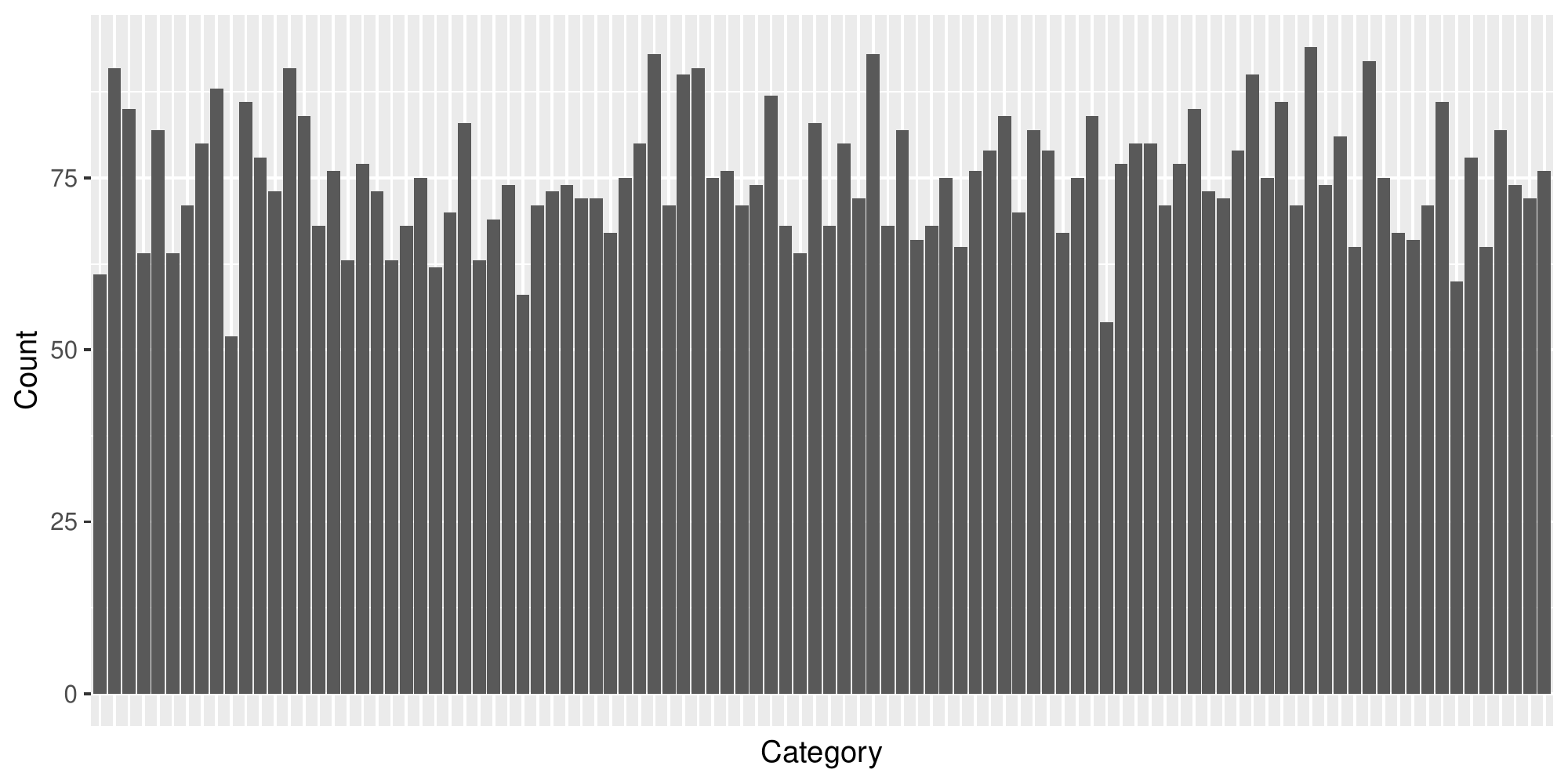} }\subfloat[Skewed\label{fig:cat-dist-2}]{\includegraphics[width=0.46\linewidth]{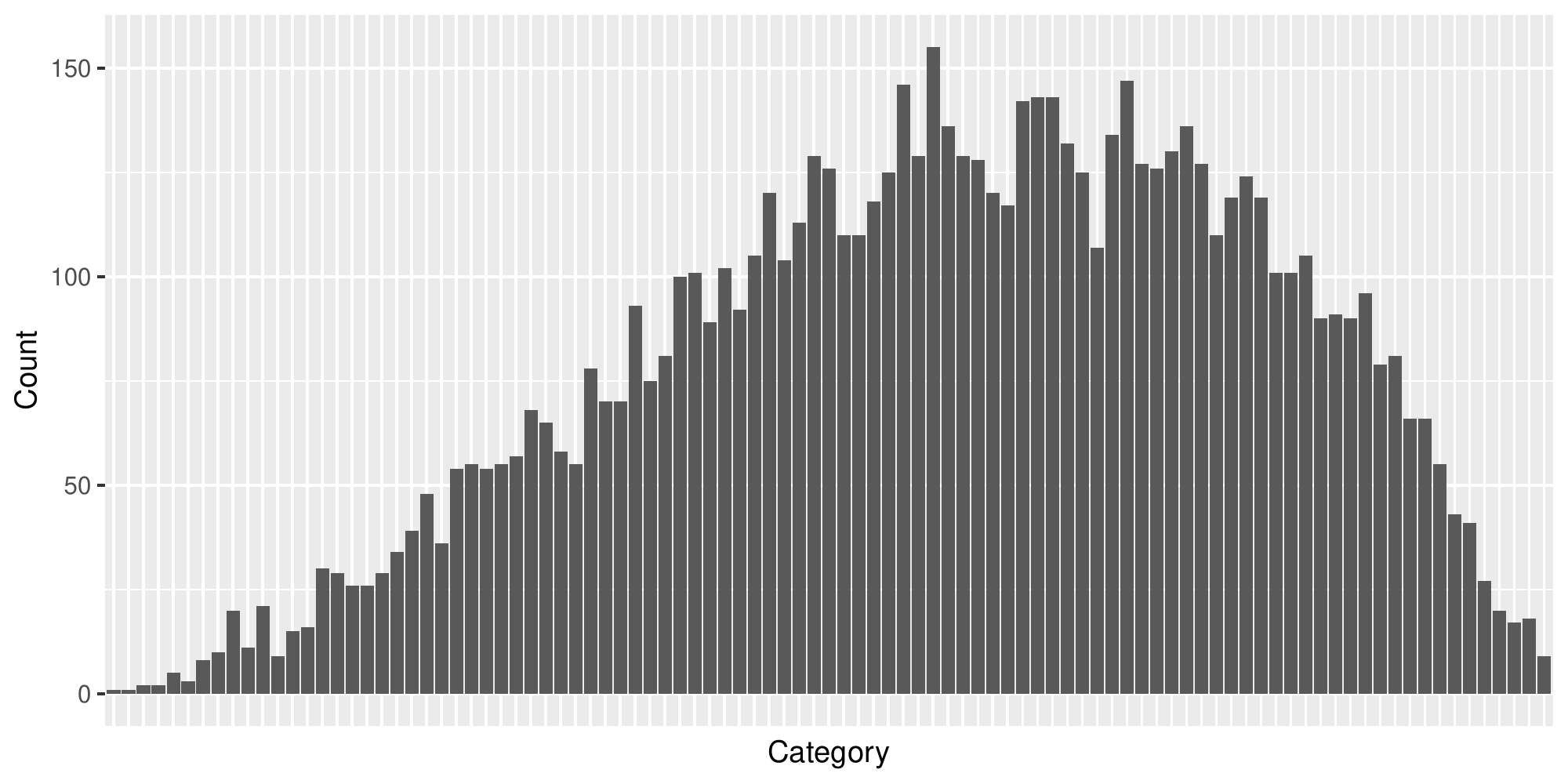} }
\caption{Number of observations by category for balanced \textit{versus} skewed distribution of categories}\label{fig:cat-dist} 
\end{center}
\end{figure}

We set up the desired simulation environments in Section \ref{sec:simulation} by adjusting the following parameters:
\begin{arcitem}
\item
  \texttt{n\_categories}: number of categories; fixed at 100.
\item
  \texttt{signal\_to\_noise}: a three-dimensional vector that captures the relative ratio of signal strength (as measured by \(\mu_f\), the mean of \(f(\mathbf{x})\)), random effects variance (\(\sigma_u^2\)), and variability of the response (\(\sigma_\epsilon^2\); noise, or equivalently the irreducible error, as this component captures the unexplained inherent randomness in the response). The vector will be normalised to sum to 1, e.g.~when \(\texttt{signal\_to\_noise} = [3, 1, 1]^\top\), the vector is first normalised to \([0.6, 0.2, 0.2]^\top\). Data points are generated as follows:
  \begin{arcenum}
  \item
    Generate a sample \(\mathbf{u} \in \mathbb{R}^q\) from \(u_j \stackrel{iid}{\sim} \mathcal{N}(0, \sigma_u^2 = 0.2^2)\).
  \item
    Rescale \(f(\mathbf{X})\), which has been pre-calculated from a deterministic formula (e.g.~the Friedman function in \eqref{eq:friedman-1}), such that \(\overline{f(\mathbf{X})} = \mu_f = 0.6\) where \(\overline{x}\) denotes the sample mean across all observations \(i = 1, \cdots, n\).
  \item
    Set the conditional mean as \(\boldsymbol{\mu} = \mathbb{E}(\mathbf{y}|\mathbf{u}) = g^{-1}(f(\mathbf{X}) + \mathbf{Z}^\top \mathbf{u})\) and \(\phi = \sigma_\epsilon^2 = 0.2^2\), where \(\phi\) is the dispersion parameter for the ED family, such that \(\operatorname{Var}(\mathbf{y}|\mathbf{u}) = \phi V(\boldsymbol{\mu})\) where \(V(\cdot)\) is the variance function for the family.
  \item
    Generate samples \(\mathbf{y}|\mathbf{u}\) from the conditional mean \(\boldsymbol{\mu}\) and conditional variance \(\operatorname{Var}(\mathbf{y}|\mathbf{u})\).
  \end{arcenum}
\item
  \texttt{y\_dist}: distributional assumption for the response, e.g.~Gaussian, gamma, or any other member of the ED family.
\item
  \texttt{inverse\_link}: inverse of the link function, i.e.~\(g^{-1}(\cdot)\).
\item
  \texttt{cat\_dist}: to use balanced or skewed distribution for the allocation of categories. A ``balanced'' distribution allocates approximately equal number of observations to each category; a ``skewed'' distribution generates categories from a (scaled) beta distribution; see Figure \ref{fig:cat-dist}.
\end{arcitem}

\vspace{-0.4cm}

\subsection{Description of the SME Building Insurance Data}

\subsubsection{Variable List} \label{app:predictors}
\begin{table}[H]
\begin{center}
\begin{footnotesize}
\begin{tabular}{p{3cm}p{11cm}}
\toprule
\textbf{Variable} & \textbf{Description}  \\ \midrule
\verb|anzsic4_desc| & ANZSIC occupation code description that describes the business's economic activity (at the Class level).  \\
\verb|year_incurred| & Calendar year of the accident associated with the claim; value between 2010 and 2015. \\
\verb|peril| & Cause of the claim, e.g. impact by third party, malicious damage, or fire. \\
\verb|years_insured| & Tenure of the policy. \\
\verb|state_risk| & State of the insured's business premises. \\
\verb|occupancy| & Type of occupancy of the insured's business premises, e.g. property owner or tenant. \\
\verb|locality| & Location of the insured's business premises, e.g. industrial or retail.\\
\verb|roof_type| & Primary roof material of the insured's business premises; re-processed into a binary indicator of whether the material is fire resistant or not. \\
\verb|wall_type| & Primary wall material of the insured's business premises. \\
\verb|floor_type| & Primary floor material of the insured's business premises. \\
\verb|log_si| & Log of total sum insured for building and contents. \\
\verb|firep_detect| & A binary indicator of whether the insured's business premises have any fire detection equipment, e.g. a fire alarm.  \\
\verb|firep_stop| & A binary indicator of whether the insured's business premises have any fire suppression equipment, e.g. a fire extinguisher. \\
\bottomrule
\end{tabular}
\end{footnotesize}
\caption{(Engineered) input features used in modelling} \label{tab:predictors} \vspace{-5mm}
\end{center}
\end{table}

\subsubsection{ANZSIC Example} \label{app:anzsic}
\begin{table}[H]
\begin{center}
\begin{small}
\begin{tabular}{@{}lll@{}}
\toprule
\textbf{Division}    & K    & Financial and Insurance Services   \\ \midrule
\textbf{Subdivision} & 63   & Insurance and Superannuation Funds \\ \midrule
\textbf{Group}       & 632  & Health and General Insurance       \\ \midrule
\textbf{Class}       & 6322 & General Insurance                  \\ \bottomrule
\end{tabular}
\caption{An example of ANZSIC occupation classification}
\label{tab:anzsic} \vspace{-5mm}
\end{small}
\end{center}
\end{table}

\clearpage
\section{Supplementary Results} \label{app:results}
\subsection{Experiments 2--3: Increasing Complexity within a Low Noise Environment} \label{app:sim23}
\begin{figure}[ht]
{\centering \subfloat[Exp 2, CRPS\label{fig:boxplots-2-1}]{\includegraphics[width=0.49\linewidth]{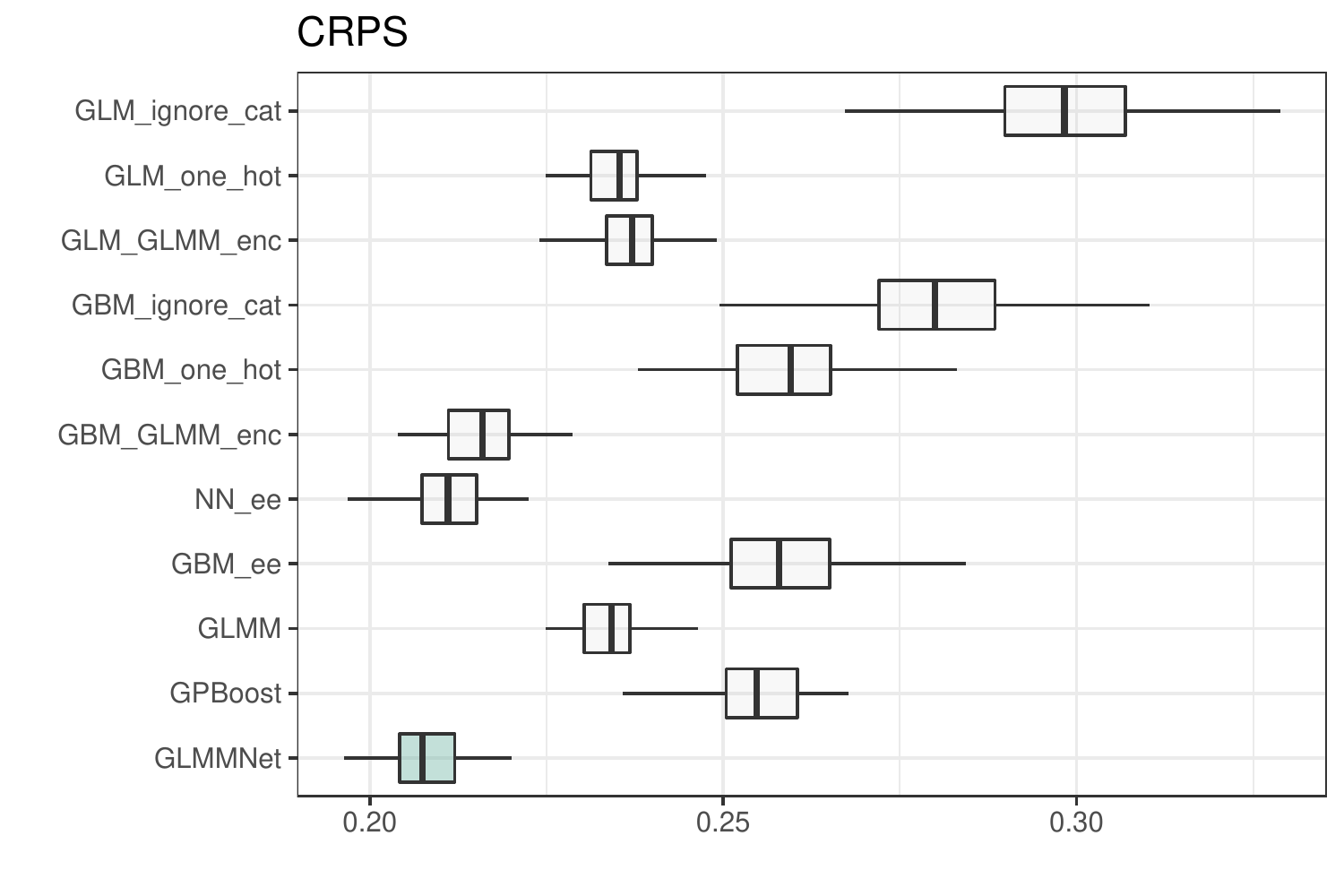} }\subfloat[Exp 2, RMSE of average prediction per category\label{fig:boxplots-2-2}]{\includegraphics[width=0.49\linewidth]{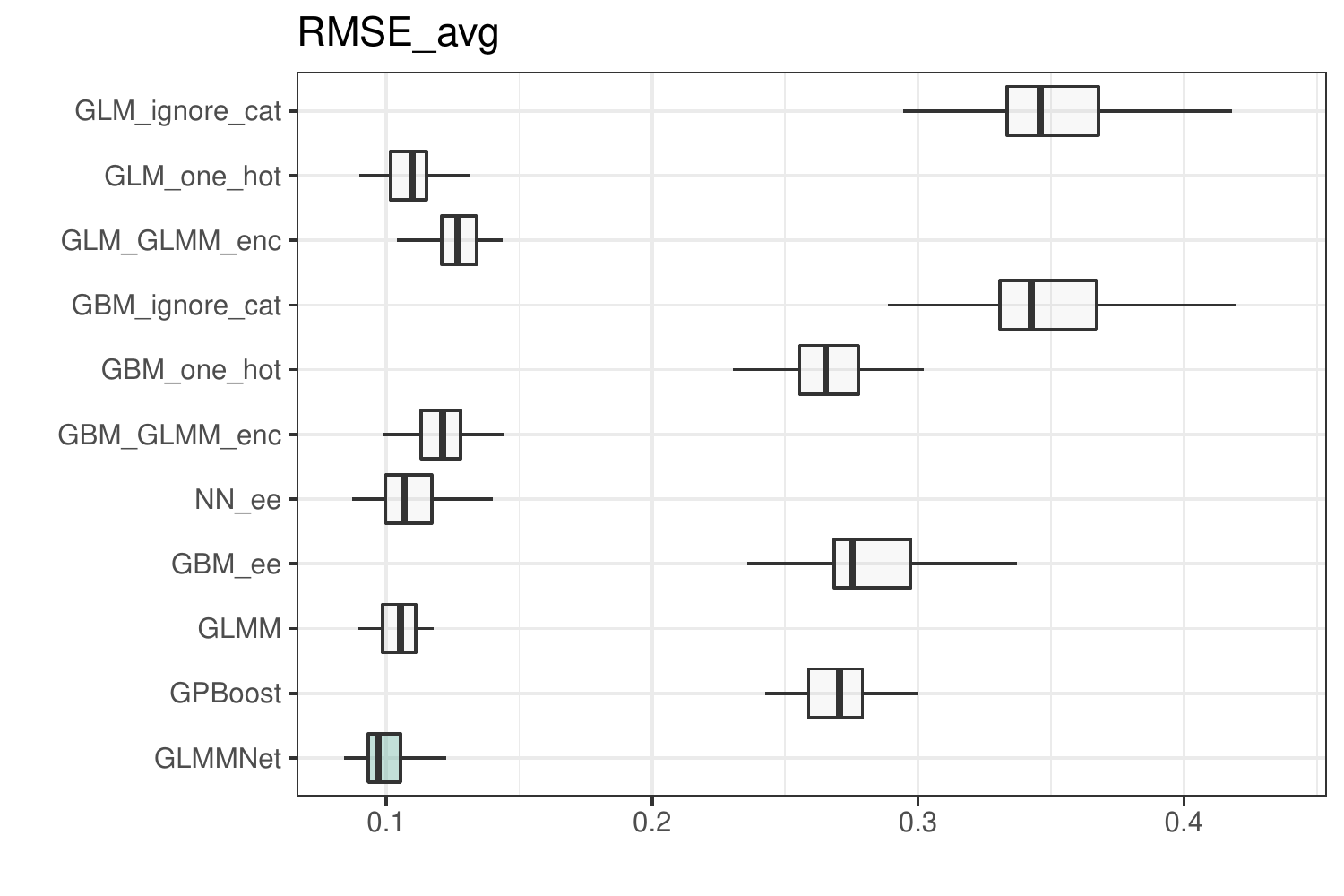} }\newline\subfloat[Exp 3, CRPS\label{fig:boxplots-2-3}]{\includegraphics[width=0.49\linewidth]{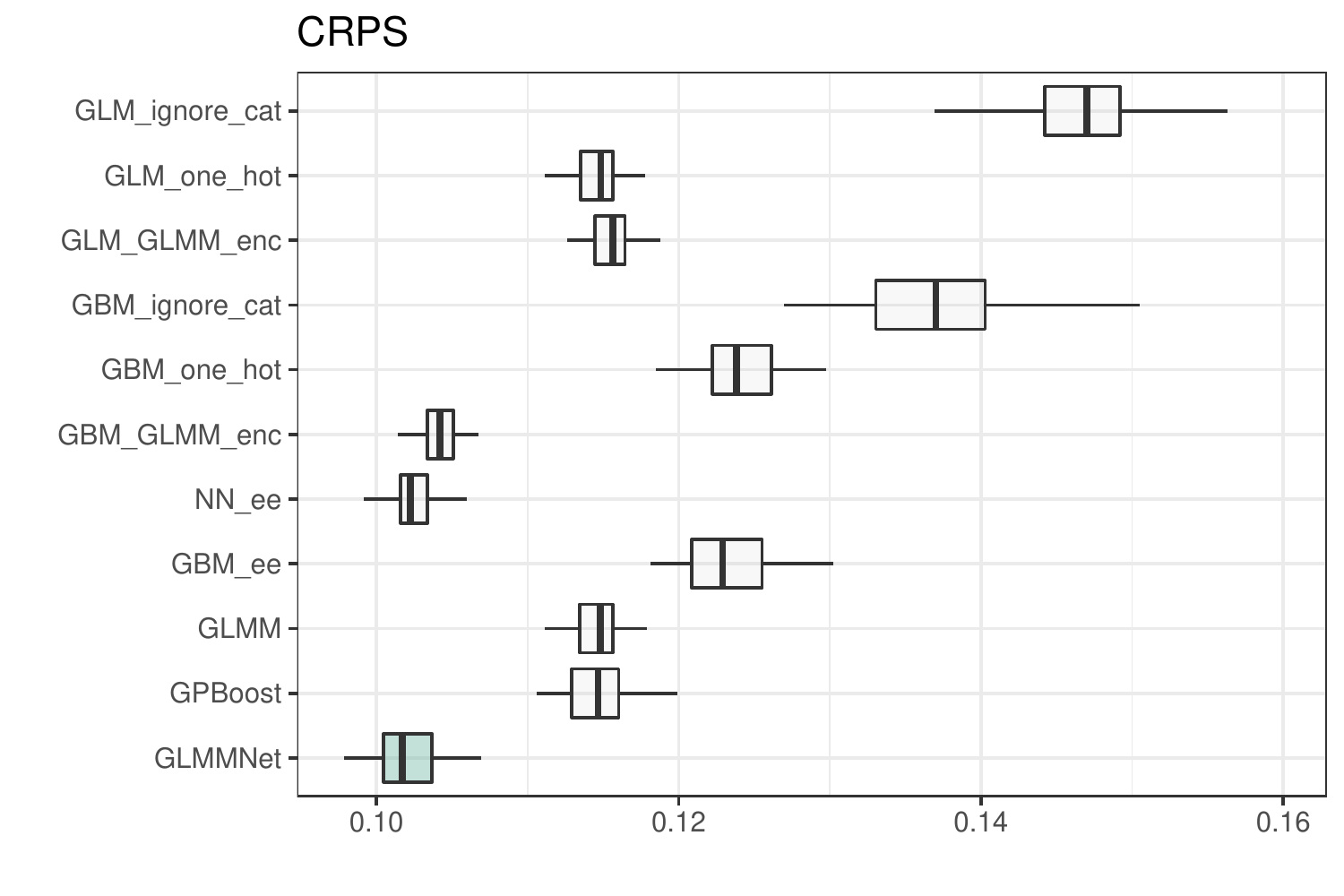} }\subfloat[Exp 3, RMSE of average prediction per category\label{fig:boxplots-2-4}]{\includegraphics[width=0.49\linewidth]{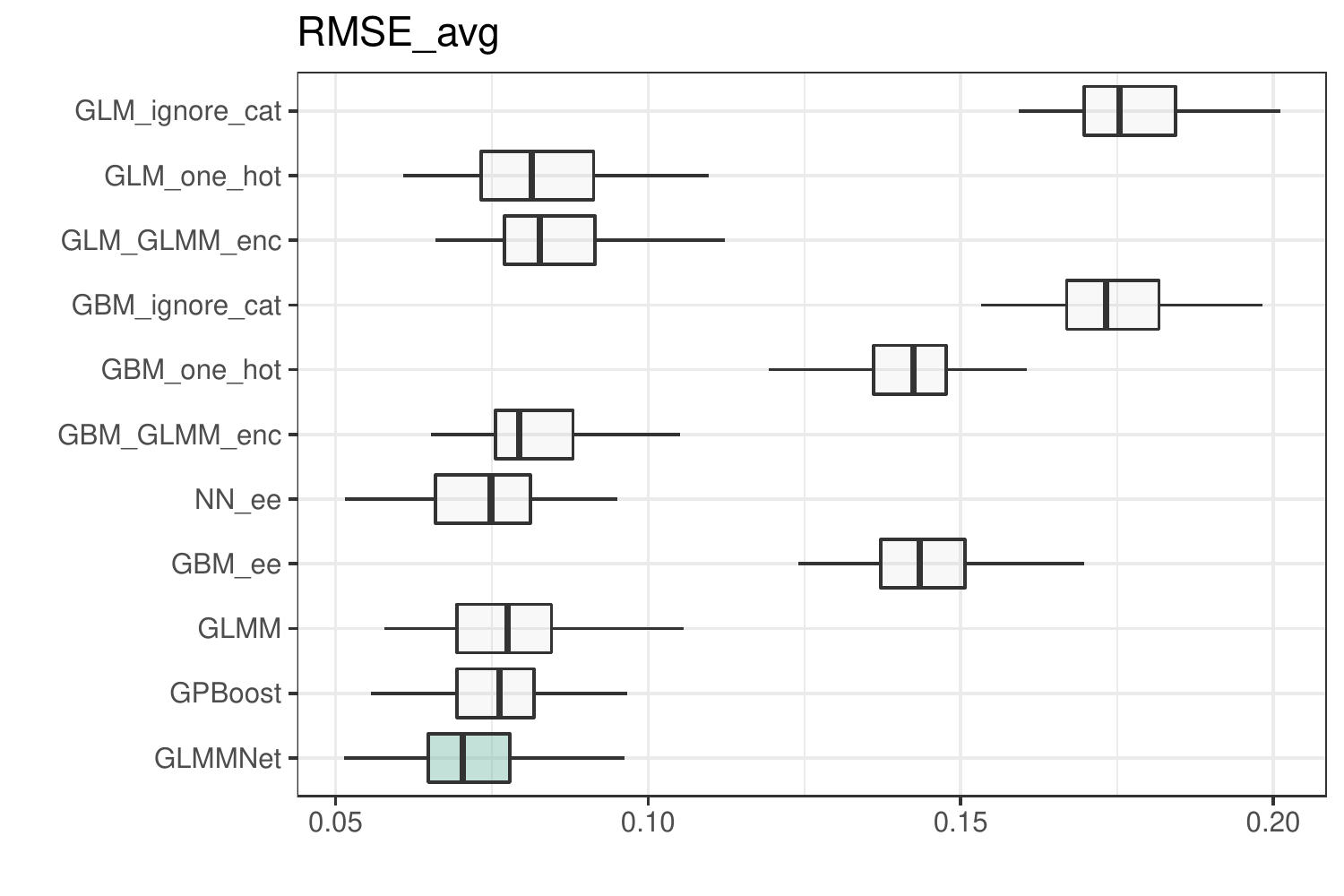} }
}
\caption{Boxplots of out-of-sample performance metrics of the different models in experiment 2 (top; gamma-distribued response) and experiment 3 (bottom; skewed distribution of categories); GLMMNet highlighted in green. Each experiment is repeated 50 times, with 5,000 training observations and 2,500 testing observations each.}\label{fig:boxplots-2}
\end{figure}

Figure \ref{fig:boxplots-2} displays the out-of-sample performance metrics of the candidate models from experiments 2 and 3, which respectively consider a gamma-distributed response variable and a skewed distribution of categories (see Table \ref{tab:scenarios}). Note that we chose not to show in Figure \ref{fig:boxplots-2} the MAE or RMSE plots for brevity of presentation; the results from the two metrics are in perfect alignment with the CRPS results.

\subsection{Experiments 4--6: Dialling up the Noise} \label{app:sim46}

In the remaining three experiments, we test the stability of GLMMNet under increasing noise in the environment, where ``noise'' refers to the unmodellable inherent uncertainty in the response due to its stochastic nature. Experiments 5--6 share the same signal-to-noise ratio of \([8, 1, 4]\) which is derived from the estimated parameters in the real data case study in Section~\ref{sec:case-study}. Experiment 4 has a signal-to-noise ratio of \([4, 1, 2]\), such that the signal strength relative to the random noise is also at \(2:1\) but the random effects are stronger (more variable) than in the case of experiments 5--6.

\begin{figure}[htbp!]
{\centering \subfloat[Exp 4, CRPS\label{fig:boxplots-4-1}]{\includegraphics[width=0.49\linewidth]{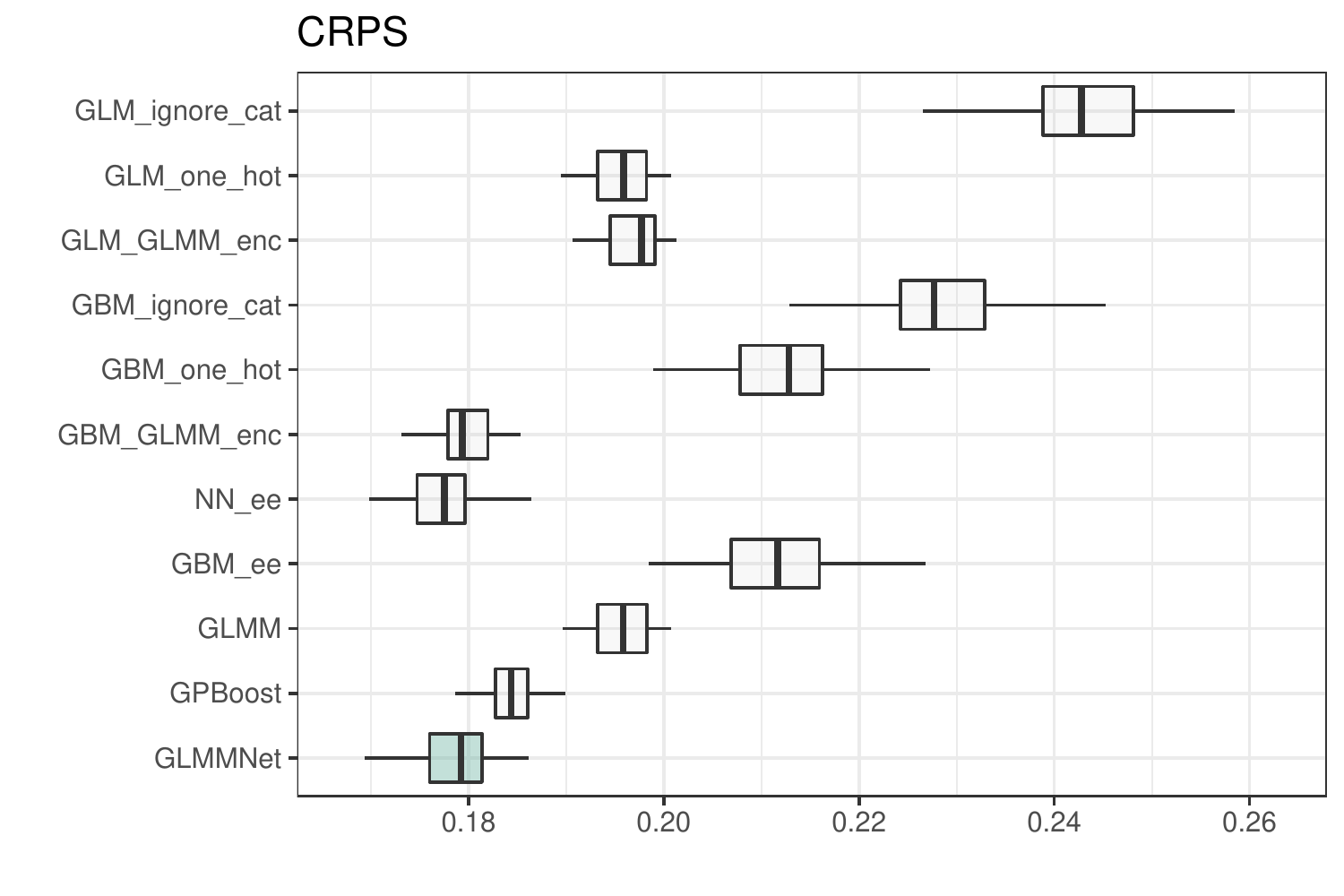} }\subfloat[Exp 4, RMSE of average prediction per category\label{fig:boxplots-4-2}]{\includegraphics[width=0.49\linewidth]{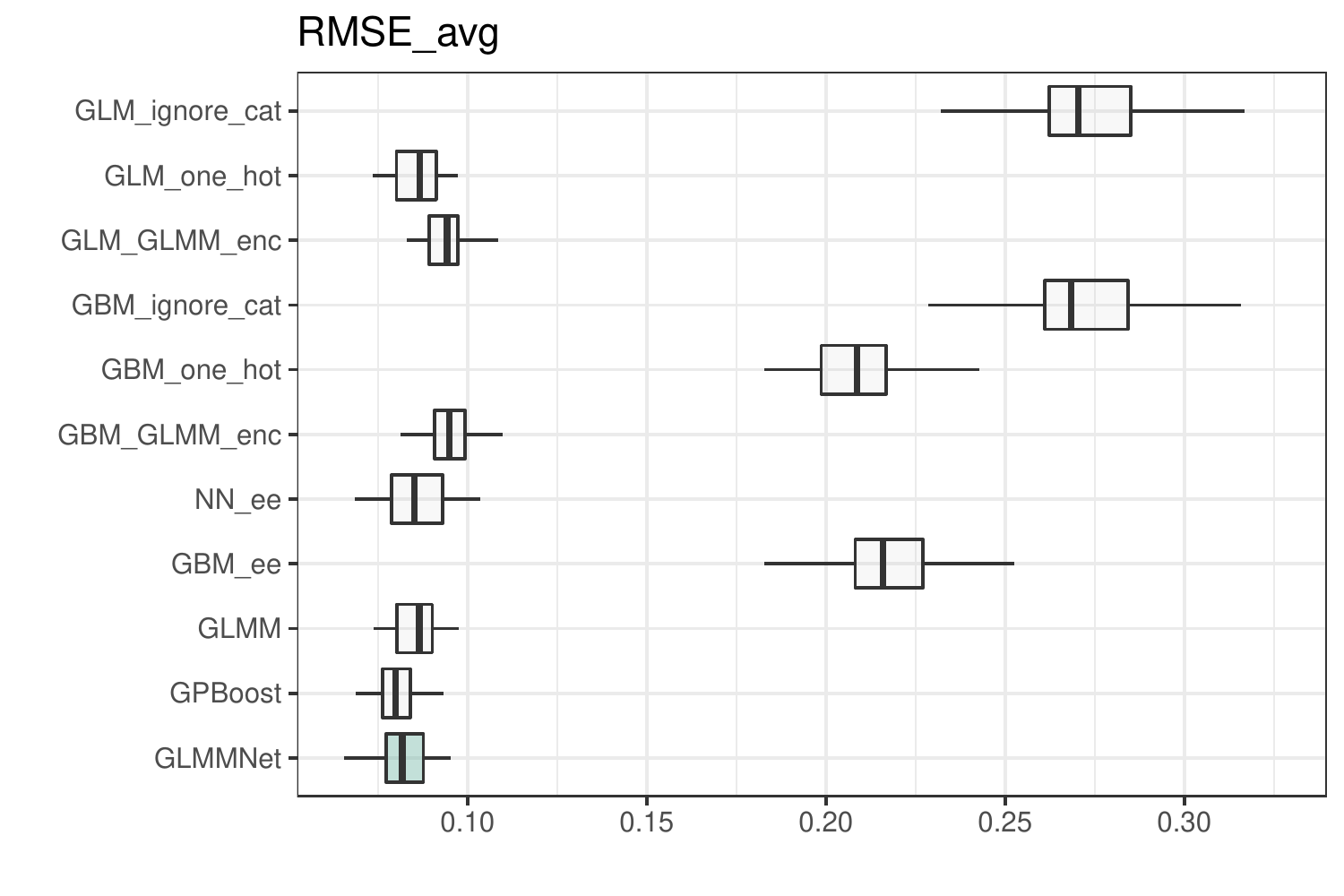} }\newline\subfloat[Exp 5, CRPS\label{fig:boxplots-4-3}]{\includegraphics[width=0.49\linewidth]{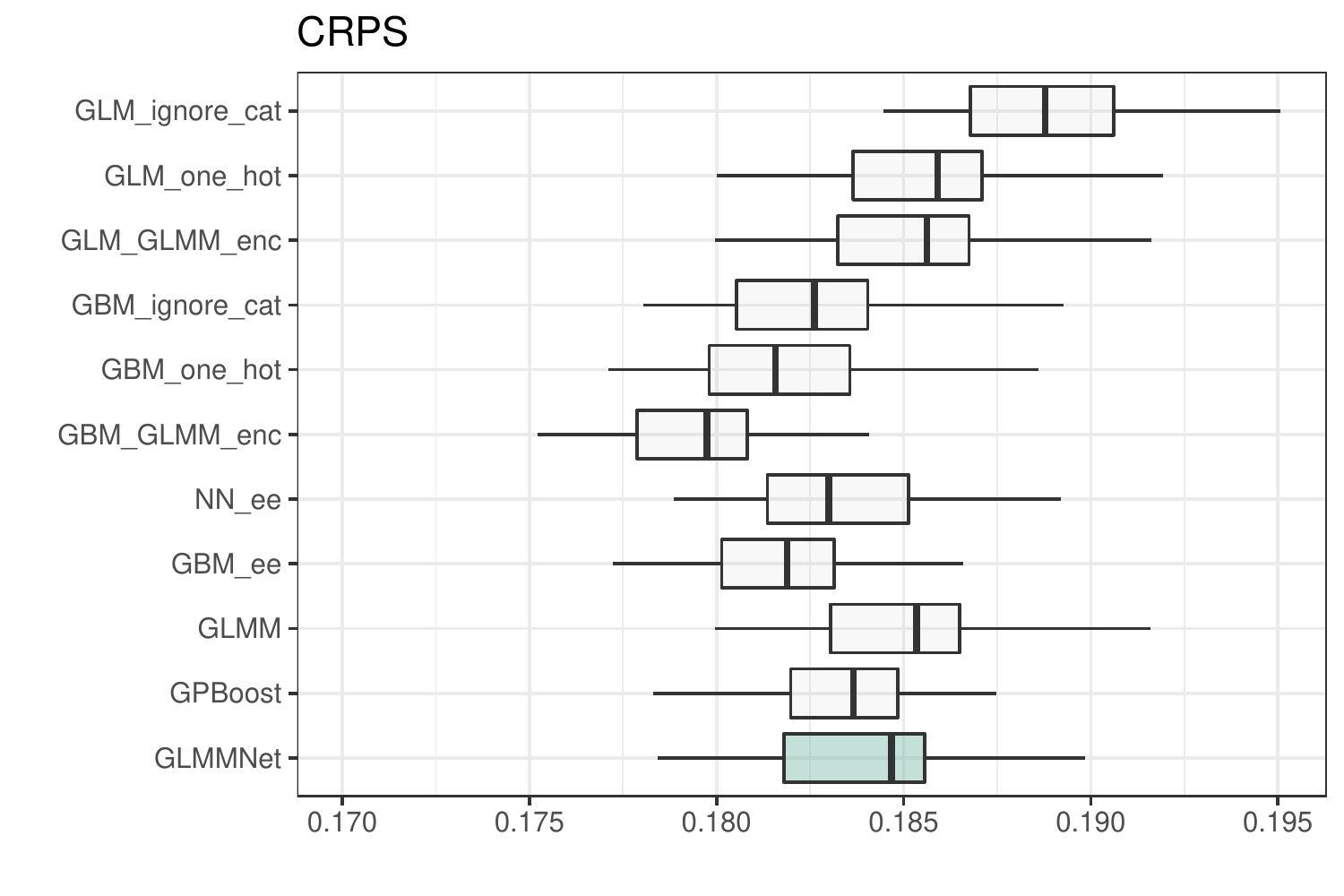} }\subfloat[Exp 5, RMSE of average prediction per category\label{fig:boxplots-4-4}]{\includegraphics[width=0.49\linewidth]{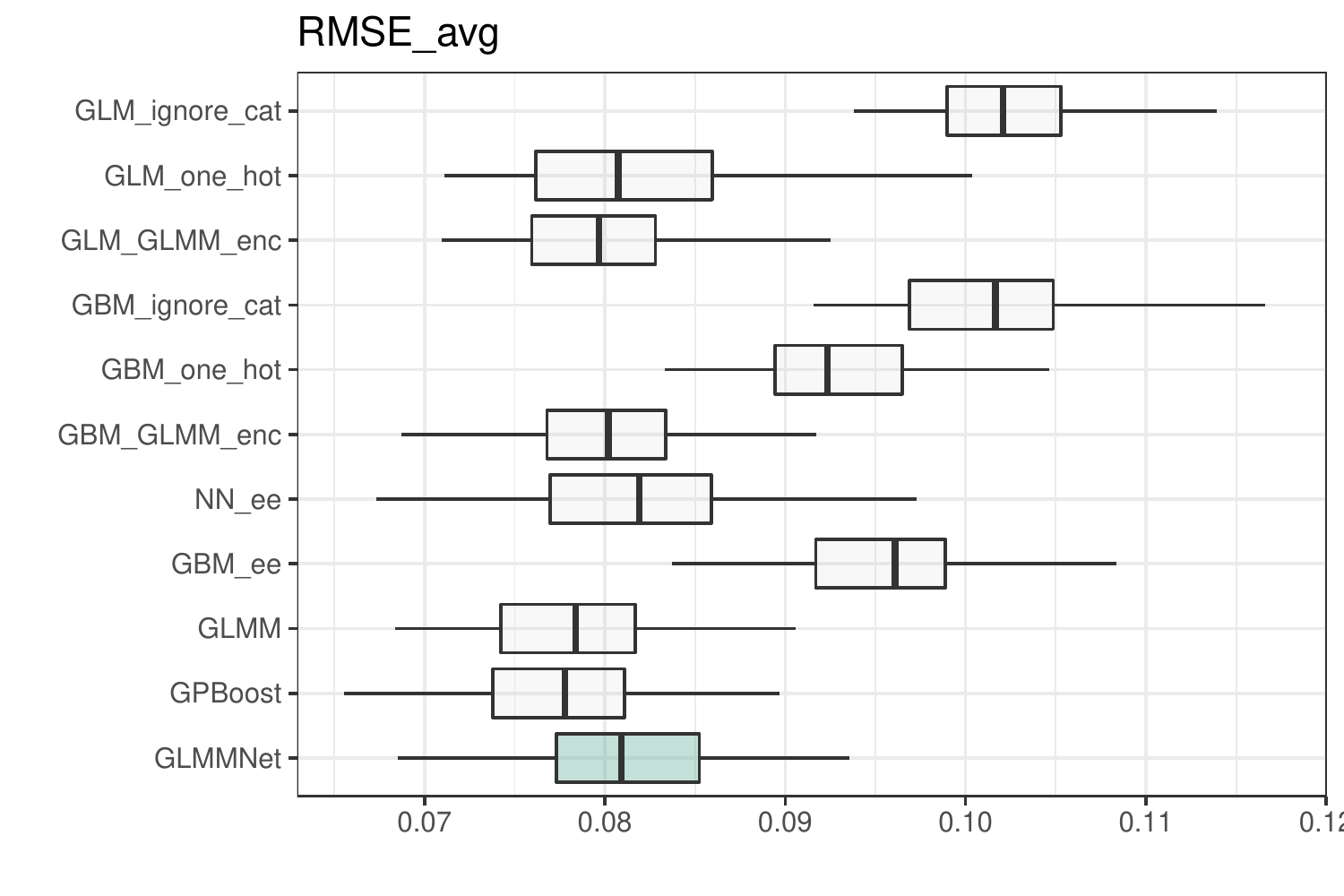} }\newline\subfloat[Exp 6, CRPS\label{fig:boxplots-4-5}]{\includegraphics[width=0.49\linewidth]{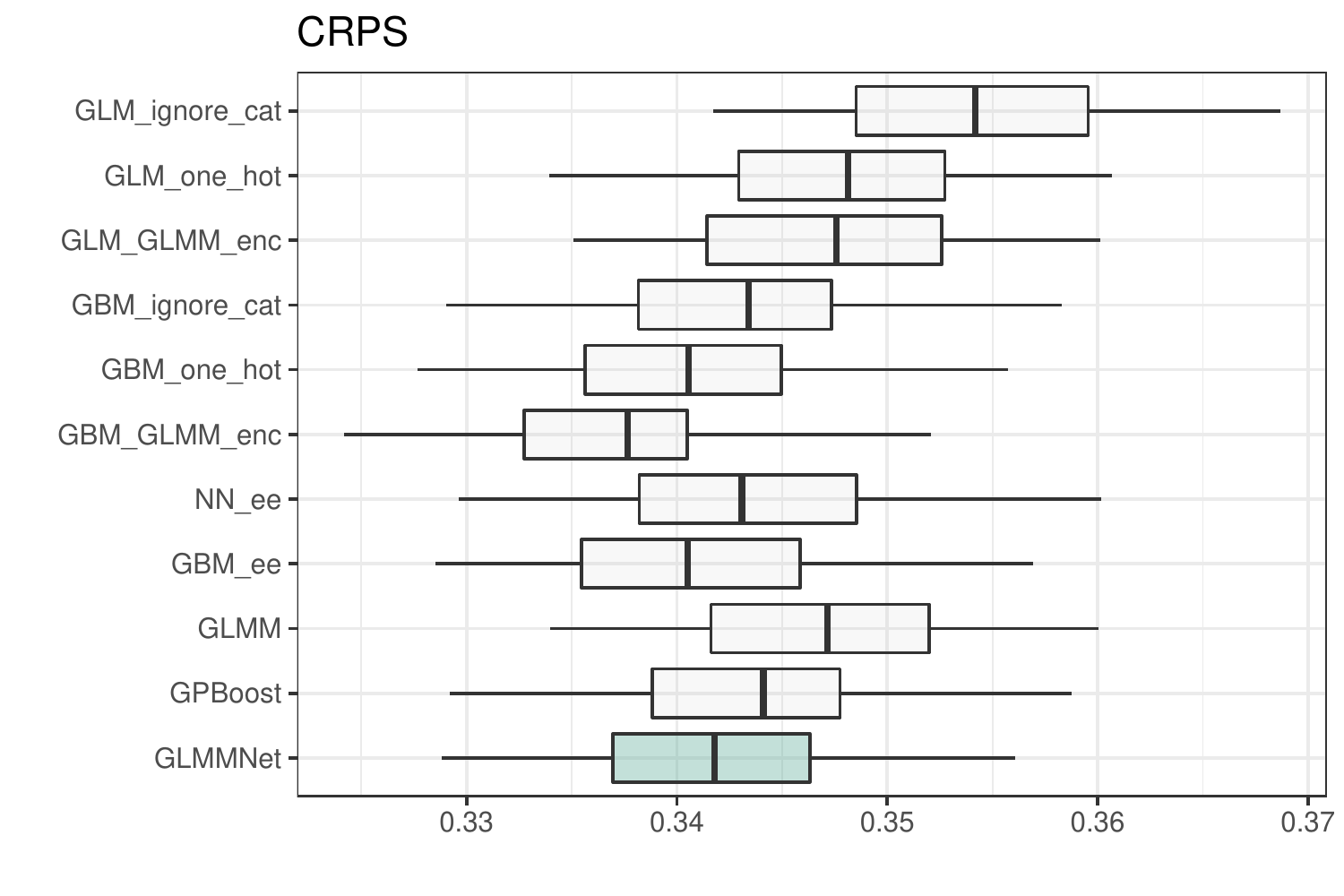} }\subfloat[Exp 6, RMSE of average prediction per category\label{fig:boxplots-4-6}]{\includegraphics[width=0.49\linewidth]{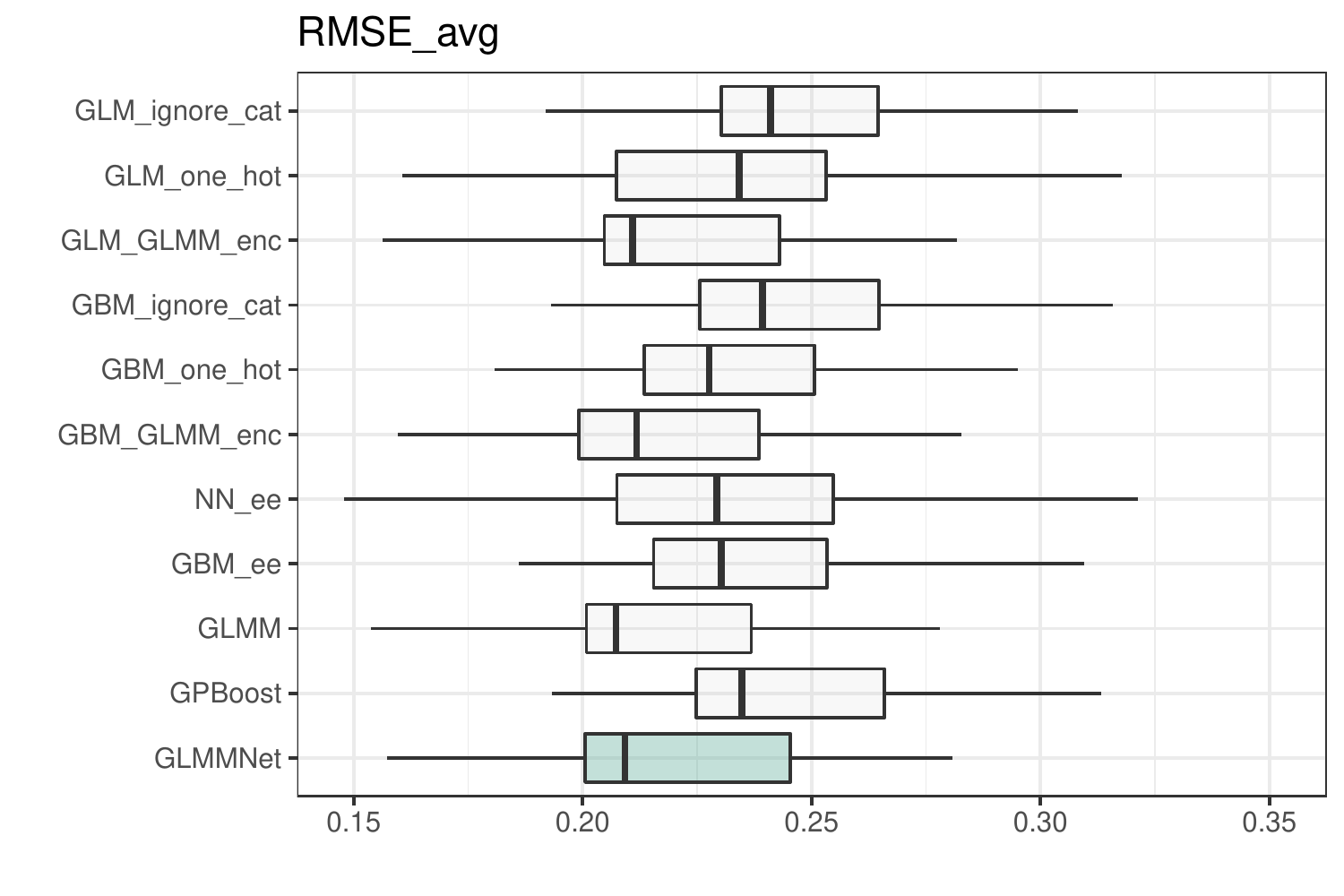} }
}
\caption{Boxplots of out-of-sample performance metrics of the different models in experiment 4 (top; medium noise Gaussian), experiment 5 (middle; high noise Gaussian) and experiment 6 (bottom; high noise, skewed categorical distribution, Gamma response); GLMMNet highlighted in green. Each experiment is repeated 50 times, with 5,000 training observations and 2,500 testing observations each.}\label{fig:boxplots-4}
\end{figure}

\begin{figure}[htbp!]
\begin{center}
\subfloat[Experiment 1\label{fig:RE-predictions-1}]{\includegraphics[width=0.35\linewidth]{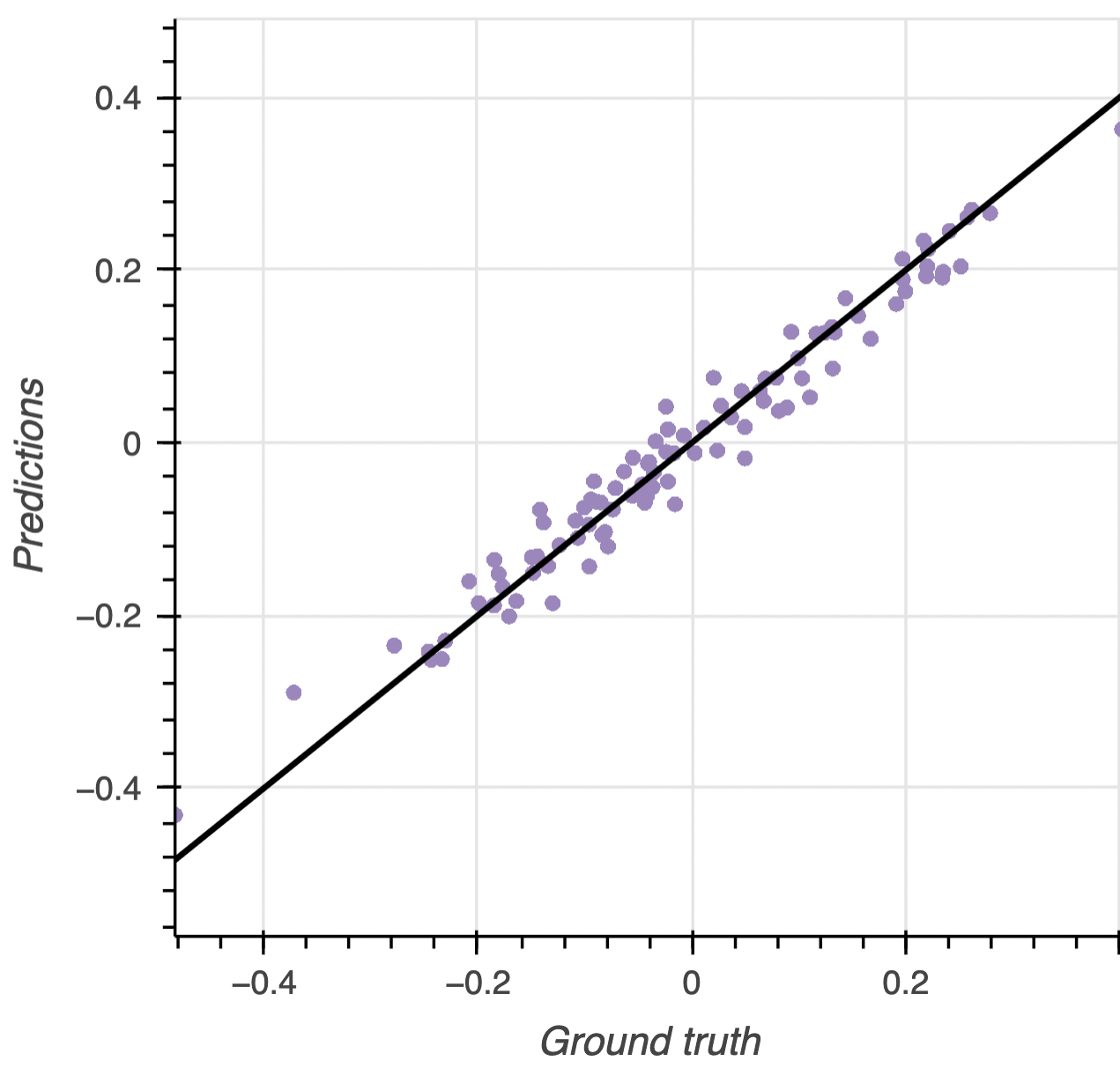} }
\subfloat[Experiment 6\label{fig:RE-predictions-6}]{\includegraphics[width=0.35\linewidth]{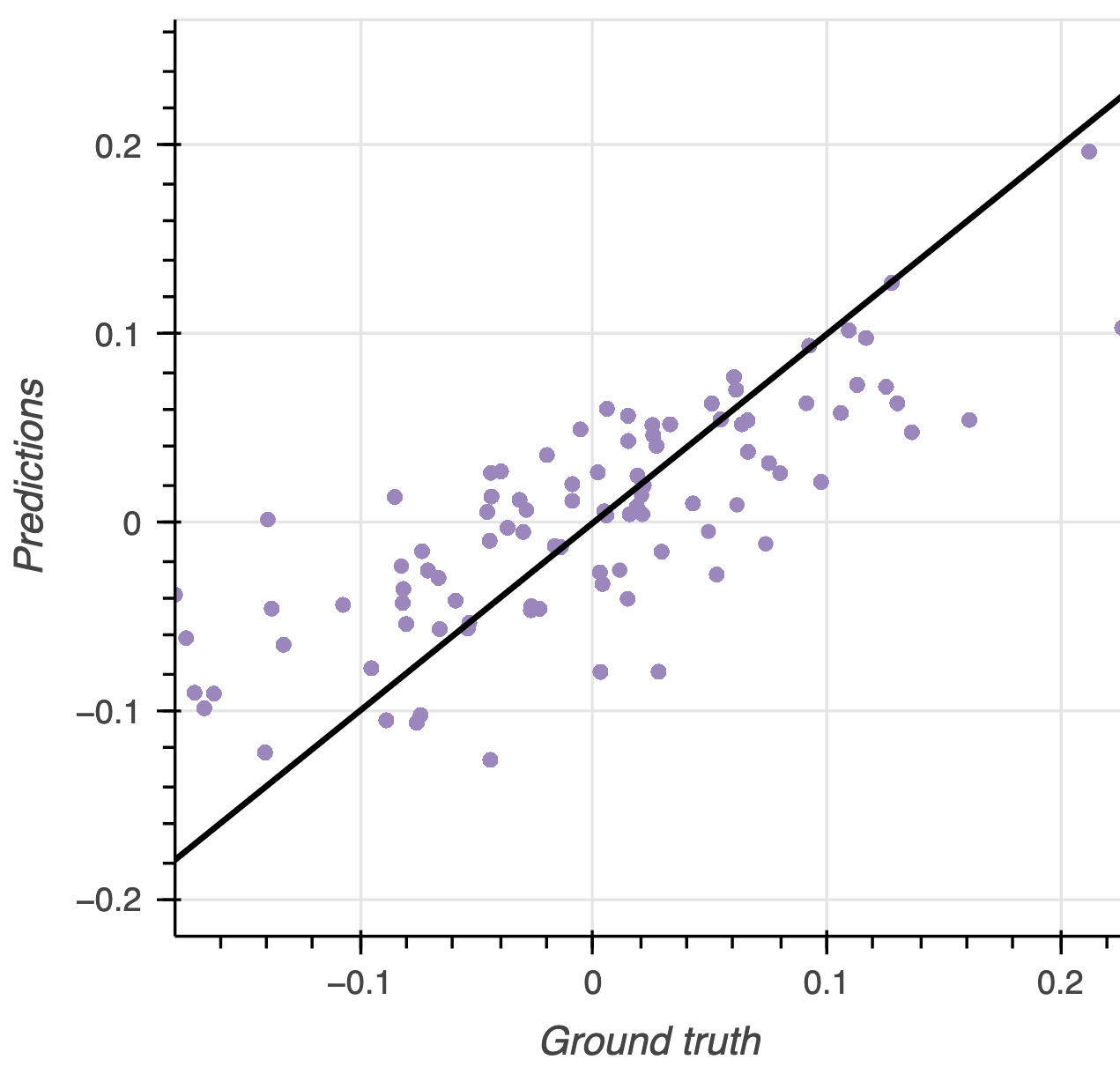} }
\caption{Random effects predictions by GLMMNet (posterior mode) against the ground truth under (a) experiment 1, i.e.~base scenario, (b) experiment 6, i.e.~high complexity, high noise scenario}\label{fig:RE-predictions}
\end{center}
\end{figure}

Figure \ref{fig:boxplots-4} plots the out-of-sample metrics for each of the three experiments. The results are discussed in Section~\ref{ssec:sim-results}. In general, as the noise level increases, the predictive advantage of the GLMMNet starts to fade away. Figure \ref{fig:RE-predictions} contrasts the random effects predictions under experiments 1 and 6: the GLMMNet is able to recover the random effects very well under low noise (experiments 1--3, plots for 2 and 3 are omitted), but the high level of noise in experiment 6 creates more difficulties for the model to accurately predict the random effects.

\subsection{Application to Real Insurance Data} \label{app:case-study}
Tables~\ref{tab:lognorm} and \ref{tab:loggamma} present the in-sample and out-of-sample metrics for the lognormal and loggamma models, respectively. The results for the top performing models are included in Table~\ref{tab:cs-results} of Section~\ref{sec:case-study}.

\begin{table}[H]
\begin{center}
\begin{small}
\begin{tabular}{@{}lcccccc@{}}
\toprule
& \multicolumn{3}{c}{\textbf{Training}} & \multicolumn{3}{c}{\textbf{Test (out-of-sample)}} \\ 
& MedAE  & CRPS  & NLL  & MedAE  & CRPS  & NLL \\ \midrule
GLM\_ignore\_cat & 4349 & 0.7934 & 9.643 & 4323 & 0.8010 & 9.635 \\
GLM\_one\_hot & 4174 & 0.7791 & 9.623 & 4108 & 0.7931 & 9.623 \\
GLM\_GLMM\_enc & 4332 & 0.7919 & 9.642 & 4304 & 0.7982 & 9.632 \\
GBM\_ignore\_cat & 3893 & 0.7676 & 9.607 & 3952 & 0.7707 & 9.589 \\
GBM\_one\_hot & 3846 & 0.7653 & 9.604 & 3903 & 0.7682 & 9.586 \\
GBM\_GLMM\_enc & 3838 & 0.7644 & 9.603 & 3870 & 0.7666 & 9.584 \\
NN\_ee & 3990 & 0.7581 & 9.596 & 4086 & 0.7666 & 9.584 \\
GBM\_ee & 3825 & 0.7646 & 9.604 & 3828 & 0.7665 & 9.584 \\
GLMM & 3835 & 0.7650 & 9.604 & 3864 & 0.7666 & 9.584 \\
GPBoost & 3878 & 0.7697 & 9.611 & 3901 & 0.7711 & 9.590 \\
GLMMNet & 3273 & 0.7378 & 9.569 & 3783 & 0.7751 & 9.595 \\
GLMMNet\_l2 & 3423 & 0.7534 & 9.590 & \textbf{3549} & \textbf{0.7634} & \textbf{9.580} \\ \bottomrule
\end{tabular}
\end{small}
\end{center}
\vspace*{-5mm}
\caption[Comparison of lognormal model performance on training and testing sets]{Comparison of \textbf{lognormal} model performance (median absolute error, CRPS, negative log-likelihood) on training and test sets. The best values for each out-of-sample metric are bolded.} \label{tab:lognorm}
\end{table}

\begin{table}[H]
\begin{center}
\begin{small}
\begin{tabular}{@{}lcccccc@{}}
\toprule
& \multicolumn{3}{c}{\textbf{Training}} & \multicolumn{3}{c}{\textbf{Test (out-of-sample)}} \\ 
& MedAE  & CRPS  & NLL  & MedAE  & CRPS  & NLL \\ \midrule
GLM\_ignore\_cat & 2036 & 0.8612 & 9.787 & 2000 & 0.8733 & 9.792 \\
GLM\_one\_hot & 1942 & 0.8334 & 9.730 & 1946 & 0.8557 & 9.751 \\
GLM\_GLMM\_enc & 2043 & 0.8606 & 9.787 & 2011 & 0.8722 & 9.792 \\
GBM\_ignore\_cat & 1591 & 0.7641 & 9.606 & 1568 & 0.7668 & 9.583 \\
GBM\_one\_hot & 1578 & 0.7619 & 9.603 & 1545 & 0.7643 & 9.580 \\
GBM\_GLMM\_enc & 1594 & 0.7610 & 9.602 & \textbf{1536} & 0.7626 & 9.578 \\
NN\_ee & 1594 & 0.7429 & 9.581 & 1606 & \textbf{0.7612} & 9.578 \\
GBM\_ee & 1589 & 0.7607 & 9.602 & 1549 & 0.7629 & 9.579 \\
GLMM & 1598 & 0.7618 & 9.602 & 1570 & 0.7629 & \textbf{9.577} \\
GPBoost & 1651 & 0.7702 & 9.613 & 1632 & 0.7726 & 9.590 \\
GLMMNet & 1612 & 0.7475 & 9.585 & 1633 & 0.7662 & 9.583 \\
GLMMNet\_l2 & 1648 & 0.7576 & 9.598 & 1618 & 0.7626 & \textbf{9.577} \\ \bottomrule
\end{tabular}
\end{small}
\end{center}
\vspace*{-5mm}
\caption{Comparison of \textbf{loggamma} model performance (median absolute error, CRPS, negative log-likelihood) on training and test sets. The best values for each out-of-sample metric are bolded.} \label{tab:loggamma}
\end{table}

\begin{figure}[H]
\begin{center} \includegraphics[width=0.75\linewidth]{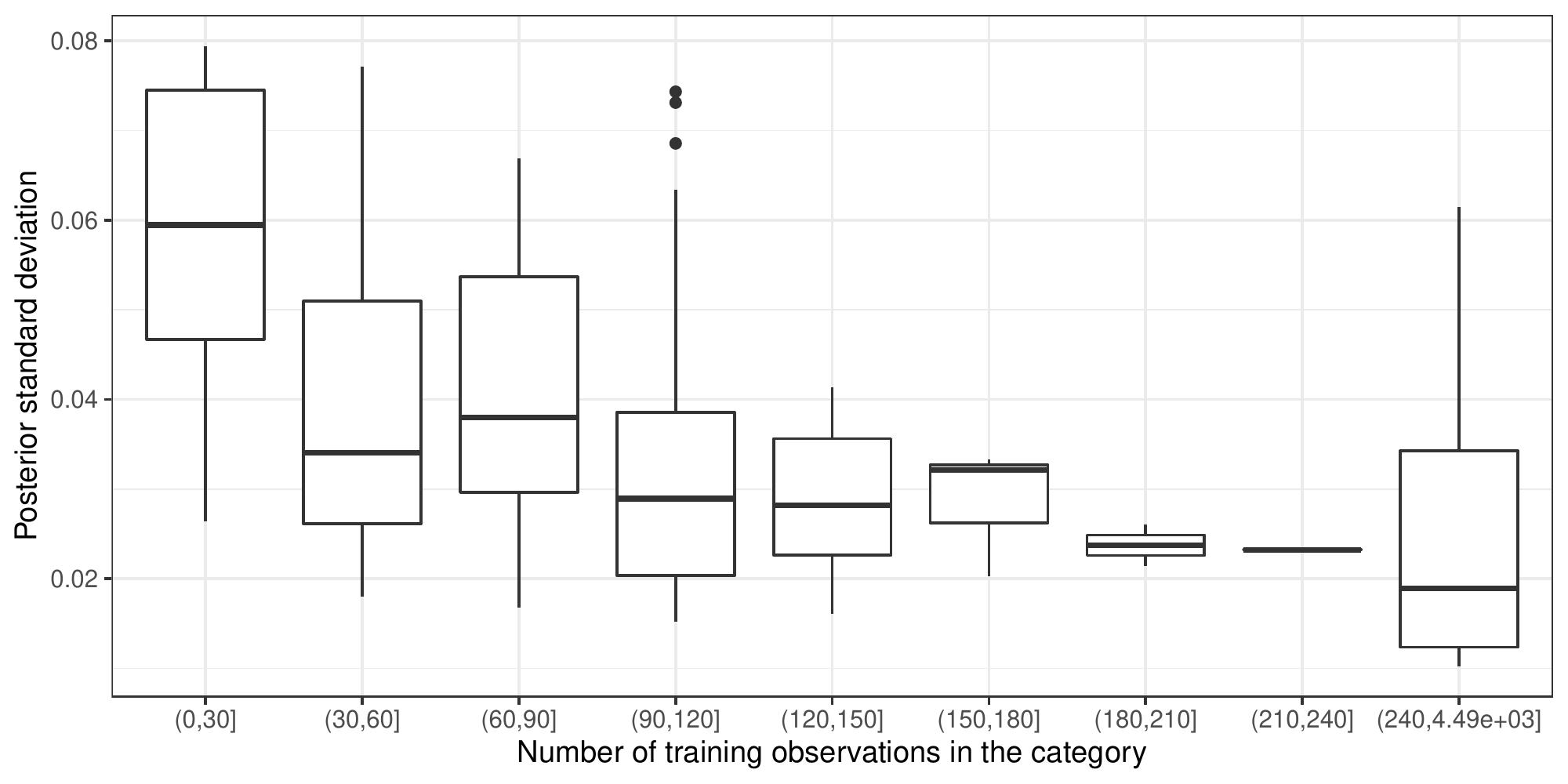} 
\caption{Posterior standard deviations plotted against the number of training observations from the occupation class}\label{fig:pr-2}
\end{center}
\end{figure}

Figure~\ref{fig:pr-2} supplements the discussion in Section~\ref{ssec:practical-insights}. It shows that, as expected, occupations with a larger number of observations generally have smaller posterior standard deviations for the random effects, indicating higher confidence in the estimates.

\end{document}